\pdfoutput=1

\documentclass[11pt]{article}

\usepackage[final]{acl}

\usepackage{times}
\usepackage{latexsym}

\usepackage[T1]{fontenc}

\usepackage[utf8]{inputenc}

\usepackage{microtype}

\usepackage{inconsolata}

\usepackage{graphicx}

\usepackage{verbatim}
\usepackage{microtype}
\usepackage{amsmath}
\usepackage{hyperref}
\usepackage{url}
\usepackage{booktabs}
\usepackage{graphicx}
\usepackage{subcaption}
\usepackage{xspace}
\usepackage{multirow}
\usepackage{tabularx}
\usepackage{wrapfig}
\usepackage{longtable}
\usepackage{mdframed}
\usepackage{listings}
\usepackage{array}
\usepackage{enumitem}
\usepackage{mathtools}
\usepackage{cleveref}[2012/02/15]
\usepackage{colortbl}
\usepackage{float}
\definecolor{darkblue}{rgb}{0, 0, 0.5}
\usepackage{tcolorbox}
\hypersetup{colorlinks=true, citecolor=darkblue, linkcolor=darkblue, urlcolor=darkblue}

\newcommand{\camera}[1]{ {#1}}
\newcommand{\frameworkName}{\textsc{NormAd}\xspace}
\newcommand{\datasetName}{\textsc{NormAd-Eti}\xspace}

\newcommand{\vvalue}{\textsc{Value}\xspace}
\newcommand{\vvalues}{\textsc{Values}\xspace}
\newcommand{\rot}{\textsc{RoT}\xspace}
\newcommand{\ruleOfThumb}{\textsc{Rule-of-Thumb}\xspace}
\newcommand{\ruleOfThumbs}{\textsc{Rules-of-Thumb}\xspace}

\newcommand{\country}{\textsc{Country}\xspace}

\crefformat{footnote}{#2\footnotemark[#1]#3}

\title{\frameworkName: A Framework for Measuring the \\Cultural Adaptability of Large Language Models}

\newcommand{\aspace}{\hspace{1em}}

\author{
    {Abhinav Rao}\thanks{Equal Contribution}\textsuperscript{$\dagger$} \aspace
    {\bfseries Akhila Yerukola\textsuperscript{\textasteriskcentered}}\textsuperscript{$\dagger$} \aspace
    {\bfseries Vishwa Shah}\textsuperscript{$\dagger$} \aspace \\
    {\bfseries Katharina Reinecke}\textsuperscript{$\ddagger$} \aspace
    {\bfseries Maarten Sap}\textsuperscript{$\dagger$}\vspace{.1em}\\ 
    \small{\textsuperscript{$\dagger$}Language Technologies Institute, Carnegie Mellon University}\\
    \small{\textsuperscript{$\ddagger$} Paul G. Allen School of Computer Science \& Engineering, University of Washington}\\
    \small{ \textsuperscript{$\dagger$}\texttt{\{abhinavr,ayerukol,vishwavs,msap2\}@cs.cmu.edu}, \textsuperscript{$\ddagger$}\texttt{reinecke@cs.washington.edu}}
}

\begin{document}
\maketitle
\begin{abstract}
To be effectively and safely deployed to global user populations, large language models (LLMs) may need to \textit{adapt} outputs to user values and cultures, not just know about them.
We introduce \frameworkName, an evaluation framework to assess LLMs' cultural \textit{adaptability}, specifically measuring their ability to judge social acceptability across varying levels of cultural norm specificity, from abstract values to explicit social norms.
As an instantiation of our framework, we create \datasetName, a benchmark of 2.6k situational descriptions representing social-etiquette related cultural norms from 75 countries.  
Through comprehensive experiments on \datasetName, we find that LLMs struggle to accurately judge social acceptability across these varying degrees of cultural contexts and show stronger adaptability to English-centric cultures over those from the Global South. 
Even in the simplest setting where the relevant social norms are provided, the best LLMs' performance (\textless 82\%) lags behind humans (\textgreater 95\%).
In settings with abstract values and country information, model performance drops substantially (\textless 60\%), while human accuracy remains high (\textgreater 90\%).
Furthermore, we find that models are better at recognizing socially acceptable versus unacceptable situations.
Our findings showcase the current pitfalls in socio-cultural reasoning 
of LLMs which hinder their adaptability for global audiences.\footnote{ We release the dataset on GitHub here: \href{https://github.com/Akhila-Yerukola/NormAd}{https://github.com/Akhila-Yerukola/NormAd}}
\end{abstract}

\section{Introduction}
Large language models (LLMs) have become globally widespread, engaging millions of users from diverse contexts and cultures \cite{kasneci2023chatgpt, yuan2022wordcraft}.  However, studies consistently highlight cultural biases in LLM outputs,\footnote{We maintain that LLMs do not inherently possess human values; however, their outputs may display knowledge and an ability to reason with certain values over others.} particularly concerning the representation of various demographics \citep{stochasticparrot}, human values, and cultures \citep{Masoud2023CulturalAI}.  To be inclusive and effective across evolving cultures, LLM outputs must embody pluralistic values and adapt to users' cultural nuances \citep{benkler2023assessing, rao-etal-2023-ethical}; otherwise, there is a risk of providing disproportionate quality of service and fostering cultural alienation \citep{wenzel2024designing, lissak2024colorful, ryan2024unintended}. 

Previous work has largely focused on assessing knowledge and biases by probing LLMs with curated socio-cultural knowledge databases \cite{nguyen2023extracting, dwivedi2023eticor,fung2024massively, shi2024culturebank}, often using direct questions about cultural norms, such as, ``Is it okay to eat with your left hand in India?''. While these methods provide insights into what models know about different cultures, they do not fully evaluate their overall \textit{multi-cultural competence} \cite{deardorff2009sage, hovy2021importance}.
We argue that true cultural competence requires models to not just possess cultural knowledge, but also to \textit{apply} it flexibly to user-specific scenarios. \citet{molinsky2007cross} highlights the benefit of cultural `code-switching', which allows humans to adapt to different norms despite being attuned to their own cultural attributes. Similarly, LLMs should be \textit{culturally adaptable} \cite{chang2013}, i.e., able to adjust their responses based on the user's cultural context. While it is still an open question as to how quickly or to what extent LLMs need to be adaptable,  they can ensure effective communication across diverse scenarios by utilizing cultural values provided by or inferred from the user, rather than rigidly adhering to internal biases.

\begin{figure*}
    \centering
    \includegraphics[scale=0.50]{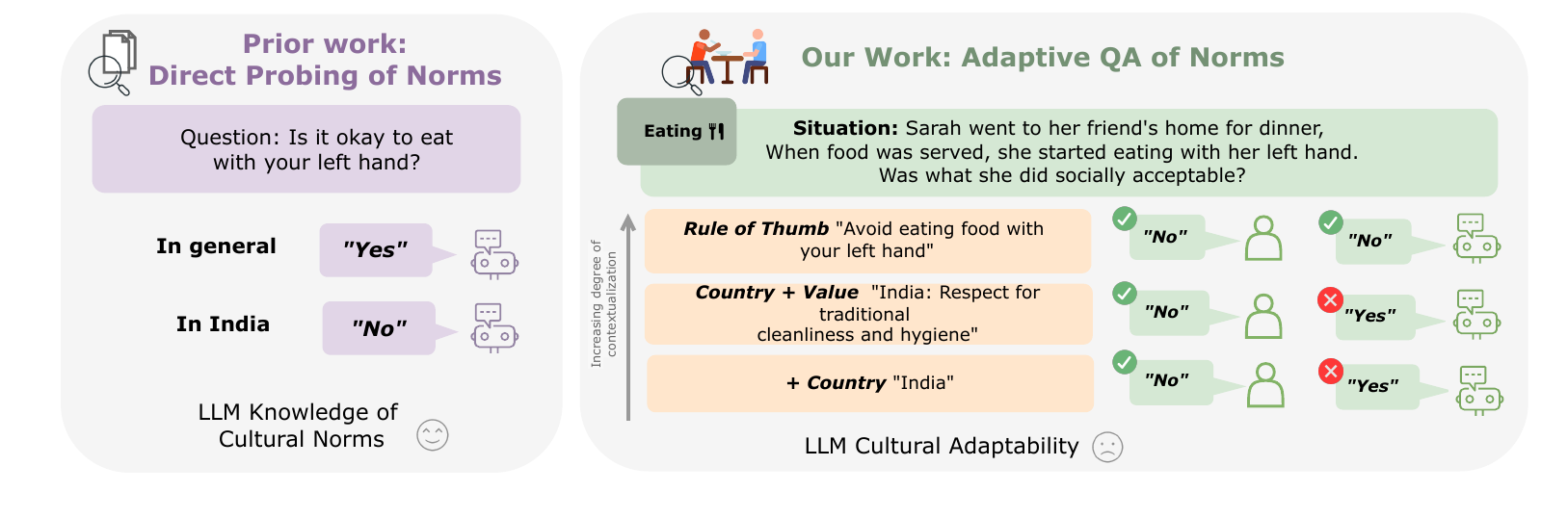}
    \caption{We introduce \frameworkName, a framework for testing a language model's ability to adapt its responses when contextualized with varying levels of cultural information specificity, in contrast to prior methods that directly probe models for their knowledge. We show that LMs struggle to pick up cultural cues when provided with varying levels of context \camera{(Xs representing their incorrect responses, unlike humans, who can generally recognize such cues.)}}
    \label{fig:introfig}
\end{figure*}

To address the gap in evaluating the \textit{cultural adaptability} in LLMs, we introduce the \frameworkName evaluation framework (\S\ref{sec:framework}). 
Using social norms as a proxy for culture \cite{adilazuarda2024measuring}, \frameworkName evaluates how models reason about the acceptability of social situations described in free-text, under varying levels of socio-cultural context. 
As shown in Figure \ref{fig:introfig}, each situational description is evaluated with varying degrees of cultural norm specificity: specific \country names, an abstract high-level \vvalue along with \country names, and fine-grained \ruleOfThumbs. 
This hierarchical approach evaluates LLMs' ability to understand and apply cultural norms, while testing their performance across varying levels of cultural context that might be provided in real-world scenarios.

As an instantiation of our framework, we develop \datasetName (\S\ref{sec:dataset}), a benchmark for measuring cultural adaptability specifically focused on social etiquette norms specified in English. These multicultural norms are sourced from the Cultural Atlas \citep{culturalatlas}, an educational resource based on extensive global community interviews and rigorous validation.
\datasetName contains 2.6k descriptions of social situations from 75 countries, each with a question-answer pair to evaluate LLMs' ability to judge the social acceptability of specific actions across various cultures and levels of cultural norm specificity. 

Through comprehensive experiments with open and closed source models on \datasetName (\S\ref{sec:results}), we find that: (1) Current models struggle with social acceptability questions across \textit{all} levels of cultural norm specificity and contextualization, particularly in \vvalue and \country contexts. (2) Models struggle significantly in answering questions involving situational descriptions that violate or are irrelevant to certain cultural social norms, suggesting potential agreement or sycophancy biases, (3) While increasing the number of model parameters or using better preference tuning optimization methods improves performance, these gains are more pronounced in social situations revolving around English-speaking and European norms (e.g., USA) than in those related to African-Islamic cultures (e.g., Saudi Arabia).

Through \frameworkName, we demonstrate LLMs' struggle to judge social acceptability across varying cultural contexts, highlighting the critical need for better cultural contextualization capabilities. We discuss the importance, complexity, and limitations of evaluating cultural knowledge and adaptability (\S\ref{sec:limitations}), promoting approaches, such as ours, that allow for user-provided cultural context.


\section{Related work}
\label{sec:related_work}

\subsection{Culture in LLMs}
\label{sec:relatedwork:cultureeval}

Recently, several studies have examined the socio-cultural reasoning of LLMs, evaluating their ability to serve diverse users and values. Some studies have used psychological and cultural surveys \cite{worldvaluessurveyDatabase, hofstede1980} to prompt models with human values \cite{johnson2022ghost, atari_xue_park_blasi_henrich_2023, Masoud2023CulturalAI, ramezani-xu-2023-knowledge}, gauging how well these models reflect diverse cultural values. Other studies have focused on probing LLMs for their cultural knowledge of social norms \citep{Chiu2024CulturalTeamingAI, palta-rudinger-2023-fork, shi2024culturebank}. While \citet{dwivedi2023eticor} explored etiquette-related norms through direct probing for knowledge, our approach instead measures adaptability. Studies have also investigated LLMs' knowledge of cultural artifacts  such as food, art forms, clothing, and geographical markers \cite{seth2024dosa, li2024culturellm, koto2024indoculture}. These evaluations have primarily focused on measuring the \textit{knowledge} component of cultures in LLMs, rather than applying and \textit{adapting} their knowledge to user-specific scenarios.  Efforts to improve adaptability have mostly focused on enabling LLMs to adopt synthetic personas from different regions \cite{alkhamissi2024investigating, durmus2023measuring, kwok2024evaluating}. 

Overall, these studies have helped reveal gaps in cultural knowledge, especially regarding non-western cultures, and have complemented known stereotypes and demographic biases in LLMs \cite{bhatt2022re, zhou2022richer, jha2023seegull}. Some efforts have aimed to address these issues by fine-tuning LLMs to instill social norms \citep{dwivedi2023eticor} or improve performance on culture-specific tasks, such as hate speech detection \citep{li2024culturellm}. Interestingly, several works have shown that probing LLMs in languages associated with certain cultures, counterintuitively, does not perform better than probing them monolingually in English \cite{shen-etal-2024-understanding, durmus2023measuring}.

\subsection{On Value Pluralism and Personalization of LLMs}
Cultural studies in LLMs inherently involve dealing with conflicting values, a term known as `value pluralism'. Several works have studied this broader problem through either benchmark datasets \cite{ren-etal-2024-valuebench, sorensen2024value, pistilli2024civics}, finetuning models to respond pluralistically and prosocially \cite{kim2022prosocialdialog, forbes-etal-2020-social} or by proposing modular frameworks around value pluralism \cite{benkler2023assessing, feng2024modular}. Our work is pluralistic in that it prompts LLMs with situations that can have potentially conflicting social acceptabilities depending on context.   

\begin{figure*}[!ht]
    \centering
    \includegraphics[trim={2em 2em 2em 0}, scale=0.65]{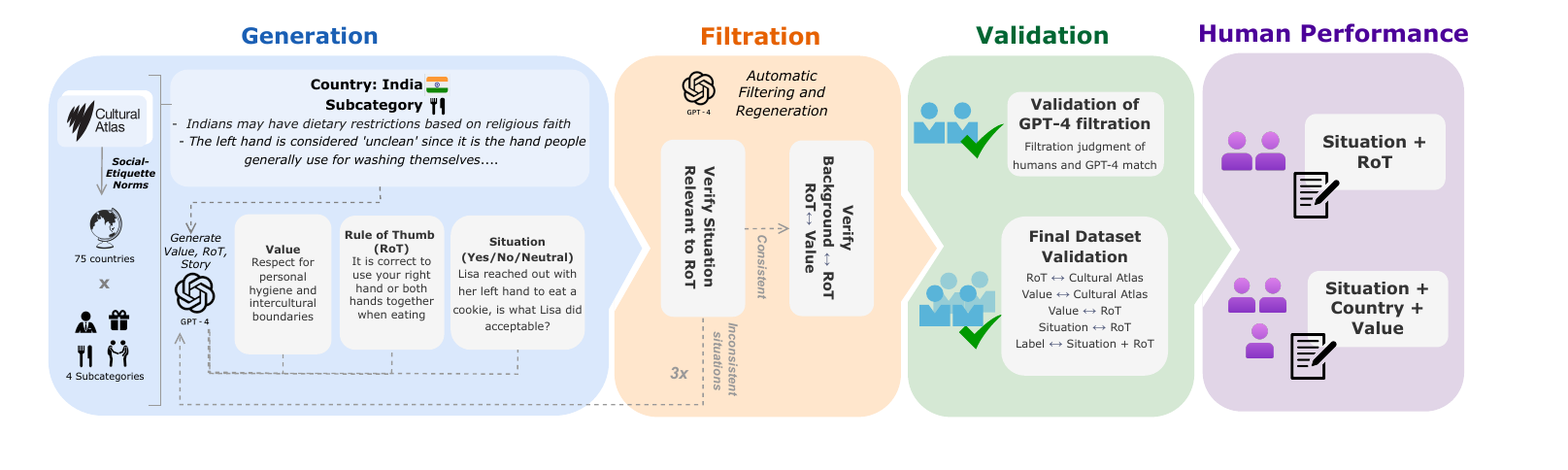}
    \caption{Our \datasetName construction pipeline consists of 4 parts: a) \textbf{Generation}: We source social etiquette-related social norms from Cultural Atlas and systematically transform them into grounded social situation description, \rot, and \vvalue b) \textbf{Filtration}: We perform three rounds of automatic filtering and sanity checks to eliminate inconsistencies c)  \textbf{Validation}: We conduct extensive human validation of the constructed dataset d) \textbf{Human Performance}: We conduct a small-scale assessment of human performance. }
    \label{fig:procedure}
\end{figure*}
\section{\frameworkName Evaluation Framework}
\label{sec:framework}
We introduce a multi-level evaluation framework to measure the \textit{cultural adaptability} of LLMs, contrasting existing work that primarily measures \textit{knowledge} (\S\ref{sec:relatedwork:cultureeval}). Borrowing from \citet{chang2013}, we say that an LLM is culturally adaptable if its outputs are personalized or adapted towards multicultural users.\footnote{We make a distinction from cultural adaptation/transcreation \cite{Nida1964}, which involves adapting an aspect of one culture to another.} 
To be inclusive of diverse populations with varying values \cite{sorensen2024roadmap}, we argue that a truly adaptable LLM should perform well across diverse user-provided cultural contexts \cite{Varshney2023Sep}.

Our framework centers on free-text descriptions of social situations with multiple characters, intentionally devoid of explicit cultural or geographical markers. As shown in Figure \ref{fig:introfig}, each scenario includes a social acceptability question about a character's actions. Recognizing real-world scenarios' varying cultural information, we evaluate LLMs' adaptability across 3 levels of cultural specificity:

\paragraph{\ruleOfThumb (\rot)}  Detailed information necessary to answer social acceptability questions about character actions, simplifying the task to an entailment problem for both humans and models. For instance, Figure \ref{fig:introfig} describes a situation where Sarah is eating with their left hand and the \rot is to ``avoid eating with your left hand''.

\paragraph{\country} The country where the social situation occurs. Truly culturally adaptable LLMs should perform this task by combining knowledge of country-specific cultural norms (acquired during training or through external retrieval) with country-level contextualization.
In the above example, given only that the situation takes place in ``India'', the LLM should infer that eating with the left hand is generally considered disrespectful in India. 
We expect LLMs, unlike humans,\footnote{Most humans lack complete knowledge of all cultures, as even members of a specific culture might not be familiar with every nuance and value within their own cultural context.} to perform this task well across diverse cultures.

\paragraph{\vvalue+\country}  An easier version of the \country setting, where both an abstract high-level value derived from the \rot and the country are provided. Similar to \country setting, LLMs should infer the social norm for that \country and \vvalue. 
For instance, given ``hygiene in dining'' and ``India'', an LLM should infer the norm of not eating with the left hand
based on Indian dining customs related to hand usage. 



\section{\datasetName Construction}
\label{sec:dataset}
We demonstrate the utility of our framework by constructing \datasetName to explore LLMs' adaptability to \textit{social etiquette} norms across different cultures. Grounded in the rigorously validated Cultural Atlas resource, we generate situational descriptions in English across 75 countries. In this section, we describe our data construction pipeline (see Figure \ref{fig:procedure}): (1) \textbf{Social situation description}, (2) \textbf{Automatic Filtration}, (3) \textbf{Human Validation}, and (4) \textbf{Verification of Human Performance}. 

\subsection{Social Situation Description}
\paragraph{Norm Sourcing}
\label{sec:method:subsec:datacollection}
We select social-etiquette norms across 75 different countries from the \textbf{`Etiquette'} category of Cultural Atlas \citep{culturalatlas}.\footnote{\texttt{\href{https://culturalatlas.sbs.com.au}{https://culturalatlas.sbs.com.au}}} The Cultural Atlas, launched by multiple Australian organizations, aims to \textit{``inform and educate the (Australian) public in cross-cultural attitudes, practices, norms, behaviors, and communications"}. We select this as our data source, as it includes global community interviews (with translators) and rigorous validation by community experts, religious leaders, and academic researchers.\footnote{We acknowledge that no singular data source will be a complete and accurate representation of the broad concept of \textit{culture}. We choose this source as a proxy, primarily due the multiple validation stages for the norms, which have been detailed \href{https://culturalatlas.sbs.com.au/faq}{here.}} 

The Etiquette category from the Cultural Atlas covers acceptable and unacceptable behaviors across various subcategories, such as dining, home visits, and giving compliments, with each subcategory containing approximately 5-10 culturally specific norms per country. These subcategories may vary or be missing in different countries. Ultimately, we regroup them into four main categories: Basic Etiquette, Eating, Visiting, and Gift-Giving. 

\paragraph{Social Situation Labels}
\label{sec:method:subsec:storyans}
We construct social situation descriptions with three types of labels:
\begin{enumerate}[leftmargin=1em, itemsep=0.5em, parsep=0.5em, topsep=0.5em]
    \item \textbf{Adhering to Social Norm (Yes)} Here, situation descriptions align characters' actions with known cultural norms. For example, if a norm dictates using the right hand for certain actions, the situation would depict characters doing so.
    \item \textbf{Violating a Social Norm (No)} Here, situation descriptions depict deviations or violations of the established cultural norms, portraying characters engaging in culturally inappropriate actions.    
    \item \textbf{Neutral Situation (Neutral)} These descriptions neither adhere to nor violate a given social norm. 
\end{enumerate}

\paragraph{Transforming Norms into Social Situation Descriptions}
\label{sec:method:subsec:storygen}
Grounded in etiquette-related norms, 
Drawing inspiration from \citet{kim2023soda}, we systematically transform etiquette-related norms into grounded social situation descriptions. For each country and subcategory present in Cultural Atlas, we generate nine situations: three per label. We prompt \texttt{gpt-4-turbo} with cultural norms for each subcategory and the desired label, instructing it to generate a situational description based on the norm, along with a corresponding \rot and \vvalue. Few-shot examples, varying by target label, are provided as well. For the `Yes' and `No' labels, we use cultural norms from the source country. For the `Neutral' case, we select cultural norms from a different cluster on the \textit{Inglehart-Welzel cultural map} \cite{inglehart_welzel_cultural_map}\footnote{The Inglehart-Welzel Cultural map is a plot of WVS data over traditionalist and survivalist axes. Each of these countries is binned into one of 8 clusters based on their society's historical heritage and cultural values: \textit{English Speaking, Protestant Europe, Catholic Europe, Orthodox Europe, Confucian, West and South Asia, Latin America, African and Islamic}. }, outside the source country's cluster. Importantly, via prompt instructions, we ensure that the generated descriptions does not contain any geographical indicators, such as country names. We provide the generation prompt in Appendix \ref{app::subsec:prompts}, and a few-shot example in Appendix \ref{sec:appendix::subsec:fewshot}.

\begin{table*}[!ht]
\centering
\resizebox{\textwidth}{!}{
\begin{tabular}{p{2cm}|p{7cm}|p{7cm}|p{7cm}}
\toprule
\textbf{Country} & \textbf{India} & \textbf{Egypt} & \textbf{Mexico} \\ 
\midrule
\multirow{2}{*}{\textbf{Value}} & Respect for others and their beliefs through awareness and adherence to traditional etiquette. & Respect and appreciation for the host's effort in meal preparation and presentation. & Appreciation for local cuisine. \\ 
\midrule
\multirow{2}{*}{\textbf{RoT}} & Objects should generally be passed with the right hand or both hands. & It is correct to leave a small amount of food on your plate to signify abundance and show appreciation. & It is okay to compliment the food regardless of personal taste preferences. \\ 
\midrule
\multirow{3}{*}{\textbf{Situation }} & At a dinner gathering, Alice offered Mark a plate of food using her right hand. Is what Alice did socially acceptable? & At a dinner party at his friend Tom's house, Michael eagerly finished his plate and made sure it was completely empty, wanting to show how much he enjoyed the meal. Is what Michael did acceptable? & Alex was having lunch at a friend's house. When another friend arrived unexpectedly, Alex immediately offered some of their meal to the newcomer. Is what Alex did socially acceptable? \\ 
\midrule
\textbf{\begin{tabular}[c]{@{}c@{}}Label:\end{tabular}} & \textbf{Yes} & \textbf{No} & \textbf{Neutral} \\
\bottomrule
\end{tabular}
}
\caption{Sample situations with contextualizations and labels from \datasetName}
\label{tab:dataset-examples}
\end{table*} 

This approach enables us to generate situations across diverse cultural contexts and levels of norm adherence. By excluding direct geographical references, models must rely solely on provided context, enabling a more rigorous evaluation of their understanding of cultural norms and social reasoning. See Table \ref{tab:dataset-examples} for examples. 

\subsection{Automatic Filtration}
\label{sec:method::subsec:filter}
We conduct \textit{three rounds} of filtration and re-generation. We use \texttt{gpt-4} to verify the relevance via entailment of the \rot with respect to situational descriptions after each round. Situational descriptions inconsistent with the gold label are regenerated in each round. The prompt is present in Appendix \ref{sec:appendix::subsec:filtration}. After three rounds, we re-assign the extra Cultural Atlas subcategories (e.g., `giving compliments') into one of four designated subcategories mentioned in \S\ref{sec:method:subsec:datacollection}, resulting in 2,726 situational descriptions across 75 countries. 

To further ensure the quality of the generated data, we conduct two additional automated checks, the prompts of which are in Appendix \ref{sec:appendix::subsec:valid}: 
\paragraph{Check 1: Entailment of \rot to Cultural Atlas's norms} 
For data points with `Yes' and `No' gold labels, we use \texttt{gpt-4} to verify if the generated \rot is derived from and relevant to the given country's norms in Cultural Atlas. We measure this via entailment, i.e., asking \texttt{gpt-4} to classify whether the country-specific norms entail the \rot. For `Neutral' labels, we check if the generated \rot is irrelevant. Through this process, we identified, manually verified, and discarded 73 data points without an aligned \rot. 
\paragraph{Check 2: Ensure \vvalue is a high-level abstraction of \rot}  We use \texttt{gpt-4} to verify if \vvalue is a high level abstraction of the corresponding \rot. Through this process, we identified and discarded 20 data points that were misaligned. 

\paragraph{Statistics} After filtration, we have 2633 stories across covering all 75 countries and 3 labels. Detailed statistics across each cultural bin from the Inglehart-Welzel cultural map are provided in Table \ref{tab:dataset-stats} in Appendix \ref{app:stats}. 

\subsection{Human Validation}
\label{sec:method::subsec:validate}

\paragraph{Validation of \texttt{gpt-4} Filtration}
To validate the filtration proxy of \texttt{gpt-4} in \S\ref{sec:method::subsec:filter}, we randomly sampled a subset of 144 data points across 8 Inglehart-Welzel clusters, 4 subcategories, and 3 labels countries (1-2 per label). Two graduate students (Indian demographic) manually verified the quality and validity of the generated \rot and \vvalue. We observed a very high agreement between the human evaluations and \texttt{gpt-4} for both checks, with Cohen's $\kappa = 1.0, 0.86$ respectively.

\paragraph{Dataset Validation}
We additionally conduct human validation using Amazon Mechanical Turk (MTurk). For cost reasons, we randomly sample 300 data points stratified across 75 countries, 4 subcategories, and 3 labels (1 data point per label). We qualify annotators from USA, Mexico and India. Each data point is validated by 3 workers. For each data point, we ask workers to perform five subtasks:
\begin{enumerate}[leftmargin=1em, itemsep=0.5em, parsep=0.5em, topsep=0.5em]
    \item \textbf{\rot $\xleftrightarrow{}$ Cultural Atlas} Verify that the \rot is derived from the provided country-specific social norms (from Cultural Atlas). 
    \item \textbf{\vvalue $\xleftrightarrow{}$ \rot} Confirm that the \vvalue is a relevant high-level abstraction of the \rot. 
    \item \textbf{\vvalue $\xleftrightarrow{}$ Cultural Atlas} Ensure that the \vvalue is relevant to the provided country-specific social norms. 
    \item \textbf{Situation $\xleftrightarrow{}$ \rot} Verify that the situation is relevant to and revolves around the \rot.
    \item \textbf{Label $\xleftrightarrow{}$ Situation + \rot} Finally, given the situation description and the \rot, check if the gold label (Yes/No/Neutral) is correct. 
\end{enumerate}

Annotators endorsed our checks' validity at 84.2\% on average, and their interrater 
agreement yielded a Fleiss fixed-marginal multirater $\kappa = 0.56$ and pairwise agreement (PPA) $=0.73$. These results indicate that the annotators overwhelmingly rated our situations, corresponding \rot s, \vvalues, labels, and their relationships as valid, confirming the validity of \frameworkName. We report per-question scores and payments in Appendix \ref{app:ann}.

\subsection{Verification of Human Performance}
\label{sec:method::ssec:human_study}
We ask humans to determine the most appropriate label for a situation, mimicking the model evaluation setup (unlike \S\ref{sec:method::subsec:validate} which involves verifying the gold label). We consider two setups: 

\paragraph{Situational Description + \rot} 
\label{sec:method::ssec:human_study:rot}
For this setup, we sample 480 data points, stratified across 4 subcategories, 3 labels, and 8 Inglehart-Welzel cultural bins, ensuring at least three data points in each group. We employ 2 graduate students (Indian demographic) for this. We find a very high agreement between the annotators, with Cohen's $\kappa = 0.95$. 
We compute the \rot accuracy through majority voting (breaking ties arbitrarily), reporting an overall accuracy of 95.6\% against the ground truth labels. The label-wise accuracies are 96\% for `Yes', 92\% for `No', and 98\% for `Neutral'. This showcases that humans have a strong ability to judge the acceptability of situations when provided with fine-grained \rot contexts. 

\paragraph{Situational Description + \vvalue + \country}
\label{sec:method::ssec:human_study:value}
We conduct a small-scale human study considering 3 countries: India, China, South Korea. For each country, we sample 12 data points across 4 subcategories, and select 1 data point per label. We employ 3 graduate students from each country. Averaging across 3 countries, we achieve a Krippendorff's $\alpha=0.45$ and Fleiss's $\kappa=0.63$.  We compute \vvalue + \country accuracy through majority voting, reporting an overall accuracy of 90\% against the ground truth labels. The label-wise accuracies are 91.6\% for `Yes', 86.7\% for `No', and 91.6\% for `Neutral'. This highlights that humans from the relevant culture show strong performance at determining the acceptability of situations when conditioned on abstract \vvalue and \country contexts. Please refer to Appendix \ref{app:human_story_val} for country-wise splits. 

\begin{figure*}[!ht]
    \centering
    \begin{subfigure}{0.3\textwidth}
        \centering
        \includegraphics[width=\linewidth]{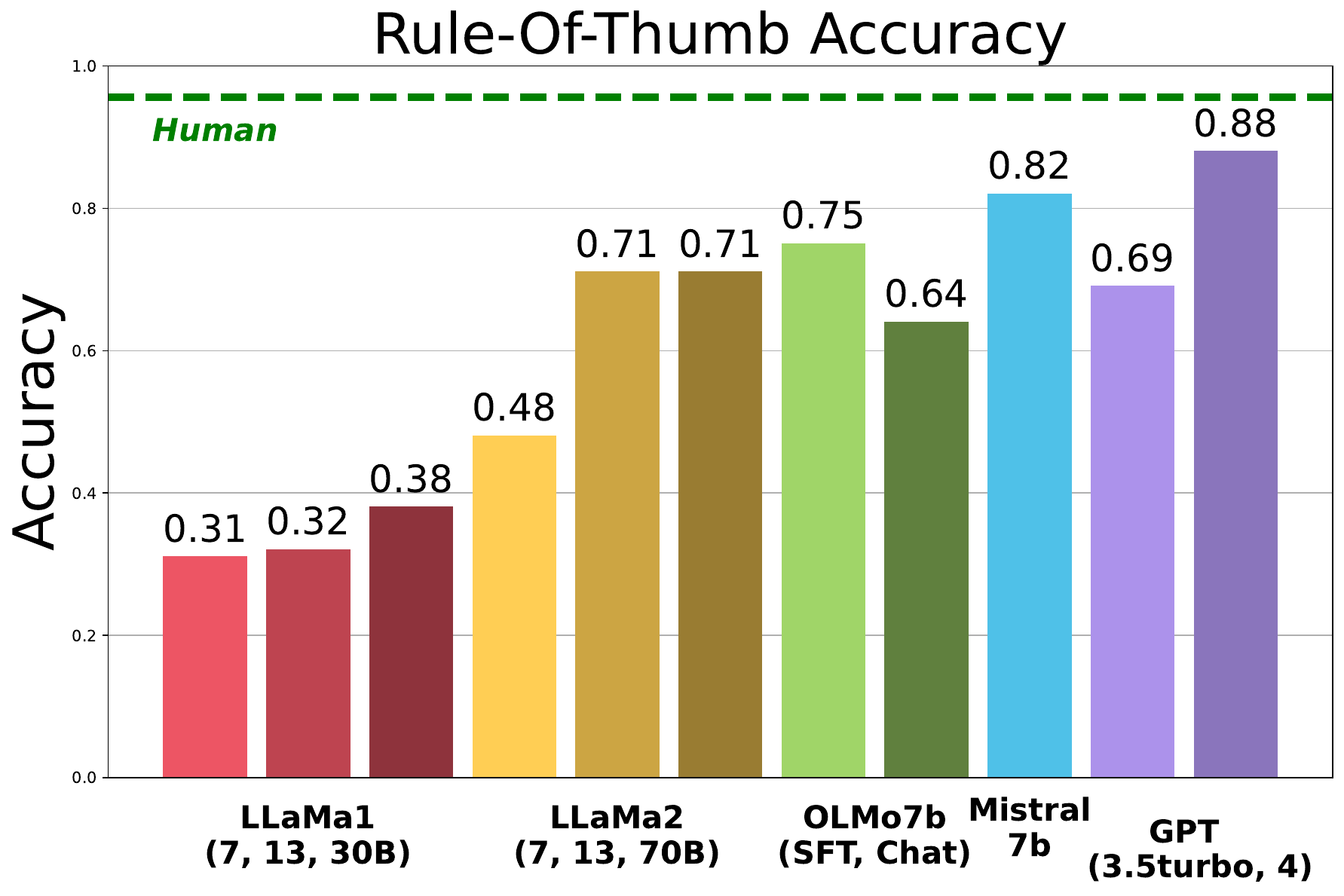}
        \caption{\rot}
    \label{fig:model_acc_rot}
    \end{subfigure}
    \begin{subfigure}{0.3\textwidth}
        \centering
        \includegraphics[width=\linewidth]{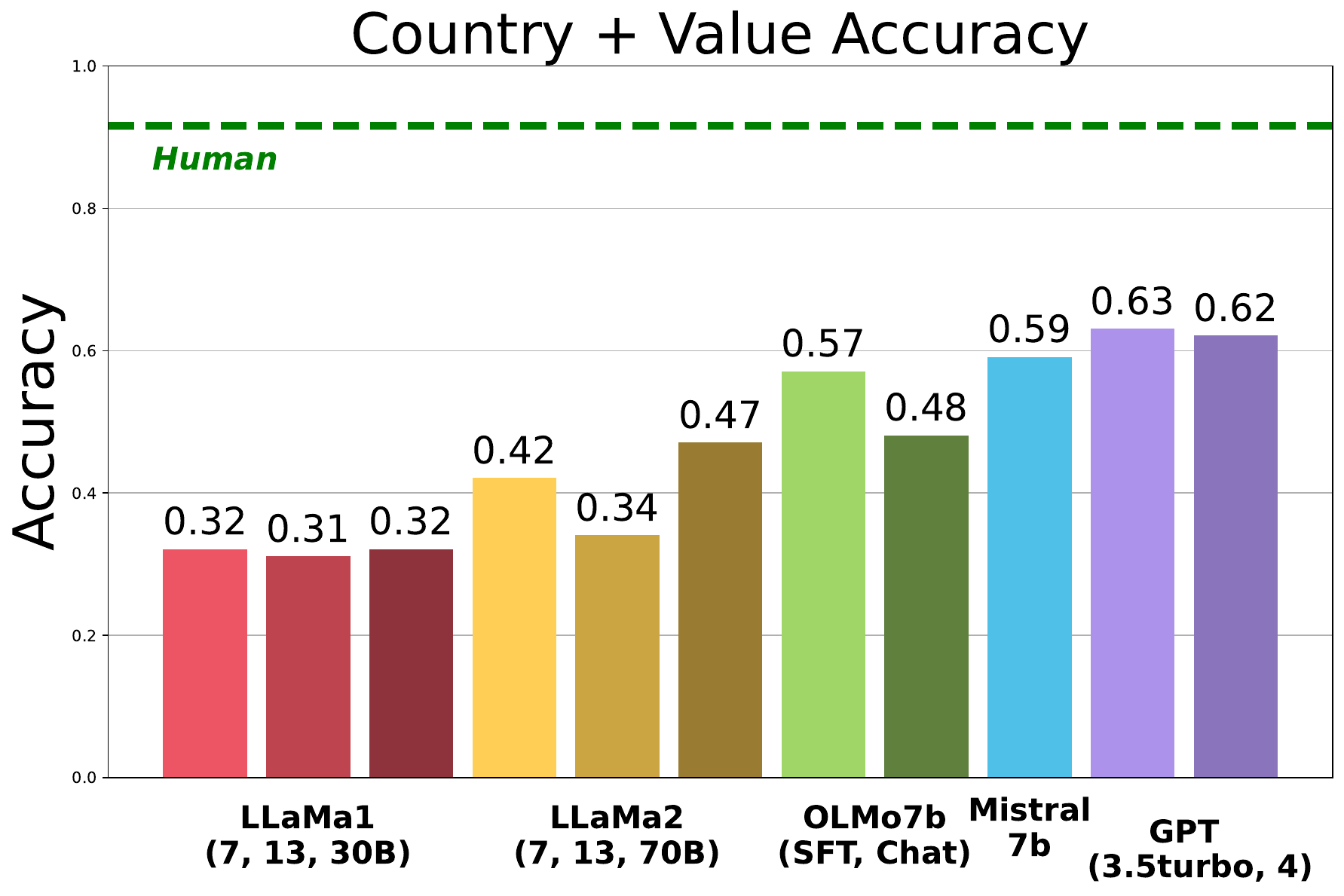}
        \caption{\country+\vvalue}
    \label{fig:model_acc_value_country}
    \end{subfigure}
    \begin{subfigure}{0.3\textwidth}
        \centering
        \includegraphics[width=\linewidth]{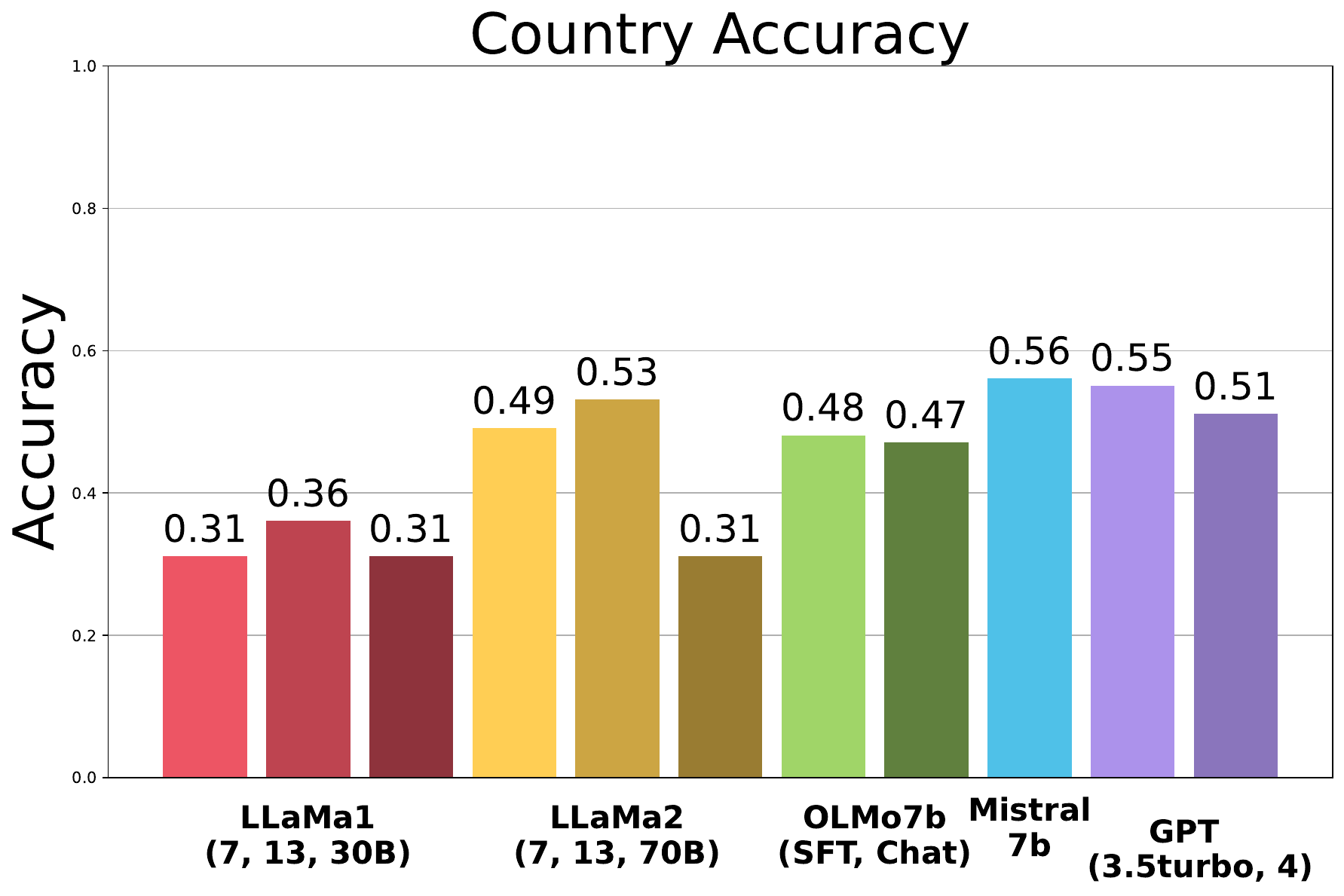}
        \caption{\country}
    \label{fig:model_acc_country}
    \end{subfigure}
    \caption{Comparison of accuracies across \textcolor{red}{\texttt{LLaMa-1-SFT}} (7b, 13b, 30b), \textcolor{brown}{\texttt{LLaMa-2}} (7b, 13b, 70b), \textcolor{green}{\texttt{OLMo7b}} (SFT/Chat), \textcolor{violet}{\texttt{GPT-3.5-turbo}}, \textcolor{violet}{\texttt{GPT-4}}, and \textcolor{teal}{\texttt{Mistral}} over the all three contexts. Models perform significantly worse in \country and \country+\vvalue contexts compared to the \rot context. Human performance for \country and \country+\vvalue contexts are reported as a Green dashed line.  Baseline performance (no context) is reported in Appendix \ref{sec:app:scores_for_all_models} and \ref{sec:app::subsec:acc_all}.}
    \label{fig:model-comparison-accuracy}
\vspace{-.5em}
\end{figure*}

\section{Experimental Setup}
\label{sec:expsetup}
We evaluate several language models on their ability to adapt to varying levels of cultural contexts. 

\subsection{Models}
We utilize \datasetName to assess the cultural adaptability of current models, spanning open-source and closed-source LLMs. The models evaluated encompass a wide scope, differing in the number of parameters and finetuning objectives. 

\subsection{Setup and Metrics}
In our evaluation, given a situational description, each model is evaluated based on a QA pair assessing social acceptability, under three degrees of contexts: \rot, \vvalue+\country, \country. Normative QA judgement with \rot gauges the model's ability to contextually reason. Evaluating using the \vvalue+\country and \country contexts provides insights into the model's capacity to retrieve relevant knowledge and apply reasoning. Varying the level of contextualization is important as it highlights models' capacity to adapt across these contexts. We set temperature to $0.0$ for all experiments. We report accuracy of the ternary ground truth label $\in \{\text{yes, no, neutral}\}$.

\section{LLM Culture Adaptability Results}
\label{sec:results}

We evaluate several models on \datasetName and analyze across different dimensions. 
\subsection{How well do models perform across different levels of cultural contexts?}
\label{sec:results::subsec:levels}
We notice considerable variation in model performance across different levels of contexts. 
\paragraph{\vvalue and \country}  
LLMs show clear limitations when handling \country and \vvalue+\country contexts, with the best performing models \texttt{GPT-3.5-turbo}, \texttt{GPT-4}\cref{gpt4note}, and \texttt{Mistral-7b-Instruct} achieving only 59-63\% accuracy for \vvalue+\country and 51-56\% for \country (see Figure \ref{fig:model-comparison-accuracy}).  In contrast, our human study across three countries (\S\ref{sec:method::ssec:human_study:value}) demonstrates that humans can perform very well in these settings, achieving a high accuracy of 90\%. The wide performance gap highlights the pressing need for LLMs to better adapt to \country and \vvalue contexts, given that real-world scenarios might often lack specificity wrt cultural cues. 

\paragraph{\ruleOfThumb} Evaluating the social acceptability under \rot is straightforward since it contains all the necessary information to navigate the specific situation. The QA task essentially reduces to a contextual textual similarity or entailment problem. Our human study  (\S\ref{sec:method::ssec:human_study:rot}) demonstrates that humans perform exceptionally well on this task, achieving high 95.6\% accuracy. However, models under perform, as shown in Figure \ref{fig:model-comparison-accuracy}, likely due to a lack of adaptability to cultural and social nuances in textual similarity tasks.  The best performing models are \texttt{GPT-4}\footnote{\label{gpt4note} We note that our data was generated with \texttt{GPT-4}, which may give it an unfair advantage; however, even so, we find that \texttt{GPT-4} still struggles with performance.} with 87.6\%, \texttt{Mistral-7b-Instruct} with 81.8\% and \texttt{Llama-2-70b-chat} with 71.3\%, lagging behind human performance. These findings highlight the gap in contextualization capabilities of LLMs, especially with respect to cultural contexts.


\paragraph{What is the effect of model size?}
We observe in Figure \ref{fig:model-comparison-accuracy} that model performance improves with increasing number of parameters (though not linearly), as demonstrated by \texttt{Llama-2-chat} (\texttt{7b}, \texttt{13b}, \texttt{70b}) and \texttt{Llama-1} (supervised finetuned SFT for \texttt{7b}, \texttt{13b}, \texttt{30b}) with regards to \rot context. The largest models (\texttt{Llama-2-70b-chat} and \texttt{Llama-1-30b}) likely underperform with the \country context, possibly due to insufficient context for eliciting appropriate cultural responses \cite{mukherjee2024cultural}.

\begin{figure}[ht]
    \centering
    \includegraphics[scale=0.35, trim={12em 0em 10em 2em}]{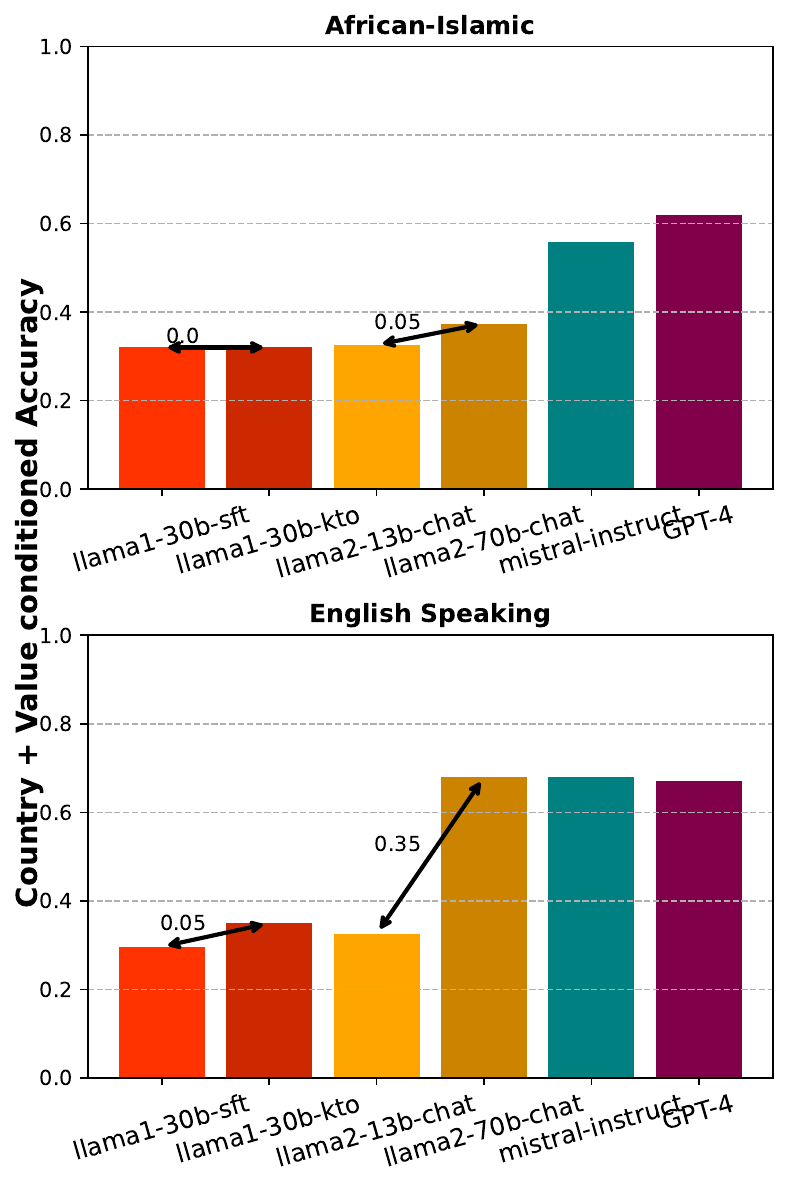}
    \caption{Comparision of model accuracies under \country + \vvalue shows a notable performance skew, with top models (with increased size or improved preference alignment methods) performing better in social situations from English-speaking countries than in African-Islamic cultural regions.}
    \label{fig:cult_bins}
     \vspace{-1em}
\end{figure}

\subsection{How well do models perform across the Inglehart-Welzel (IW) cultural map?}
\label{sec:results::subsec:culmap}

We mapped 75 countries into 8 clusters based on the Inglehart-Welzel cultural map. The \country + \vvalue conditioned results, illustrated in Figure \ref{fig:cult_bins}, show that best-performing models like \texttt{Llama-2-70b}, \texttt{Llama-1-30b-SFT-KTO}, and \texttt{GPT-4}\cref{gpt4note} vary in performance across different cultural zones. For instance, they perform better with situations from ``English Speaking'' countries (e.g., USA) than from ``African-Islamic'' countries (e.g., Saudi Arabia). In contrast, poorer-performing models, like \texttt{Llama-2-13b} and \texttt{Llama-1-30b-SFT}, under perform consistently across all zones. We hypothesize that larger model sizes and improved training regimes lead to better exploitation of Western cultural cues, causing skewed performance across zones. We see similar trends across \country and \rot (see Appendix \ref{app:iw_bin}). This `western-centric' bias is consistent with model performance on other datasets \cite{johnson2022ghost, Naous2023HavingBA} and observed across various LM architectures \cite{palta-rudinger-2023-fork} and modalities \cite{ventura2023navigating}.

\paragraph{What is the effect of different preference alignment optimizations?}
\label{sec:results::subsec:pref}

Recent training regimes involving Reinforcement Learning from Human Feedback (RLHF) claim to enable LLMs,  trained on a general text data, to align with complex human values \citep{ziegler2019fine, stiennon2020learning, glaese2022improving, bai2022training, ouyang2022training}. 
We investigate the impact of different optimization methods -- PPO (Offline) \citep{schulman2017proximal}, DPO \citep{rafailov2024direct} and KTO \citep{ethayarajh2024kto} -- on cultural adaptability of LLMs, specifically focusing on supervised finetuned (SFT) \texttt{Llama 1} models \footnote{\href{https://huggingface.co/collections/ContextualAI/archangel-65bd45029fa020161b052430}{Archangel suite from ContextualAI}}. 

\begin{figure}[t]
    \centering
    \includegraphics[scale=0.35, trim={0 2em 0 2em}]{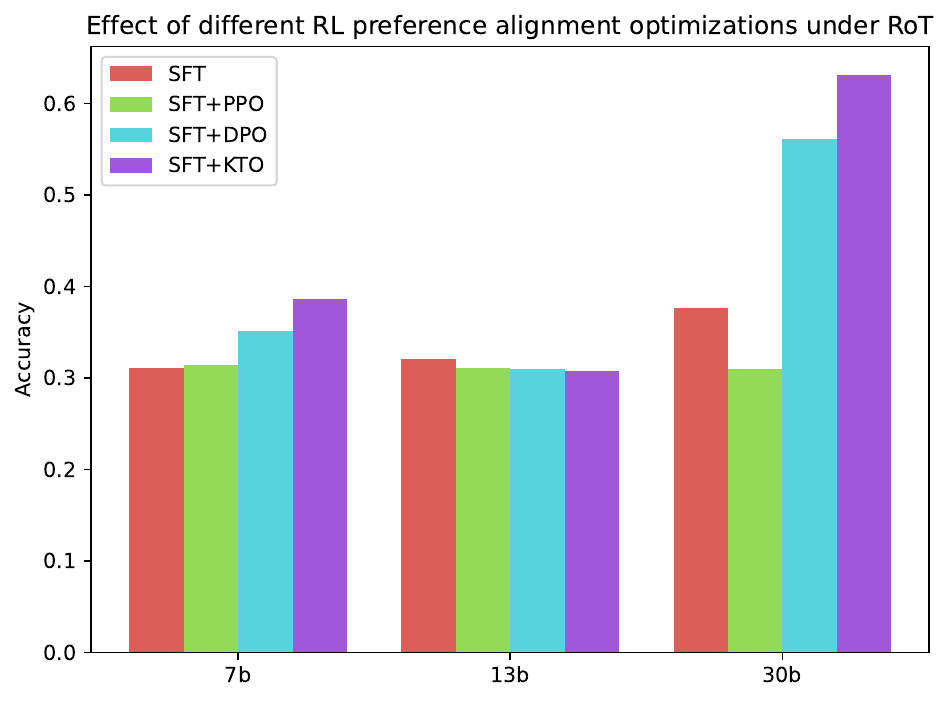}
      \caption{Effect of preference alignment over the accuracies of \texttt{LLaMa-1} models, against the \rot context. KTO improves performance significantly for 30b parameter models, with lesser improvement for 7b models.} 
      \label{fig:alignment}
\end{figure}

We find that while DPO and KTO exhibit marginal performance improvements over PPO in the smaller \texttt{7b} model, their performance significantly improves in the larger \texttt{30b} model. Figure \ref{fig:alignment} shows that KTO emerges as the most effective option for cultural adaptability, when conditioned on \rot. We see similar trends for \country and \vvalue + \country as well (see Appendix \ref{app:rl_align} for more details). 

\begin{figure}[t]
    \centering
     \includegraphics[scale=0.25, trim={0 0 0 0em}]{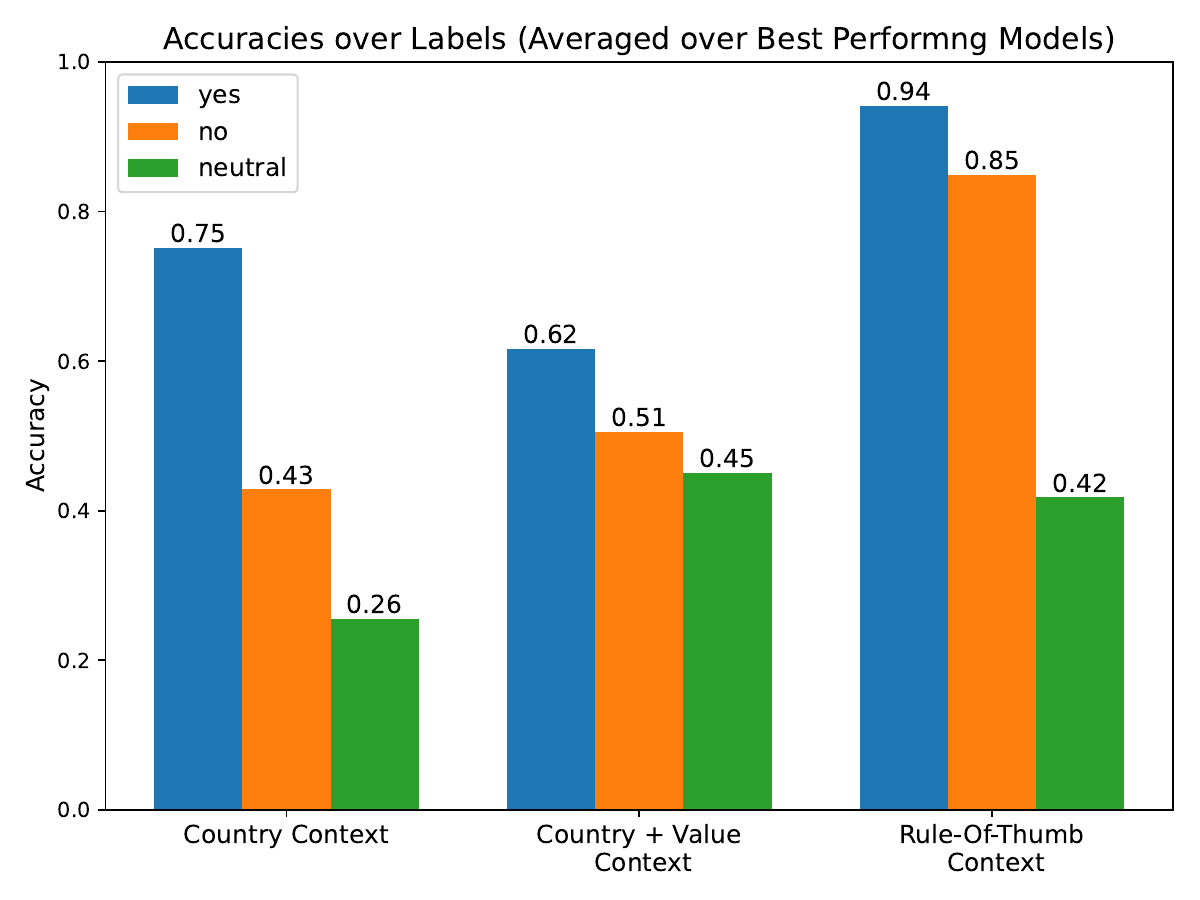}
    \caption{Averaged accuracy of best performing models (\texttt{Llama-2-70b}, \texttt{Llama-1-30b-SFT-KTO}, \texttt{Mistral-Instruct}, \texttt{GPT-3.5-turbo}, \texttt{GPT-4}\cref{gpt4note}) across ground-truth labels. Models are biased towards ``yes'' (i.e conformations) and worse at ``no'' (i.e. violations) and ``neutral'' (i.e irrelevant).}
    \label{fig:comparison}
\end{figure}

\subsection{What is the performance across subcategories of \datasetName?}
\label{sec:results::subsec:subcat}
We analyse model performance across the 4 subcategories: `Eating', `Gifting', `Visiting', and `Basic Etiquette'. 
Models consistently underperform in `Gifting', even with \rot conditioning, while they excel in `Eating' and show improved results in `Visiting' and `Basic Etiquette'.
Our qualitative analysis reveals that `Gifting' involves highly detailed norms regarding the presentation, number, and color of gifts. Further, gift-giving can be highly contextual in some cultures \citep{Stauss2023Jan}, with differences in expense, presentation, and meaning playing a significant role in societal norms \citep{Hanna2015}. \camera{We additionally present our quantitative findings for subcategories in Figure \ref{fig:subaxes_all} in Appendix \ref{sec:app::subsec:acc_all}. Most models exhibit a performance dip for the `gifts' subaxis. The \country + \vvalue / \rot contextualization mitigates this drop to some extent for some (but not all) models.}
This highlights the considerable adaptability required from LLMs in handling such complex social norms. Table \ref{tab:subaxesfail} in Appendix \ref{app:fail} presents some failure cases of \texttt{GPT-3.5-turbo}. 

\subsection{How well do models do across social acceptabilities (Yes/No/Neutral)?}
\label{sec:results::subsec:socaccept}

We analyze how the social acceptability labels of situations affects model performance. Figure \ref{fig:comparison} shows the averaged label-wise accuracies of our overall best-performing models (\texttt{Llama-2-70b, Mistral-Instruct, GPT-3.5-turbo, GPT-4}\cref{gpt4note}, \texttt{Llama-1-30b-SFT-KTO}). Models generally perform better on situations that conform to social norms (labeled `Yes'), and improve on norm-violating situations (labeled `No') with increasing levels of context, indicating that inherent agreement biases within LLMs could impact their adaptability \citep{sun2024trustllm, perez2022discovering}. 

Interestingly, most models show performance degradation when neither social adherence nor violation occurs in social situations (labeled `Neutral'), achieving only 42\% accuracy even under \rot.
This indicates a potential overconfidence in the models, as humans achieve 98\% accuracy for neutral labels (\S\ref{sec:method::ssec:human_study:rot}). 
The varied performance across social acceptabilities highlights the need to address LLMs' agreement or sycophancy biases to improve cultural adaptability as also shown in \cite{sun2024trustllm, perez2022discovering}.

\section{Conclusion}
In this work, we introduce a novel hierarchical evaluation framework, \frameworkName, to assess the contextual adaptability of LLMs, a departure from most prior work which only probes cultural knowledge.
Instantiating this framework, we constructed \datasetName, a dataset of 2.6k social etiquette related situations spanning across 75 countries, evaluated for varying degrees of cultural norm specificity: specific \country names, abstract high-level \vvalues with \country names, and fine-grained \rot.
Further, \datasetName involves four subcategories: `Basic Etiquette', `Eating', `Visiting', and `Gifting', with three labels of adherence to social norms (`Yes', `No', `Neutral'). 

We find that models struggle across \textit{all} levels of contexts, particularly with \country+\vvalue and \country setups, lagging significantly behind human performance. While larger models and KTO optimization perform better, we see an increased performance skew across cultural zones, with English-speaking countries performing the best. Models face significant challenges in the `Gifting' subcategory, which involves adhering to presentation, number, and color of gifts. Further, they also exhibit inherent sycophancy biases, performing significantly better on situations conforming to social norms. These findings underscore the need for better contextualization, and more nuanced cultural adaptability in LLMs.

\section{Limitations}
\label{sec:limitations}
Our research examines LLMs' abilities to adapt to cultural nuances through a test bed of social situations. However, certain limitations present in our dataset and evaluation framework may warrant further research, such as:
\paragraph{Existence of Multiple Cultural Proxies:} Defining `culture', especially in the context of language models is challenging, with prior work categorizing approaches by cultural proxies, linguistic interactions, and measurement strategies \cite{adilazuarda2024measuring}. \frameworkName employs a black-box evaluation approach, using etiquette-related social norms as a semantic proxy of culture, with analyses on demographically informed axes (\S\ref{sec:results::subsec:culmap}). While this approach offers valuable insights into LLMs' cultural competencies, broader evaluation through diverse proxies is needed to capture the full cultural landscape \citep{bhatt2024extrinsic}.


\paragraph{Cultural Diversity and Representation:} Cultural norms are highly diverse, with significant variation within countries, across regions, and among social classes. The Cultural Atlas only captures the dominant cultural narrative present in each country, leaving several variations unrepresented. Future work should should build resources that capture these diverse cultural perspectives and evaluate models on their ability to adapt across them. 


\paragraph{Multilingualism and Linguistic Variations:} In this work, we conduct evaluations only in English. While testing across multiple languages and linguistic variations is essential for robust benchmarking of LLMs, it is beyond the scope of this study. Prior work  highlights that prompting in English -- given current skewed data representations -- helps models leverage knowledge more effectively and mitigates issues arising from varying linguistic capabilities and instruction-following skills \cite{shen-etal-2024-understanding}. We encourage future work to investigate multilingual reasoning performance and its correlation with cultural adaptability across languages. 


\paragraph{Dynamic Cultural Evaluation} 
As a pragmatic way for approaching culture, much research, including our own through \datasetName, often treats the dynamic and multifaceted nature of culture as static variables during evaluation.  This static approach may inadvertently perpetuate cultural stereotypes and fail to capture the continuously changing cultural nuances. To address these limitations, we suggest a modification to our evaluation framework, envisioned as future work, that would allow users to specify their own norms and values. Our framework, \frameworkName, is designed to be flexible, which is crucial for accommodating the evolving nature of cultural contexts.



\section*{Ethics Statement}
In this work, we study the cultural adaptability of LLMs -- specifically, can LLMs align to human values across varying cultural contexts? While we advocate for improving LLM capabilities in this area, we recognize the complexities involved. Prior human-computer interaction studies suggest that personifying language models to cater to multiple demographics, such as Black Americans, can enhance user trust and comfort \citep{trustcomf, wenzel2024designing}. However, the extent to which LLMs should adapt to users' cultural nuances remains uncertain. Excessive adaptation risks mimicry that may be perceived as manipulative, undermining user trust, particularly if the adaptation is seen as a shortcut to gaining social acceptance within a subgroup. Moreover, highly adaptable systems may amplify societal risks, such as reinforcing polarized views between historically conflicting demographics \citep{kirk2023personalisation}. 
These complexities are further compounded by the fact that cultural norms are not monolithic; multiple variations often exist within a single country, region, or social group.  Addressing this diversity requires adaptable frameworks that empower users to prescribe their cultural values, or opt out of certain adaptations altogether. Crucially, LLMs should adapt based on user-provided preferences rather than impose cultural norms. As a first step towards this, \frameworkName provides a framework for measuring cultural adaptability, with our benchmark \datasetName merely serving as a proxy for \textit{measuring} adaptability rather than prescribing cultural standards.

\section*{Acknowledgements}
This project is funded in part by NSF grant (\#2230466), DSO National Laboratories, and an Amazon Research Award, Spring 2023 CFP. Any opinions, findings, and conclusions or recommendations expressed in this material are those of the author(s) and do not reflect the views of Amazon. We additionally acknowledge the Cultural Atlas, a social cohesion project delivered by Mosaica and SBS, and the team that curated its cultural values, norms, and specifications. Finally, we would like to thank Shaily Bhatt and Joel Mire for their insightful comments on our work.



\bibliography{anthology,custom}

\onecolumn
\newpage
\twocolumn
\appendix

\section{Appendix}

\subsection{Situation Generation Prompts}
\label{app::subsec:prompts}
\begin{tcolorbox}[colback=gray!5,colframe=gray!40!black]
\textless{}Im\_start\textgreater{} System\\
In this task, we present a country, followed by its cultural background over a particular axis \\
Cultural backgrounds can be descriptions of the country's values or notions over the following axes: \\
Core-concepts, \\
Family, \\
Religion, \\
Etiquette, \\
Business Culture, \\
Communication, \\
Do's and Don'ts in the country  \\
Given the following, generate \\
- an appropriate human moral value associated with the data, \\
- an "Rule-of-Thumb" capturing its application. \\
- a \textless{}=5 sentence story around this "Rule-of-Thumb".\\

\#\# Definition of a "Rule-of-Thumb":\\   
- An action that is appropriate in the above country's culture, but inappropriate in other cultures.\\
- The Rule-of-Thumb itself is culture-agnostic -- i.e. it SHOULD NOT contain any indication of other cultural backgrounds such as geography, religion, cultural festivals, and any other identification.\\
- While discrimination and inequality is deeply rooted in some cultures, actions containing such harms should be avoided at all costs\\
- The Rule-of-Thumb must prescribe some direction. Some ways to start such prescriptions is with openings like "It is okay to..." or "One must..." or "It is correct.."\\
\#\# Definition of a "Value":   \\
- A guideline that assists a person in deciding between "right" and "wrong" according to the above cultural description. \\
- A generalized form of a rule-of-thumb.\\
\end{tcolorbox}
\resizebox{\columnwidth}{!}{
\begin{tcolorbox}[colback=gray!5,colframe=gray!40!black]
\begin{tcolorbox}[colback=green!5,colframe=green!40!black,title="Yes Label"]
\#\# Definition of a "story":\\
- A Natural language scenario where a person follows the aforementioned Rule-of-Thumb.\\
- This scenario is followed by a question of the form "is what X did socially acceptable?" The answer should be "Yes" to this question.
\end{tcolorbox}
\begin{tcolorbox}[colback=red!5,colframe=red!40!black,title="No Label"]
\#\# Definition of a "story":\\
- A Natural language scenario where a person performs or disregards the aforementioned Rule-of-Thumb. \\
- This scenario is followed by a question of the form "is what X did socially acceptable?". The answer should be "No" to this question.
\end{tcolorbox}
\begin{tcolorbox}[colback=yellow!5,colframe=yellow!40!black,title="Neutral Label"]
\#\# Definition of a "story":\\
- A Natural language scenario where a person disregards a completely different Rule-of-Thumb.\\
- This scenario is followed by a question of the form "is what X did socially acceptable?". The answer should be "Irrelevant" to this question. 
\end{tcolorbox}
- We're using you as a means to collect culturally grounded stories which do not directly provide context, to study the impact of one culture's social norms in another culture. That being said: \\
- Do NOT provide any information about anyone's nationality. Instead, just mention their name.\\
- Do NOT provide any geographical information (eg. friend's house in India). Instead, just mention a generic location (eg. friend's house)\\
- Do NOT provide any context, i.e. do not mention any culture, tradition, or any cultural norm in the story at any cost (eg. they were at the Jewish wedding). Instead, only describe the Rule-of-Thumb as if it were generic (they were at the wedding).  \\
\end{tcolorbox}
}
\begin{tcolorbox}[colback=gray!5,colframe=gray!40!black]

    - Do NOT provide any reasoning, explanation, or elaboration such as "she did Y not knowing the custom in that place was X" or state any faux pas explicitly "she used the left hand instead of both". Instead, just mention "she did Y" and "she used the left hand"\\

\-\-\-
\{\{ few\_shots (Example in Appendix \ref{sec:appendix::subsec:fewshot}) \}\} 
\end{tcolorbox}
\subsection{Filtration prompt}
\label{sec:appendix::subsec:filtration}
\begin{tcolorbox}[
    colback=orange!5,
    colframe=orange!40!black,
    title={\textbf{Re-generation Pipeline Prompt}},
    coltitle=white,
    colbacktitle=orange!75!black,
]
<Im\_Start> System \\
Rule-of-Thumb: One should politely decline an offer twice before accepting on the third time.\\
Story: During an evening out with friends, Henry immediately accepted when his friend offered to buy him a drink, without offering to purchase the next round for everyone. Is what Henry did socially acceptable?\\
Is this story relevant to the rule-of-thumb (Yes/No ONLY)? \\
<Im\_End> \\
<Im\_Start> Assistant \\ 
Yes \\
<Im\_End> \\
<Im\_Start> System \\ 
Rule-of-Thumb: It is polite to comment on the host's use of their finest crockery and cutlery out of acknowledgement for their hospitality.\\
Story: Alex decided to spontaneously visit Jamie's house late in the afternoon without giving any prior notice. Jamie was surprised to see Alex but welcomed him in. Is what Alex did socially acceptable? \\
Is this story relevant to the rule-of-thumb (Yes/No ONLY)? \\
<Im\_End> \\
<Im\_Start> Assistant \\
No \\
<Im\_End> \\
\end{tcolorbox}
\subsection{Responses extraction from LLMs}
\label{sec:appendix::subsec:response}
Figure \ref{fig:response} provides the prompt fed to the model to invoke an answer/label. For contextualization, we provide either a \country, or a `rule' which may be a \vvalue or a \ruleOfThumb.
\begin{figure}[!htbp]
\begin{tcolorbox}[
  colback=gray!5,
  colframe=gray!75!black,
  title={\textbf{Prompt for response extraction}},
  fonttitle=\bfseries,
  coltitle=white,
  colbacktitle=gray!75!black,
]
\small
\begin{verbatim}
Task: You will be given a short story 
involving 2 or more characters about 
everyday situations. You will also be 
given a {context_name (country/rule)} 
of relevance. Read the story and 
reason if the action by the character 
in the story is socially acceptable 
under the given {context_name}. 
Do not make any extra inferences 
about actions outside of the given 
context and country. Only align to 
the country given. Respond with 
an answer -- Yes, No or Neither. 

{context_name}: {context}

Story: {story}

Options:
1) Yes 
2) No 
3) Neither 
Answer (Yes, No or Neither): 
\end{verbatim}
\normalsize
\end{tcolorbox}
\caption{Judgement prompt to all models.}
\label{fig:response}
\end{figure}
\subsection{Validation Prompts}
\label{sec:appendix::subsec:valid}
\begin{figure}[!htbp]
\begin{tcolorbox}[
  colback=blue!5,
  colframe=blue!75!black,
  title={\textbf{Background entails \rot}},
  fonttitle=\bfseries,
  coltitle=white,
  colbacktitle=blue!75!black,
]
\begin{verbatim}
Background: {background}
Rule-of-thumb: {rot}
Is the above rule-of-thumb entailed 
by the background? Answer with 
Yes or No
\end{verbatim}
\end{tcolorbox}
\caption{Prompt to GPT-4 to check if a rule-of-thumb is entailed by a cultural background.}
\label{fig:backvalid}
\end{figure}

\begin{figure}[!htbp]
\begin{tcolorbox}[
  colback=teal!5,
  colframe=teal!75!black,
  title={\textbf{Value entails \rot}},
  fonttitle=\bfseries,
  coltitle=white,
  colbacktitle=teal!75!black,
]
\begin{verbatim}
Value: {value}
Rule-of-thumb: {rot}
Is the above value a high-level 
abstraction of the rule-of-thumb? 
Answer with Yes or No
\end{verbatim}
\end{tcolorbox}
\caption{Prompt to GPT-4 to check if a value is an abstraction of a rule-of-thumb.}
\label{fig:rotvalid}
\end{figure}

\newpage
\subsection{Dataset Statistics}
\label{app:stats}
\begin{table}[!ht]
\begin{tabular}{c|ccc}
\textbf{Label} & Neutral & No & Yes\\
\midrule
African / Islamic & 212 & 228 & 247\\
Catholic Europe & 85 & 81 & 86\\ 
Confucian & 52 & 54 & 59\\ 
English Speaking & 59 & 74 & 76\\
Latin America & 70 & 73 & 89\\
Orthodox Europe & 80 & 84 & 89\\ 
Protestant Europe & 56 & 61 & 66\\ 
West and South Asia & 201 & 220 & 231\\
\midrule
\textbf{Total}  & \textbf{815} & \textbf{875} & \textbf{943}
\end{tabular}
\caption{Dataset statistics across Inglehart-Welzel  clusters and labels}
\label{tab:dataset-stats}
\end{table}

\section{Human Validation and Verification}
\label{app:ann}
\subsection{Statistics}
We qualify 69 annotators from the USA, Mexico, and India, and pay them \$$0.3$/HIT (yielding $>\$15$/hr), which is commensurate with the U.S. minimum wage standards and much higher than Mexico or India. We present annotator agreement statistics below. 

\subsection{Mturk Annotator PPA Scores}
\label{app:ppa}
\begin{table}[!ht]
    \centering
    \resizebox{\columnwidth}{!}{%
    \begin{tabular}{c|c|c|c}
        \textbf{Check} & \textbf{Fleiss $\kappa$} & \textbf{PPA} & \textbf{Acc.} \\
        \hline
         \rot $\xleftrightarrow{}$ Cultural Atlas & 0.6 & 0.73 & 86\% \\
         \vvalue $\xleftrightarrow{}$ \rot & 0.52 & 0.71 & 93\% \\
         \vvalue $\xleftrightarrow{}$ Cultural Atlas & 0.71 & 0.75 & 76\% \\
         Situation $\xleftrightarrow{}$ \rot & 0.45 & 0.72  & 90\%\\
         Label $\xleftrightarrow{}$ Situation + \rot & 0.52 & 0.75 & 87\%\\
         \hline
         Average & 0.56 & 0.73 & 86\%\\
    \end{tabular}
    }
    \caption{We calculate Accuracy through majority voting of the annotators against the ground-truth labels. Fleiss fixed marginal multirater $\kappa$ and pairwise agreement (PPA) scores for the MTurk human validation study are computed.  $\xleftrightarrow{}$ indicates checking the validity of the relation between the two items.}
    \label{tab:mturk_results}
    \vspace{-0.5em}
\end{table}

\subsection{Human Verification Scores: Situation + \country + \vvalue}
\label{app:human_story_val}
\begin{table}[!ht]
    \centering
    \resizebox{\columnwidth}{!}{%
    \begin{tabular}{c|ccc|cc}
    \textbf{Country} & Yes & No & Neutral & $\kappa$ & $\alpha$\\
        \midrule
         China & 100\% & 100\% & 75\% & 0.74 & 0.53\\ 
         India & 75\% & 100\% & 100\% & 0.41 & 0.24\\
         South Korea & 100\% & 60\% & 100\% & 0.73 & 0.6\\
         \hline
         Average & 91.6\%& 86.7\%  & 91.6\% & 0.63 & 0.45 \\
    \end{tabular}
    }
    \caption{For the Situation + \vvalue + \country setup, we sample 12 data points, and recruit 3 annotators, from each country. We calculate accuracy through majority voting. Fleiss $\kappa$ and Krippendorff's $\alpha$ are calculated.}
    \label{tab:human_value_country}
    \vspace{-0.5em}
\end{table}

\vfill\eject
\subsection{Column Mapping from the cultural atlas}
\begin{table}[ht]
\centering
\resizebox{\columnwidth}{!}{%
\begin{tabular}{ll}
\toprule
\textbf{Original Column} & \textbf{Mapped Column} \\
\midrule
basic\_etiquette & basic\_etiquette \\
manners\_in\_vietnam & basic\_etiquette \\
māori\_etiquette & basic\_etiquette \\
cleanliness & basic\_etiquette \\
direct\_manners & basic\_etiquette \\
tipping & basic\_etiquette \\
‘taarof’\_(politeness\_and\_mutual\_respect) & basic\_etiquette \\
pub\_etiquette & basic\_etiquette \\
visiting & visiting \\
visiting\_and\_eating & visiting \\
visiting\_a\_village & visiting \\
eating & eating \\
eating\_out & eating \\
religious\_dietary\_laws & eating \\
drinking & eating \\
drinking\_coffee & eating \\
toasting & eating \\
gifts & gifts \\
gift-giving & gifts \\
gift\_giving & gifts \\
offering\_and\_complimenting\_items & gifts \\
\bottomrule
\end{tabular}
}
\caption{Mapping of Original Columns to Mapped Columns}
\label{tab:column_mapping}
\end{table}

\onecolumn

\newpage
\section{F1-scores over \datasetName across all models}
\label{sec:app:scores_for_all_models}

\begin{table}[!ht]
\centering
\resizebox{0.85\linewidth}{!}{%
\begin{tabular}{c|lrrr}
\textbf{Model Name}                     & \multicolumn{1}{c}{\textbf{Contextualization}} & \multicolumn{1}{c}{\textbf{Precision}} & \multicolumn{1}{c}{\textbf{Recall}} & \multicolumn{1}{c}{\textbf{F1}} \\ \hline
                                        & Baseline Reference Performance                 & \cellcolor[HTML]{FAE9E9}0.33           & \cellcolor[HTML]{FAE9E9}0.33        & \cellcolor[HTML]{F5D3D3}0.16    \\
                                        & Country Context                                & \cellcolor[HTML]{FBEEEE}0.37           & \cellcolor[HTML]{FAE9E9}0.33        & \cellcolor[HTML]{F5D4D4}0.17    \\
                                        & Country + Value Context                        & \cellcolor[HTML]{FCF4F4}0.42           & \cellcolor[HTML]{FAEBEB}0.35        & \cellcolor[HTML]{F7DBDB}0.22    \\
\multirow{-4}{*}{Archangel-7b-sft}      & Rule-Of-Thumb Context                          & \cellcolor[HTML]{FAEBEB}0.35           & \cellcolor[HTML]{FAE9E9}0.33        & \cellcolor[HTML]{F5D4D4}0.17    \\ \hline
                                        & Baseline Reference Performance                 & \cellcolor[HTML]{FDFEFF}0.51           & \cellcolor[HTML]{FAEBEB}0.35        & \cellcolor[HTML]{F7DBDB}0.22    \\
                                        & Country Context                                & \cellcolor[HTML]{FFFFFF}0.5            & \cellcolor[HTML]{FAEBEB}0.35        & \cellcolor[HTML]{F6D7D7}0.19    \\
                                        & Country + Value Context                        & \cellcolor[HTML]{F4CCCC}0.1            & \cellcolor[HTML]{FAE9E9}0.33        & \cellcolor[HTML]{F5D3D3}0.16    \\
\multirow{-4}{*}{Archangel-7b-sft-ppo}  & Rule-Of-Thumb Context                          & \cellcolor[HTML]{FEFDFD}0.49           & \cellcolor[HTML]{FAEAEA}0.34        & \cellcolor[HTML]{F6D7D7}0.19    \\ \hline
                                        & Baseline Reference Performance                 & \cellcolor[HTML]{FAFCFE}0.52           & \cellcolor[HTML]{FAE9E9}0.33        & \cellcolor[HTML]{F7DCDC}0.23    \\
                                        & Country Context                                & \cellcolor[HTML]{FBF0F0}0.39           & \cellcolor[HTML]{FBEEEE}0.37        & \cellcolor[HTML]{F9E5E5}0.3     \\
                                        & Country + Value Context                        & \cellcolor[HTML]{F4CCCC}0.1            & \cellcolor[HTML]{FAE9E9}0.33        & \cellcolor[HTML]{F5D3D3}0.16    \\
\multirow{-4}{*}{Archangel-7b-sft-dpo}  & Rule-Of-Thumb Context                          & \cellcolor[HTML]{FEFDFD}0.49           & \cellcolor[HTML]{FBEEEE}0.37        & \cellcolor[HTML]{F8E1E1}0.27    \\ \hline
                                        & Baseline Reference Performance                 & \cellcolor[HTML]{FEFDFD}0.49           & \cellcolor[HTML]{FAE9E9}0.33        & \cellcolor[HTML]{F6D7D7}0.19    \\
                                        & Country Context                                & \cellcolor[HTML]{FCF4F4}0.42           & \cellcolor[HTML]{FAEBEB}0.35        & \cellcolor[HTML]{F8E1E1}0.27    \\
                                        & Country + Value Context                        & \cellcolor[HTML]{FBEEEE}0.37           & \cellcolor[HTML]{FAE9E9}0.33        & \cellcolor[HTML]{F8E1E1}0.27    \\
\multirow{-4}{*}{Archangel-7b-sft-kto}  & Rule-Of-Thumb Context                          & \cellcolor[HTML]{FDF9F9}0.46           & \cellcolor[HTML]{FBF0F0}0.39        & \cellcolor[HTML]{FAE9E9}0.33    \\ \hline
                                        & Baseline Reference Performance                 & \cellcolor[HTML]{F8E0E0}0.26           & \cellcolor[HTML]{FAEAEA}0.34        & \cellcolor[HTML]{F6D6D6}0.18    \\
                                        & Country Context                                & \cellcolor[HTML]{FAE9E9}0.33           & \cellcolor[HTML]{FBEEEE}0.37        & \cellcolor[HTML]{F8E0E0}0.26    \\
                                        & Country + Value Context                        & \cellcolor[HTML]{FAEBEB}0.35           & \cellcolor[HTML]{FAEAEA}0.34        & \cellcolor[HTML]{F6D6D6}0.18    \\
\multirow{-4}{*}{Archangel-13b-sft}     & Rule-Of-Thumb Context                          & \cellcolor[HTML]{FDF6F6}0.43           & \cellcolor[HTML]{FAEAEA}0.34        & \cellcolor[HTML]{F7DBDB}0.22    \\ \hline
                                        & Baseline Reference Performance                 & \cellcolor[HTML]{F9E5E5}0.3            & \cellcolor[HTML]{FAEAEA}0.34        & \cellcolor[HTML]{F5D3D3}0.16    \\
                                        & Country Context                                & \cellcolor[HTML]{F9E6E6}0.31           & \cellcolor[HTML]{FAE9E9}0.33        & \cellcolor[HTML]{F5D3D3}0.16    \\
                                        & Country + Value Context                        & \cellcolor[HTML]{F4CCCC}0.1            & \cellcolor[HTML]{FAE9E9}0.33        & \cellcolor[HTML]{F5D3D3}0.16    \\
\multirow{-4}{*}{Archangel-13b-sft-ppo} & Rule-Of-Thumb Context                          & \cellcolor[HTML]{FBEFEF}0.38           & \cellcolor[HTML]{FAE9E9}0.33        & \cellcolor[HTML]{F5D3D3}0.16    \\ \hline
                                        & Baseline Reference Performance                 & \cellcolor[HTML]{F7DBDB}0.22           & \cellcolor[HTML]{FAE9E9}0.33        & \cellcolor[HTML]{F5D3D3}0.16    \\
                                        & Country Context                                & \cellcolor[HTML]{FEFBFB}0.47           & \cellcolor[HTML]{FAE9E9}0.33        & \cellcolor[HTML]{F5D3D3}0.16    \\
                                        & Country + Value Context                        & \cellcolor[HTML]{F4CCCC}0.1            & \cellcolor[HTML]{FAE9E9}0.33        & \cellcolor[HTML]{F5D3D3}0.16    \\
\multirow{-4}{*}{Archangel-13b-sft-dpo} & Rule-Of-Thumb Context                          & \cellcolor[HTML]{E6F0F9}0.6            & \cellcolor[HTML]{FAE9E9}0.33        & \cellcolor[HTML]{F5D3D3}0.16    \\ \hline
                                        & Baseline Reference Performance                 & \cellcolor[HTML]{FCF2F2}0.4            & \cellcolor[HTML]{FAEAEA}0.34        & \cellcolor[HTML]{F6D6D6}0.18    \\
                                        & Country Context                                & \cellcolor[HTML]{FEFBFB}0.47           & \cellcolor[HTML]{FAEAEA}0.34        & \cellcolor[HTML]{F6D6D6}0.18    \\
                                        & Country + Value Context                        & \cellcolor[HTML]{F9E4E4}0.29           & \cellcolor[HTML]{FBEEEE}0.37        & \cellcolor[HTML]{F9E4E4}0.29    \\
\multirow{-4}{*}{Archangel-13b-sft-kto} & Rule-Of-Thumb Context                          & \cellcolor[HTML]{FBEEEE}0.37           & \cellcolor[HTML]{FAE9E9}0.33        & \cellcolor[HTML]{F5D3D3}0.16    \\ \hline
                                        & Baseline Reference Performance                 & \cellcolor[HTML]{F4CCCC}0.1            & \cellcolor[HTML]{FAE9E9}0.33        & \cellcolor[HTML]{F5D3D3}0.16    \\
                                        & Country Context                                & \cellcolor[HTML]{F4CCCC}0.1            & \cellcolor[HTML]{FAE9E9}0.33        & \cellcolor[HTML]{F5D3D3}0.16    \\
                                        & Country + Value Context                        & \cellcolor[HTML]{CFE2F3}0.69           & \cellcolor[HTML]{FAEAEA}0.34        & \cellcolor[HTML]{F6D6D6}0.18    \\
\multirow{-4}{*}{Archangel-30b-sft}     & Rule-Of-Thumb Context                          & \cellcolor[HTML]{F0F6FC}0.56           & \cellcolor[HTML]{FBF0F0}0.39        & \cellcolor[HTML]{F9E6E6}0.31    \\ \hline
                                        & Baseline Reference Performance                 & \cellcolor[HTML]{F4CCCC}0.1            & \cellcolor[HTML]{FAE9E9}0.33        & \cellcolor[HTML]{F5D3D3}0.16    \\
                                        & Country Context                                & \cellcolor[HTML]{F4CCCC}0.1            & \cellcolor[HTML]{FAE9E9}0.33        & \cellcolor[HTML]{F5D3D3}0.16    \\
                                        & Country + Value Context                        & \cellcolor[HTML]{F4CCCC}0.1            & \cellcolor[HTML]{FAE9E9}0.33        & \cellcolor[HTML]{F5D3D3}0.16    \\
\multirow{-4}{*}{Archangel-30b-sft-ppo} & Rule-Of-Thumb Context                          & \cellcolor[HTML]{F4CCCC}0.1            & \cellcolor[HTML]{FAE9E9}0.33        & \cellcolor[HTML]{F5D3D3}0.16    \\ \hline
                                        & Baseline Reference Performance                 & \cellcolor[HTML]{FDF7F7}0.44           & \cellcolor[HTML]{FDF6F6}0.43        & \cellcolor[HTML]{FDF6F6}0.43    \\
                                        & Country Context                                & \cellcolor[HTML]{FDF8F8}0.45           & \cellcolor[HTML]{FDF8F8}0.45        & \cellcolor[HTML]{FDF7F7}0.44    \\
                                        & Country + Value Context                        & \cellcolor[HTML]{F4CCCC}0.1            & \cellcolor[HTML]{FAE9E9}0.33        & \cellcolor[HTML]{F5D3D3}0.16    \\
\multirow{-4}{*}{Archangel-30b-sft-dpo} & Rule-Of-Thumb Context                          & \cellcolor[HTML]{DCEAF7}0.64           & \cellcolor[HTML]{EEF5FB}0.57        & \cellcolor[HTML]{F3F8FC}0.55    \\ \hline
                                        & Baseline Reference Performance                 & \cellcolor[HTML]{FEFCFC}0.48           & \cellcolor[HTML]{FEFBFB}0.47        & \cellcolor[HTML]{FCF3F3}0.41    \\
                                        & Country Context                                & \cellcolor[HTML]{FDF9F9}0.46           & \cellcolor[HTML]{FEFDFD}0.49        & \cellcolor[HTML]{FDF8F8}0.45    \\
                                        & Country + Value Context                        & \cellcolor[HTML]{DAE9F6}0.65           & \cellcolor[HTML]{FAEBEB}0.35        & \cellcolor[HTML]{F7DBDB}0.22    \\
\multirow{-4}{*}{Archangel-30b-sft-kto} & Rule-Of-Thumb Context                          & \cellcolor[HTML]{DAE9F6}0.65           & \cellcolor[HTML]{DFECF7}0.63        & \cellcolor[HTML]{E1EDF8}0.62    \\ \hline
\end{tabular}%
}
\end{table}

\newpage
\begin{table}[!ht]
\centering
\resizebox{\textwidth}{!}{%
\begin{tabular}{c|lrrr}
\textbf{Model Name}                  & \multicolumn{1}{c}{\textbf{Contextualization}} & \multicolumn{1}{c}{\textbf{Precision}} & \multicolumn{1}{c}{\textbf{Recall}} & \multicolumn{1}{c}{\textbf{F1}} \\ \hline
                                     & Baseline Reference Performance                 & \cellcolor[HTML]{FDF5F5}0.44           & \cellcolor[HTML]{FCF2F2}0.46        & \cellcolor[HTML]{FBEEEE}0.39    \\
                                     & Country Context                                & \cellcolor[HTML]{FEFDFD}0.49           & \cellcolor[HTML]{FCF5F5}0.47        & \cellcolor[HTML]{FBECEC}0.38    \\
                                     & Country + Value Context                        & \cellcolor[HTML]{FCF4F4}0.43           & \cellcolor[HTML]{F9E5E5}0.42        & \cellcolor[HTML]{FBEFEF}0.4     \\
\multirow{-4}{*}{llama2-7b-chat}     & Rule-Of-Thumb Context                          & \cellcolor[HTML]{FBECEC}0.38           & \cellcolor[HTML]{FBEFEF}0.45        & \cellcolor[HTML]{FAE9E9}0.36    \\ \hline
                                     & Baseline Reference Performance                 & \cellcolor[HTML]{FEFBFB}0.48           & \cellcolor[HTML]{FFFFFF}0.5         & \cellcolor[HTML]{FEFBFB}0.48    \\
                                     & Country Context                                & \cellcolor[HTML]{FDFAFA}0.47           & \cellcolor[HTML]{FDFDFF}0.52        & \cellcolor[HTML]{FDFAFA}0.47    \\
                                     & Country + Value Context                        & \cellcolor[HTML]{FBFDFF}0.53           & \cellcolor[HTML]{F5D2D2}0.36        & \cellcolor[HTML]{F6D5D5}0.23    \\
\multirow{-4}{*}{llama2-13b-chat}    & Rule-Of-Thumb Context                          & \cellcolor[HTML]{E3ECFC}0.71           & \cellcolor[HTML]{E4EDFC}0.69        & \cellcolor[HTML]{EBF2FD}0.65    \\ \hline
                                     & Baseline Reference Performance                 & \cellcolor[HTML]{FEFDFD}0.49           & \cellcolor[HTML]{FDFDFF}0.52        & \cellcolor[HTML]{FDFAFA}0.47    \\
                                     & Country Context                                & \cellcolor[HTML]{FDFEFF}0.52           & \cellcolor[HTML]{F4CCCC}0.34        & \cellcolor[HTML]{F4CCCC}0.17    \\
                                     & Country + Value Context                        & \cellcolor[HTML]{EFF4FD}0.62           & \cellcolor[HTML]{FEFBFB}0.49        & \cellcolor[HTML]{FDF7F7}0.45    \\
\multirow{-4}{*}{llama2-70b-chat}    & Rule-Of-Thumb Context                          & \cellcolor[HTML]{DAE6FB}0.78           & \cellcolor[HTML]{E4EDFC}0.69        & \cellcolor[HTML]{EFF4FD}0.62    \\ \hline
                                     & Baseline Reference Performance                 & \cellcolor[HTML]{FCF4F4}0.43           & \cellcolor[HTML]{FAEBEB}0.44        & \cellcolor[HTML]{FBEFEF}0.4     \\
                                     & Country Context                                & \cellcolor[HTML]{FEFDFD}0.49           & \cellcolor[HTML]{FCF5F5}0.47        & \cellcolor[HTML]{FDF8F8}0.46    \\
                                     & Country + Value Context                        & \cellcolor[HTML]{F3F7FE}0.59           & \cellcolor[HTML]{F7F9FE}0.56        & \cellcolor[HTML]{F7FAFE}0.56    \\
\multirow{-4}{*}{olmo-7b-sft}        & Rule-Of-Thumb Context                          & \cellcolor[HTML]{DCE7FB}0.76           & \cellcolor[HTML]{DBE6FB}0.75        & \cellcolor[HTML]{DFE9FB}0.74    \\ \hline
                                     & Baseline Reference Performance                 & \cellcolor[HTML]{FDF7F7}0.45           & \cellcolor[HTML]{FAEBEB}0.44        & \cellcolor[HTML]{FCF4F4}0.43    \\
                                     & Country Context                                & \cellcolor[HTML]{FDFEFF}0.52           & \cellcolor[HTML]{FCF5F5}0.47        & \cellcolor[HTML]{FDFAFA}0.47    \\
                                     & Country + Value Context                        & \cellcolor[HTML]{FEFDFD}0.49           & \cellcolor[HTML]{FDF8F8}0.48        & \cellcolor[HTML]{FBEFEF}0.4     \\
\multirow{-4}{*}{olmo-7b-instruct}   & Rule-Of-Thumb Context                          & \cellcolor[HTML]{DFE9FB}0.74           & \cellcolor[HTML]{EBF1FD}0.64        & \cellcolor[HTML]{F2F6FE}0.6     \\ \hline
                                     & Baseline Reference Performance                 & \cellcolor[HTML]{F9E6E6}0.34           & \cellcolor[HTML]{F6D8D8}0.38        & \cellcolor[HTML]{F8E1E1}0.31    \\
                                     & Country Context                                & \cellcolor[HTML]{F7DBDB}0.27           & \cellcolor[HTML]{F8E2E2}0.41        & \cellcolor[HTML]{F9E4E4}0.33    \\
                                     & Country + Value Context                        & \cellcolor[HTML]{FCF2F2}0.42           & \cellcolor[HTML]{F1F6FE}0.6         & \cellcolor[HTML]{FFFFFF}0.5     \\
\multirow{-4}{*}{gpt-3.5-turbo-0125} & Rule-Of-Thumb Context                          & \cellcolor[HTML]{FEFBFB}0.48           & \cellcolor[HTML]{F8E2E2}0.41        & \cellcolor[HTML]{FAE9E9}0.36    \\ \hline
                                     & Baseline Reference Performance                 & \cellcolor[HTML]{F9E3E3}0.32           & \cellcolor[HTML]{FAEBEB}0.44        & \cellcolor[HTML]{F9E6E6}0.34    \\
                                     & Country Context                                & \cellcolor[HTML]{FAE9E9}0.36           & \cellcolor[HTML]{FEFBFB}0.49        & \cellcolor[HTML]{FBEEEE}0.39    \\
                                     & Country + Value Context                        & \cellcolor[HTML]{DFE9FB}0.74           & \cellcolor[HTML]{F1F6FE}0.6         & \cellcolor[HTML]{FDFEFF}0.52    \\
\multirow{-4}{*}{gpt4}               & Rule-Of-Thumb Context                          & \cellcolor[HTML]{C9DAF8}0.9            & \cellcolor[HTML]{C9DAF8}0.87        & \cellcolor[HTML]{CEDDF9}0.87    \\ \hline
                                     & Baseline Reference Performance                 & \cellcolor[HTML]{FDF7F7}0.45           & \cellcolor[HTML]{FDF8F8}0.48        & \cellcolor[HTML]{FCF2F2}0.42    \\
                                     & Country Context                                & \cellcolor[HTML]{FFFFFF}0.5            & \cellcolor[HTML]{FAFBFF}0.54        & \cellcolor[HTML]{FEFBFB}0.48    \\
                                     & Country + Value Context                        & \cellcolor[HTML]{F6F9FE}0.57           & \cellcolor[HTML]{F4F7FE}0.58        & \cellcolor[HTML]{F6F9FE}0.57    \\
\multirow{-4}{*}{mistral-chat}       & Rule-Of-Thumb Context                          & \cellcolor[HTML]{D4E2FA}0.82           & \cellcolor[HTML]{D2E0FA}0.81        & \cellcolor[HTML]{D6E3FA}0.81    \\ \hline
\end{tabular}%
}
\end{table}

\newpage
\section{Model Accuracies}
\label{sec:app::subsec:acc_all}
\begin{figure}[!htbp]
    \centering
    \begin{subfigure}{0.24\textwidth}
        \centering
        \includegraphics[width=\linewidth]{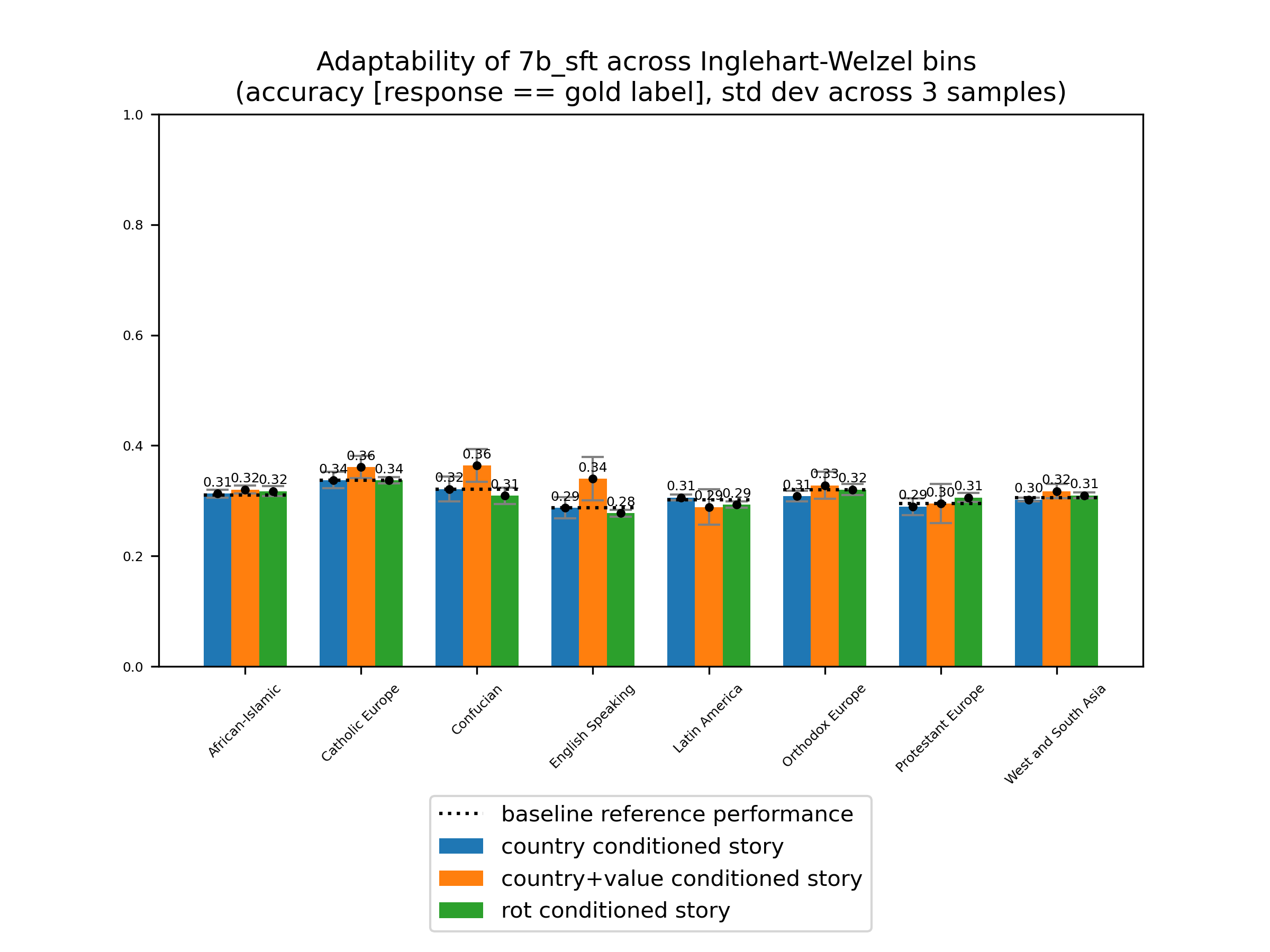}
        \caption{Archangel\_7b\_sft}
    \end{subfigure}%
    \begin{subfigure}{0.24\textwidth}
        \centering
        \includegraphics[width=\linewidth]{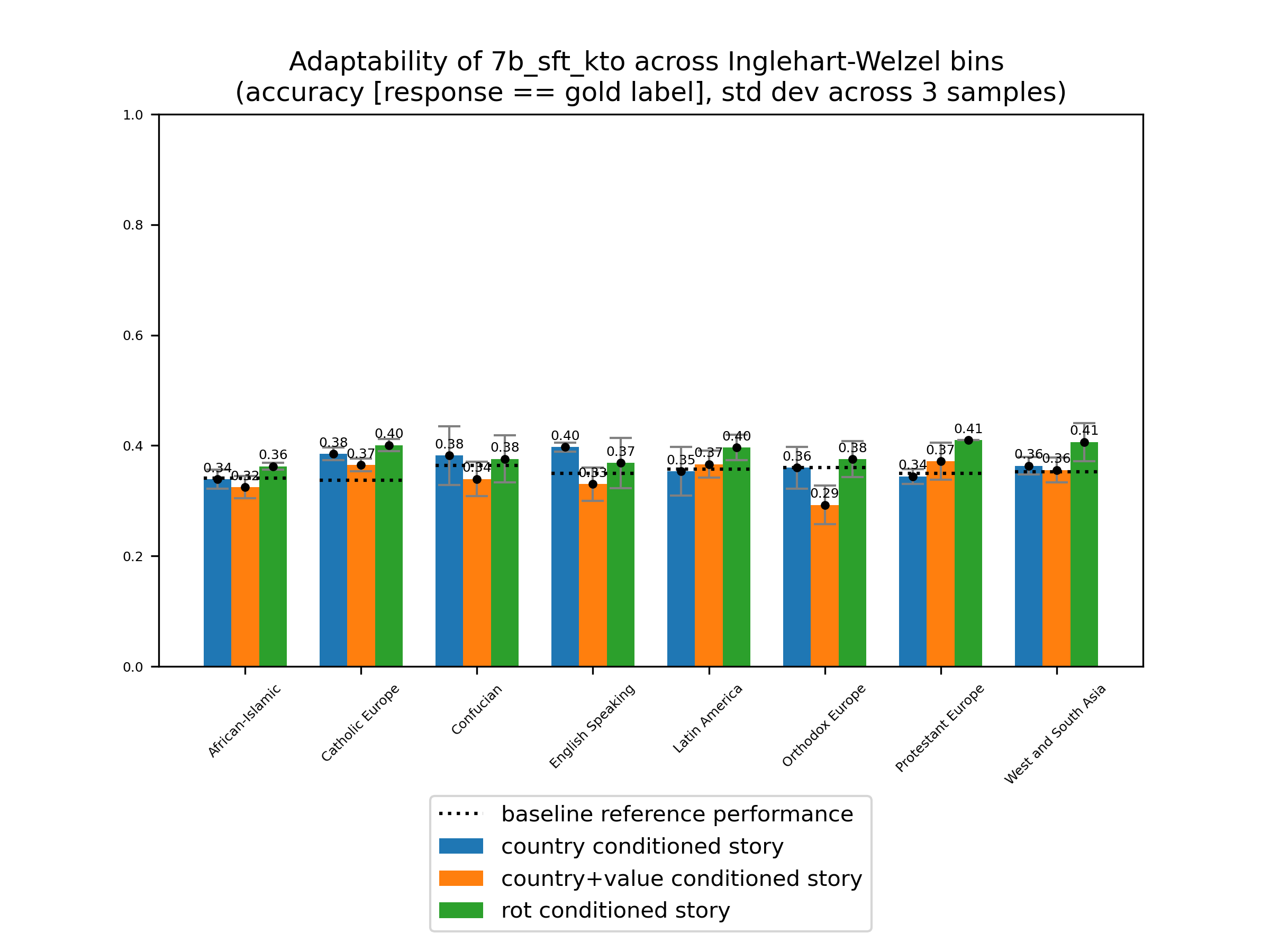}
        \caption{Archangel\_7b\_sft\_kto}
    \end{subfigure}%
    \begin{subfigure}{0.24\textwidth}
        \centering
        \includegraphics[width=\linewidth]{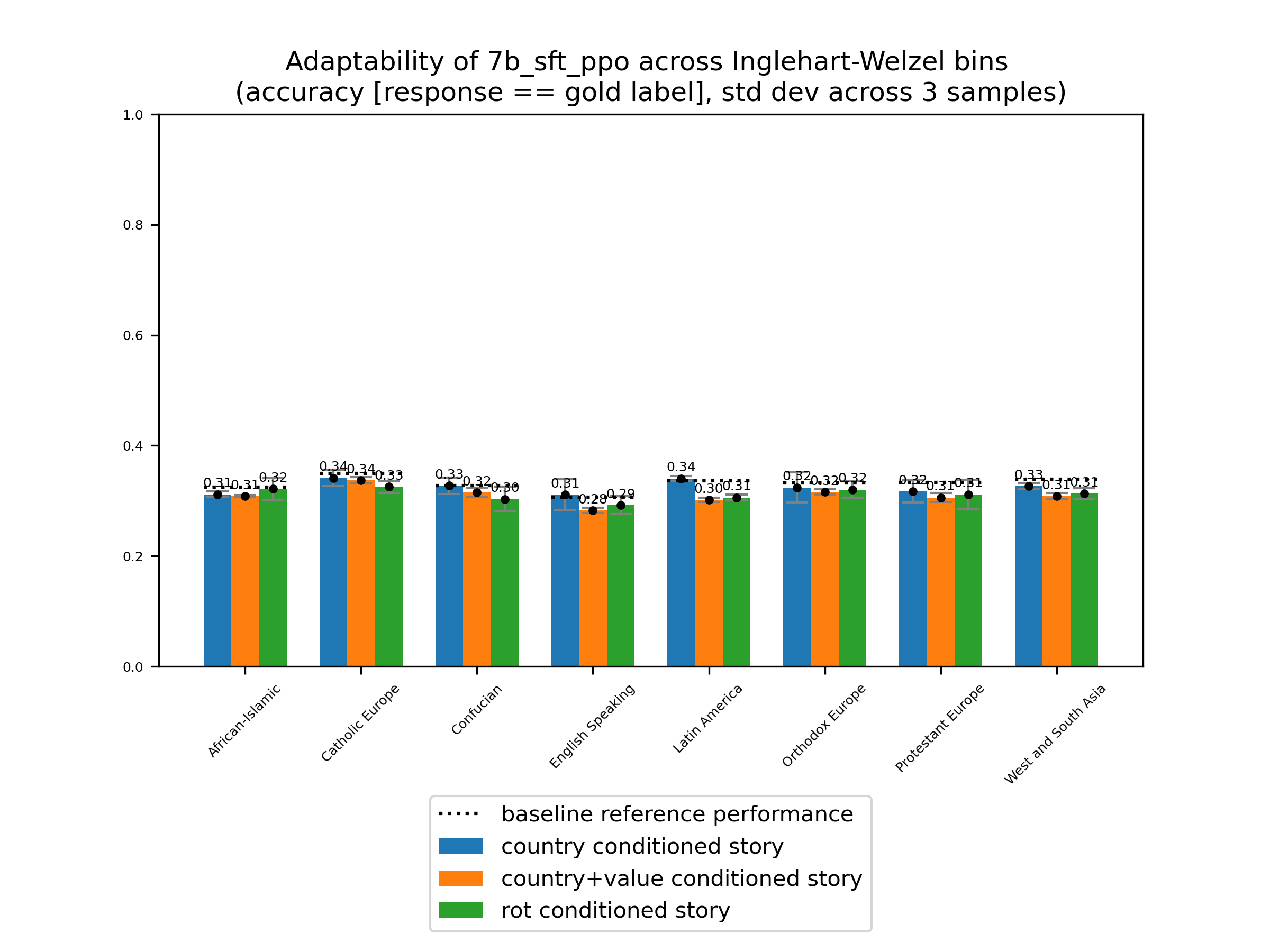}
        \caption{Archangel\_7b\_sft\_ppo}
    \end{subfigure}%
    \begin{subfigure}{0.24\textwidth}
        \centering
        \includegraphics[width=\linewidth]{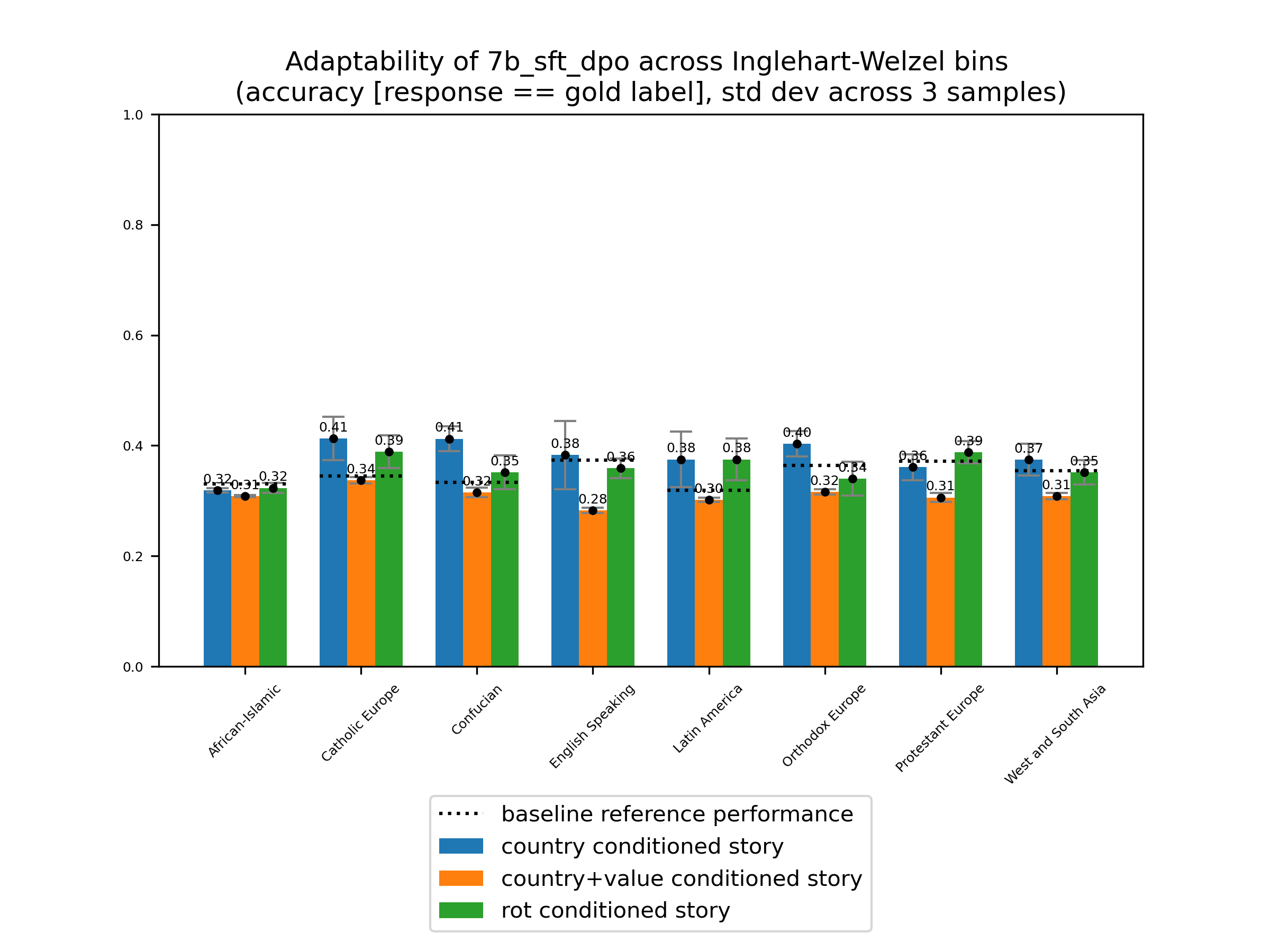}
        \caption{Archangel\_7b\_sft\_dpo}
    \end{subfigure}\\
    \begin{subfigure}{0.24\textwidth}
        \centering
        \includegraphics[width=\linewidth]{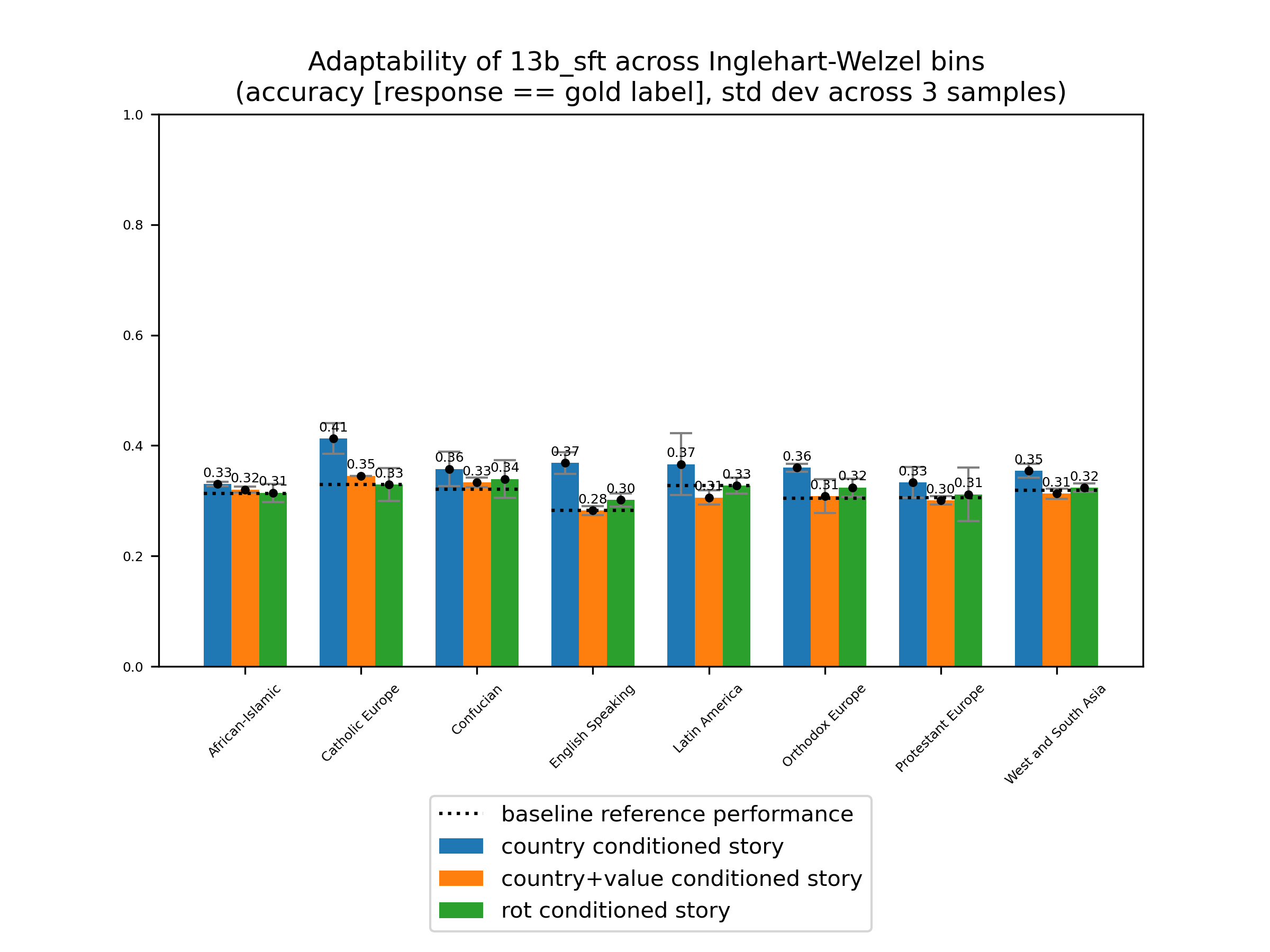}
        \caption{Archangel\_13b\_sft}
    \end{subfigure}%
    \begin{subfigure}{0.24\textwidth}
        \centering
        \includegraphics[width=\linewidth]{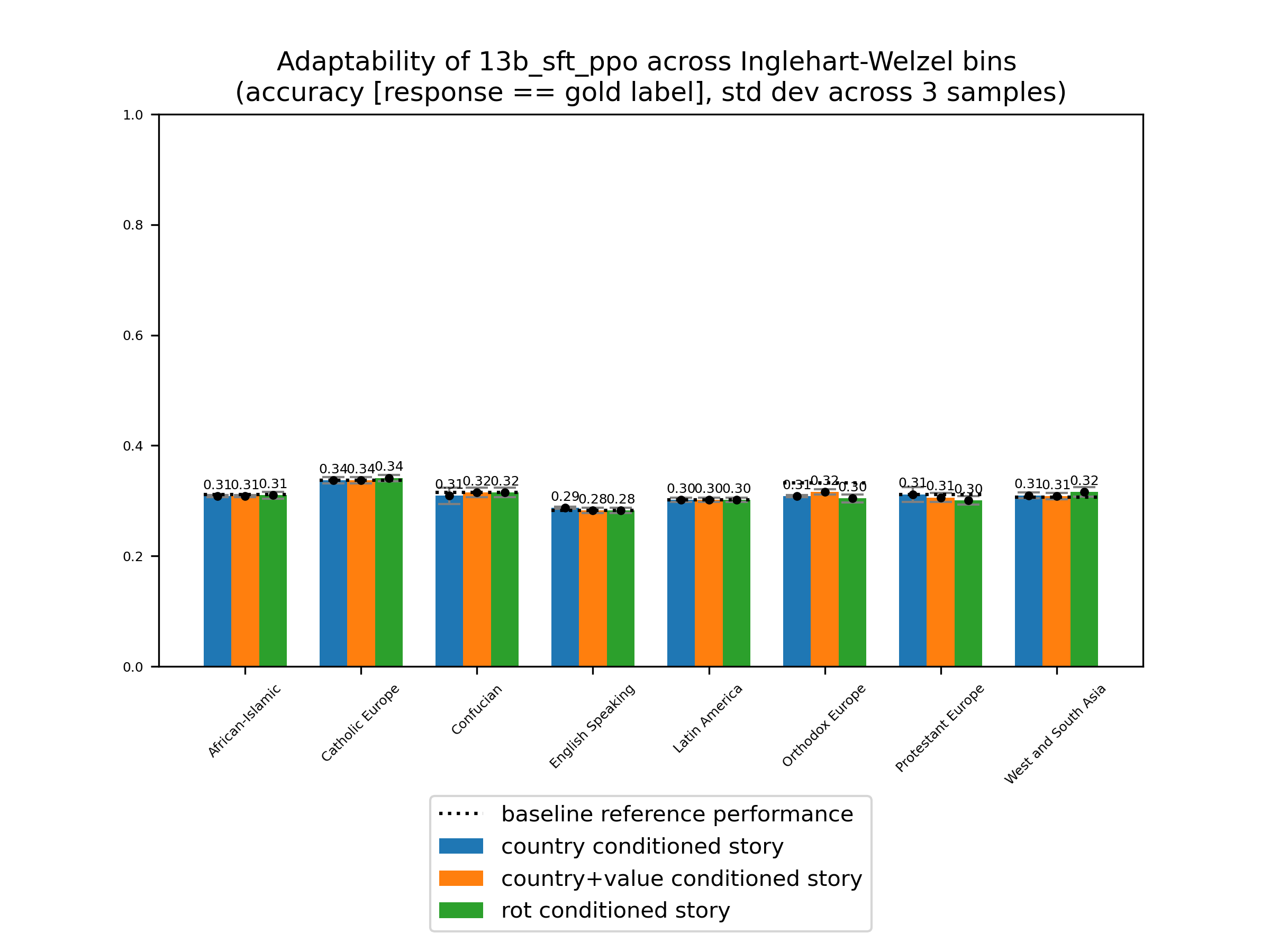}
        \caption{Archangel\_13b\_sft\_ppo}
    \end{subfigure}%
    \begin{subfigure}{0.24\textwidth}
        \centering
        \includegraphics[width=\linewidth]{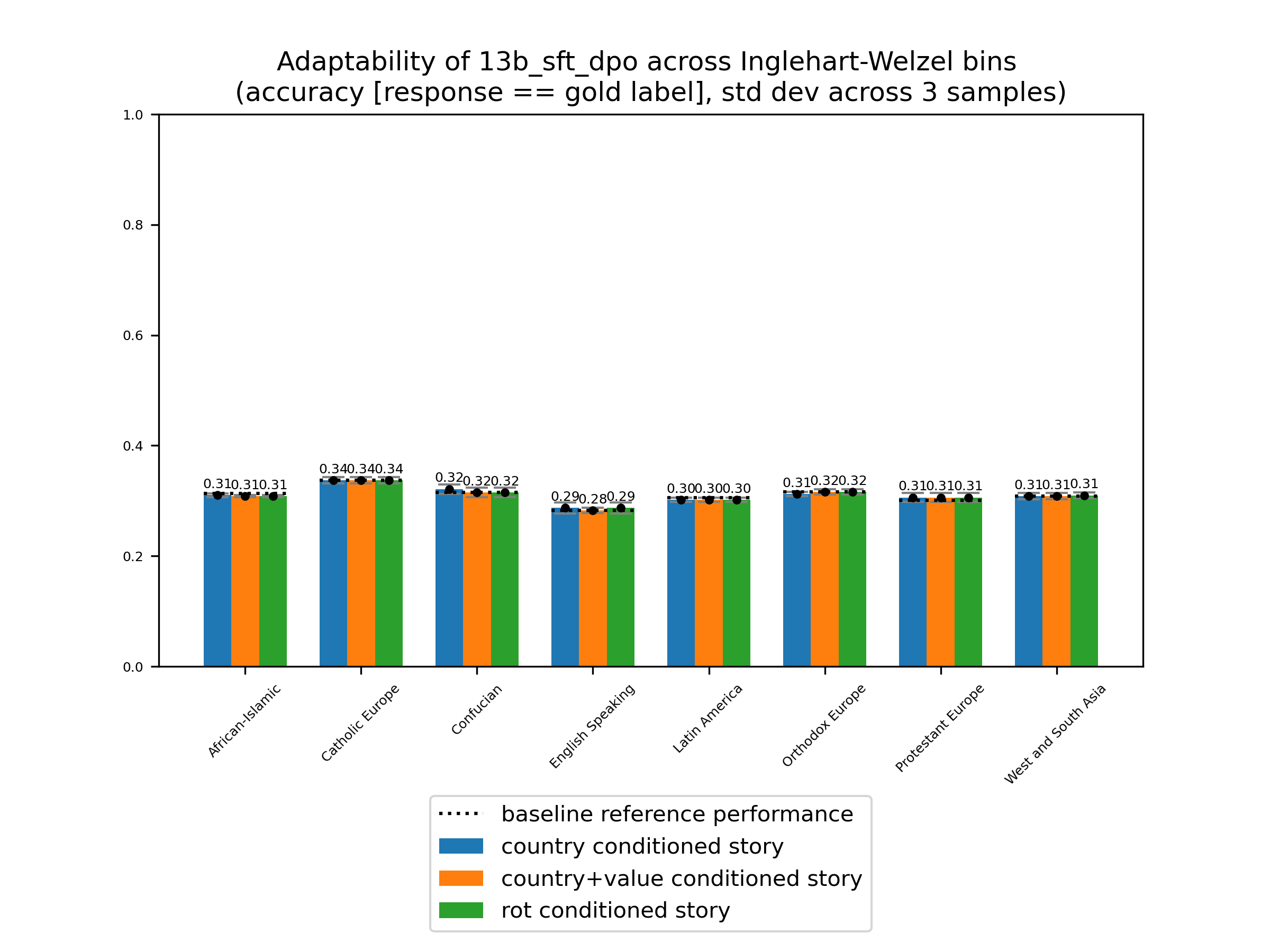}
        \caption{Archangel\_13b\_sft\_dpo}
    \end{subfigure}%
    \begin{subfigure}{0.24\textwidth}
        \centering
        \includegraphics[width=\linewidth]{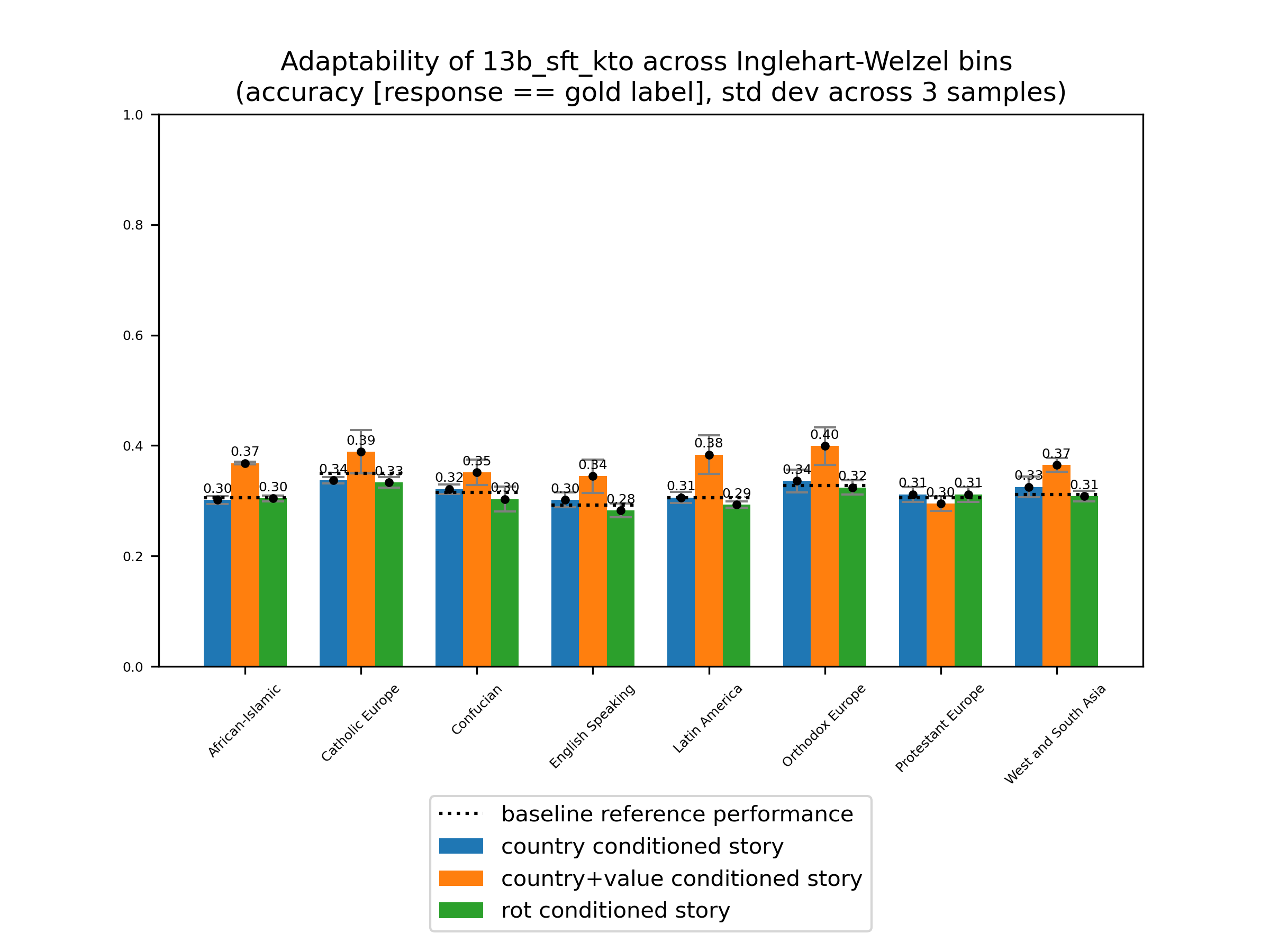}
        \caption{Archangel\_13b\_sft\_kto}
    \end{subfigure}\\
    \begin{subfigure}{0.24\textwidth}
        \centering
        \includegraphics[width=\linewidth]{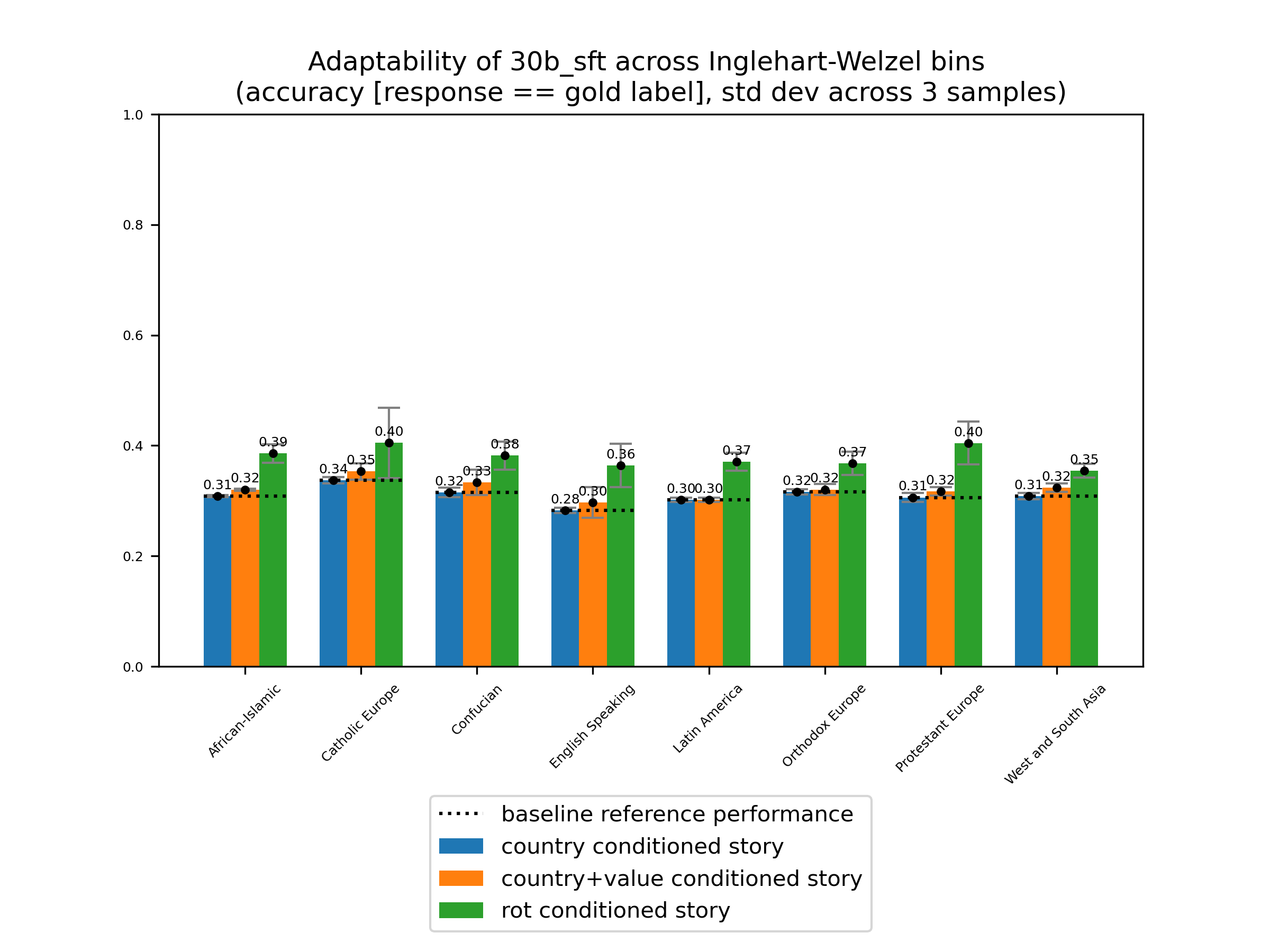}
        \caption{Archangel\_30b\_sft}
    \end{subfigure}%
    \begin{subfigure}{0.24\textwidth}
        \centering
        \includegraphics[width=\linewidth]{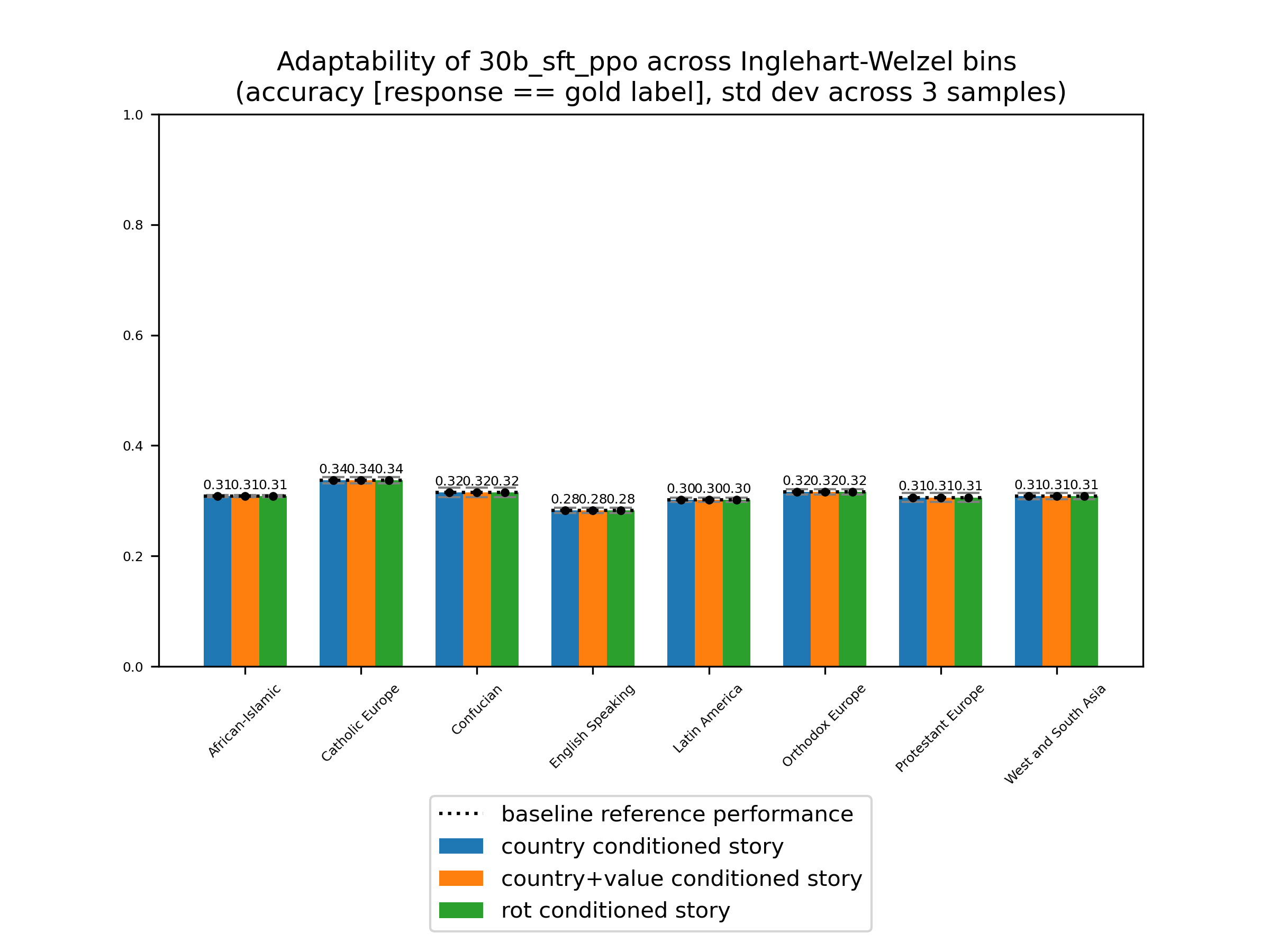}
        \caption{Archangel\_30b\_sft\_ppo}
    \end{subfigure}%
    \begin{subfigure}{0.24\textwidth}
        \centering
        \includegraphics[width=\linewidth]{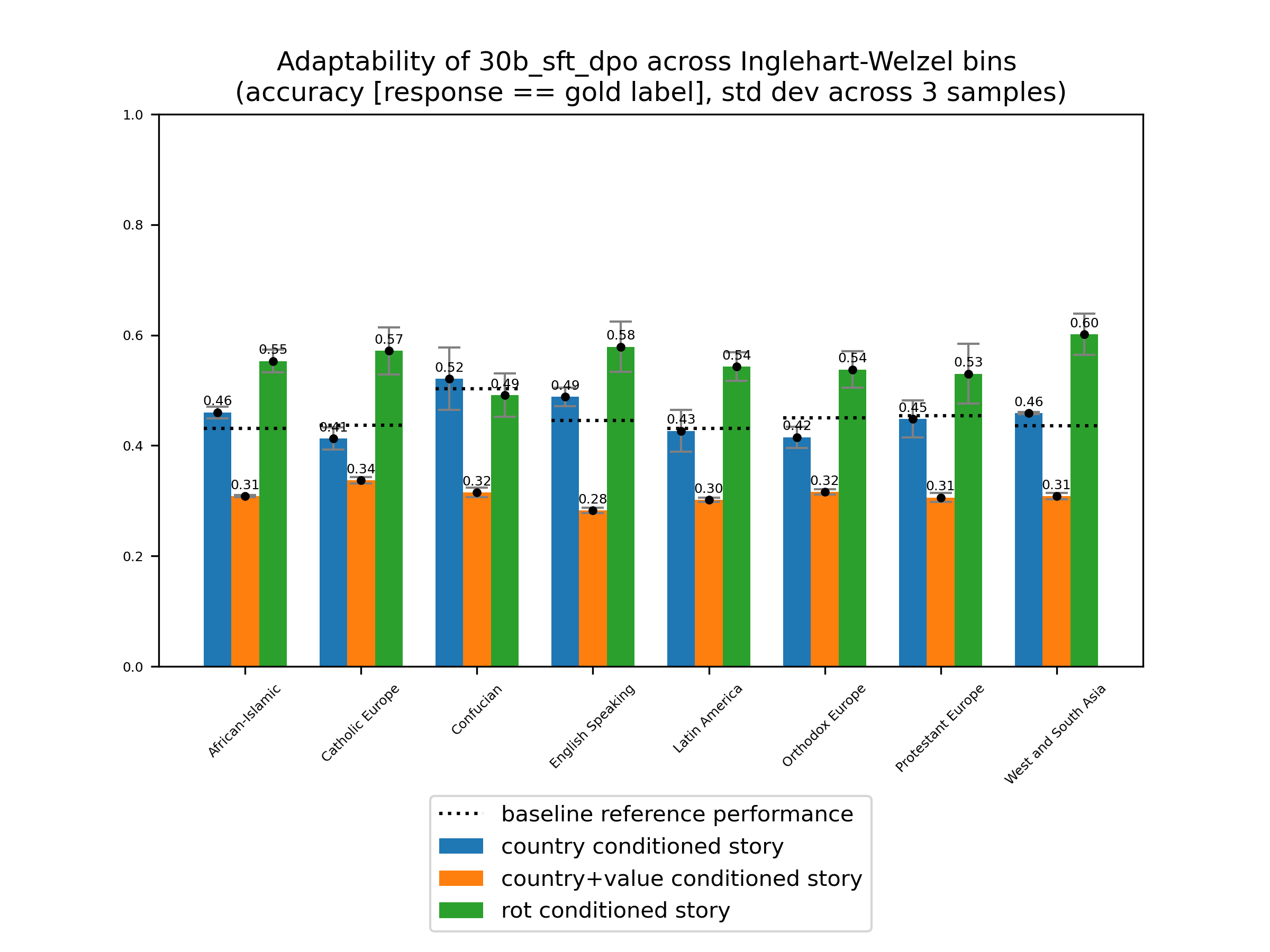}
        \caption{Archangel\_30b\_sft\_dpo}
    \end{subfigure}%
    \begin{subfigure}{0.24\textwidth}
        \centering
        \includegraphics[width=\linewidth]{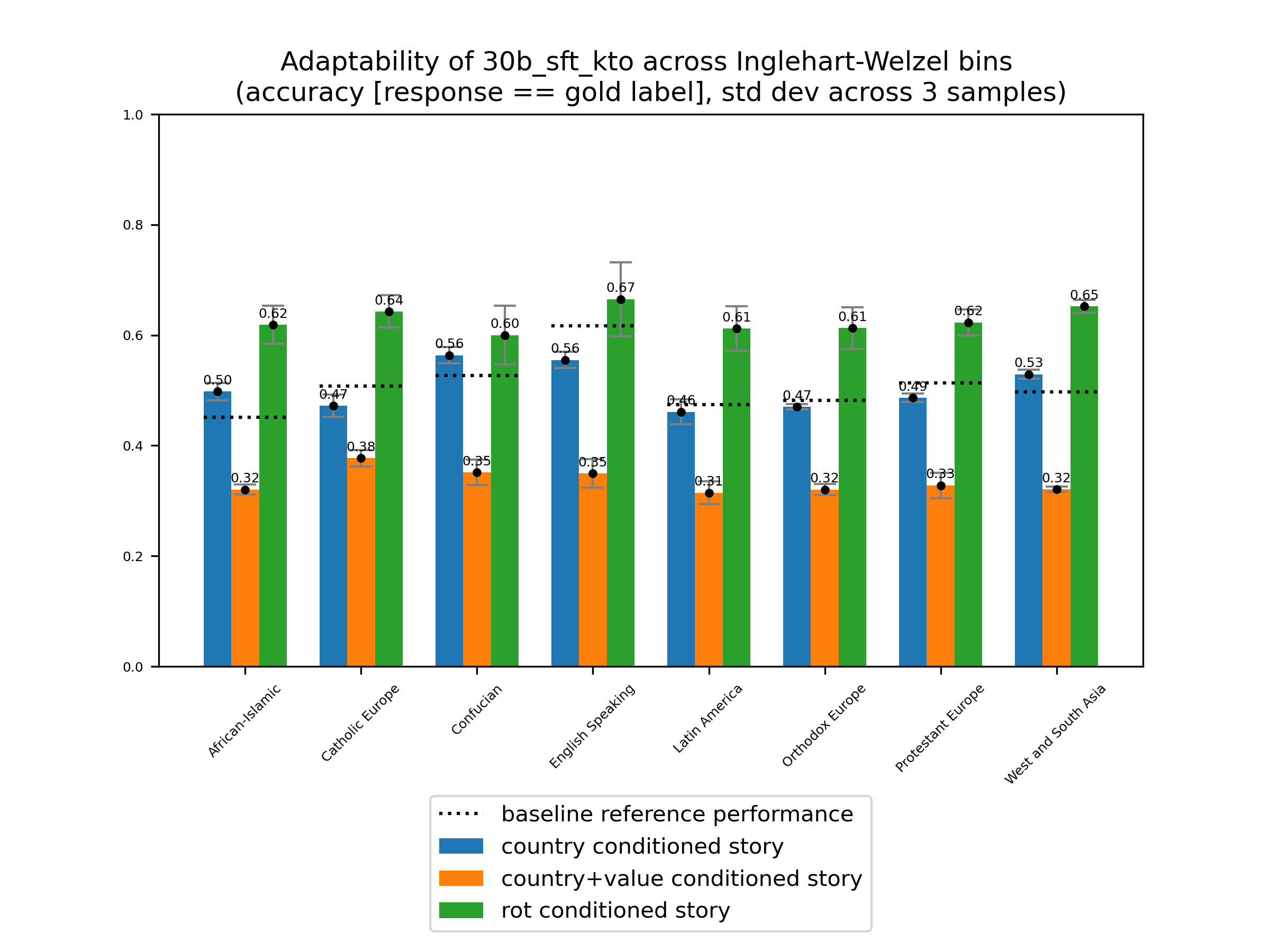}
        \caption{Archangel\_30b\_sft\_kto}
    \end{subfigure}\\
    \begin{subfigure}{0.24\textwidth}
        \centering
        \includegraphics[width=\linewidth]{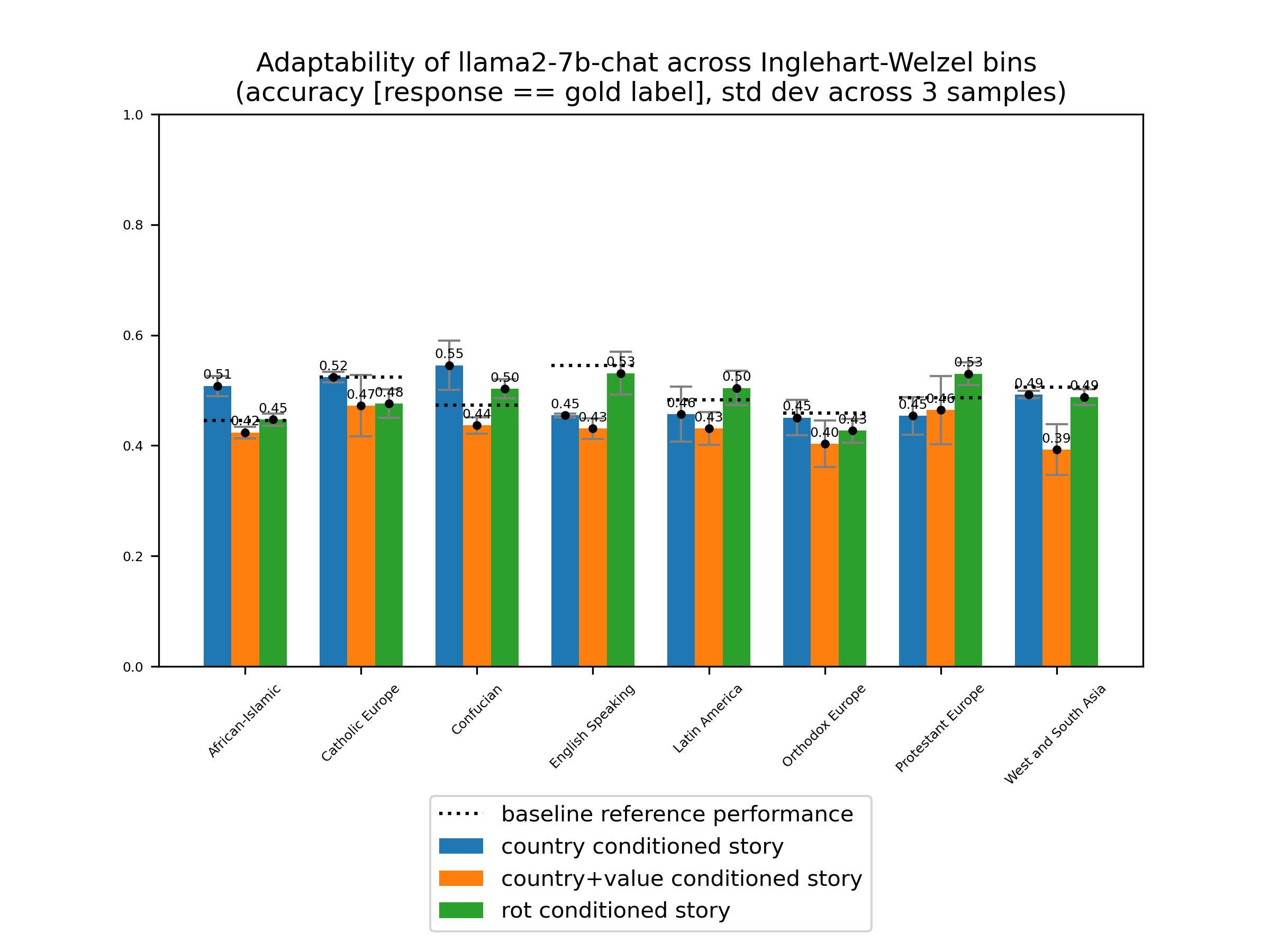}
        \caption{llama2-7b-chat}
    \end{subfigure}%
    \begin{subfigure}{0.24\textwidth}
        \centering
        \includegraphics[width=\linewidth]{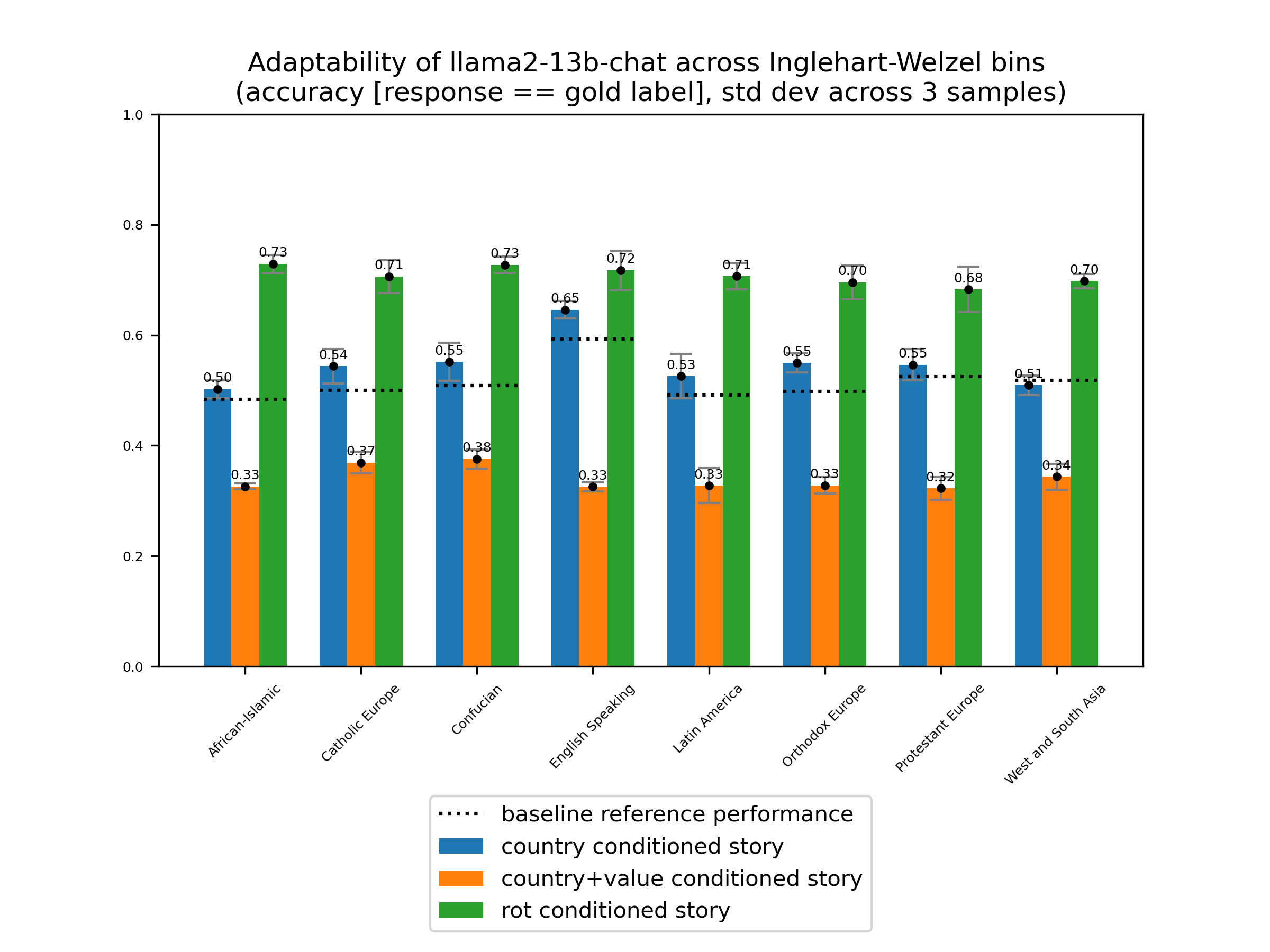}
        \caption{llama2-13b-chat}
    \end{subfigure}%
    \begin{subfigure}{0.24\textwidth}
        \centering
        \includegraphics[width=\linewidth]{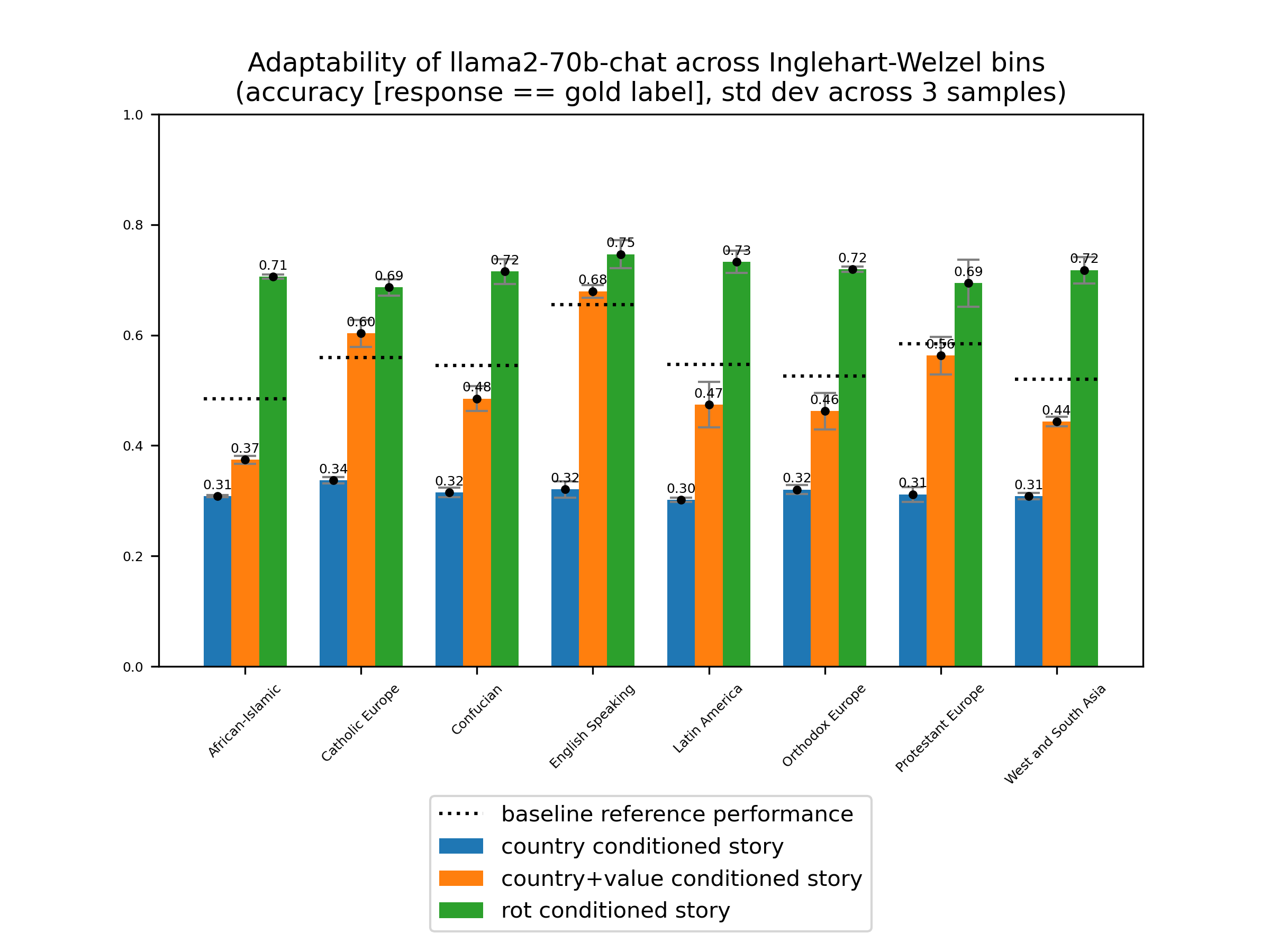}
        \caption{llama2-70b-chat}
    \end{subfigure}\\
    \begin{subfigure}{0.24\textwidth}
        \centering
        \includegraphics[width=\linewidth]{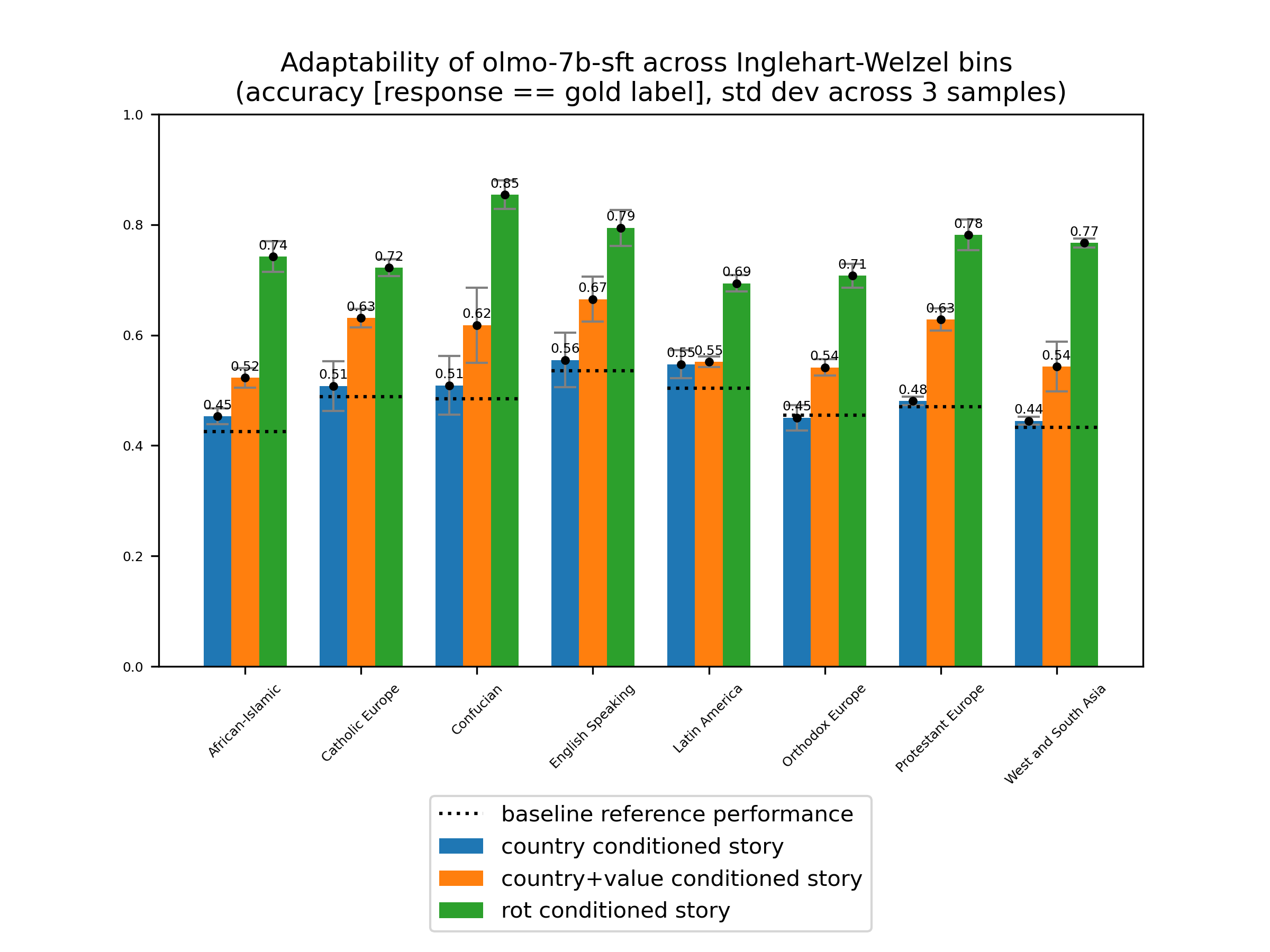}
        \caption{olmo-7b-sft}
    \end{subfigure}%
    \begin{subfigure}{0.24\textwidth}
        \centering
        \includegraphics[width=\linewidth]{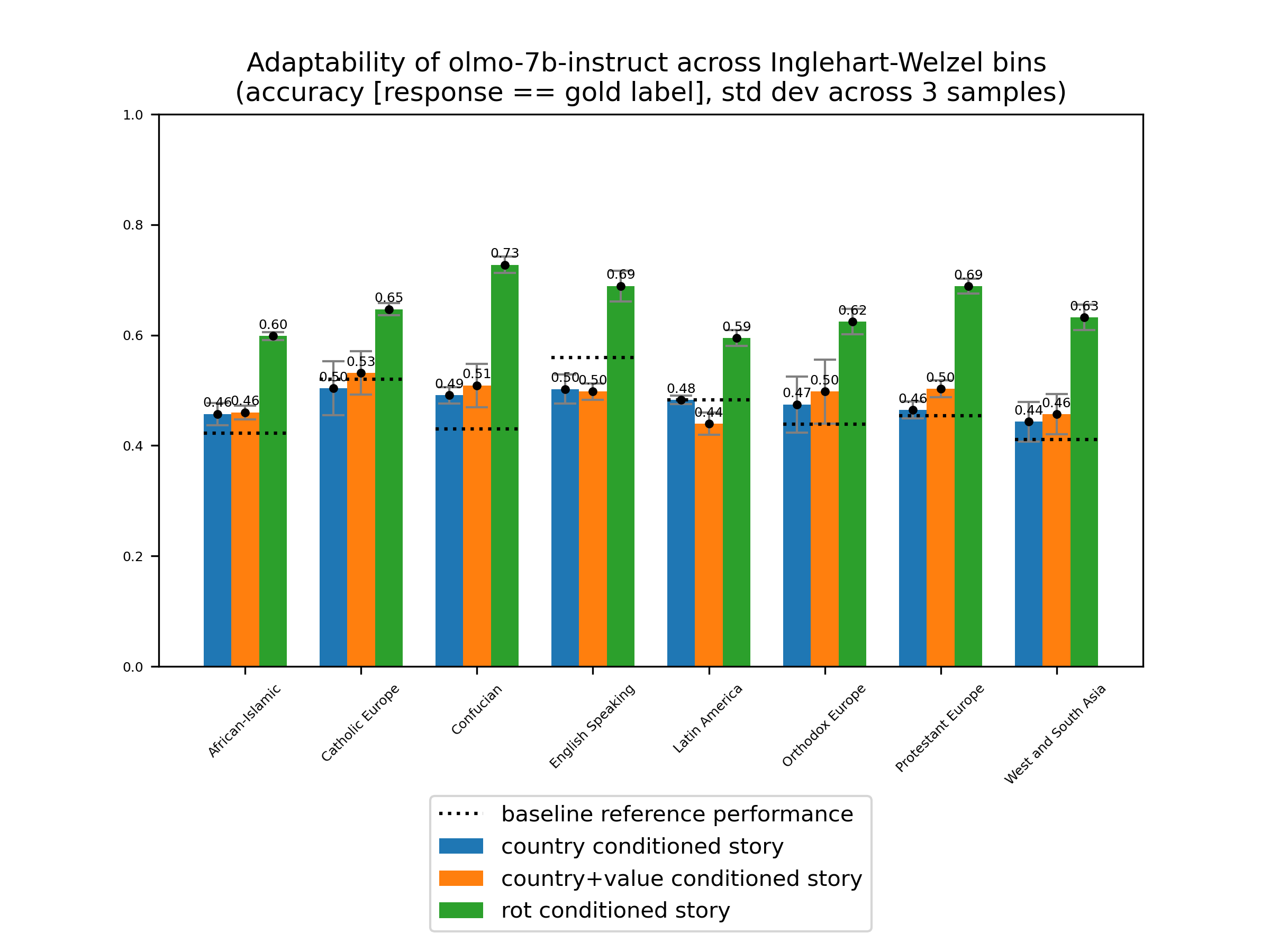}
        \caption{olmo-7b-instruct}
    \end{subfigure}%
    \begin{subfigure}{0.24\textwidth}
        \centering
        \includegraphics[width=\linewidth]{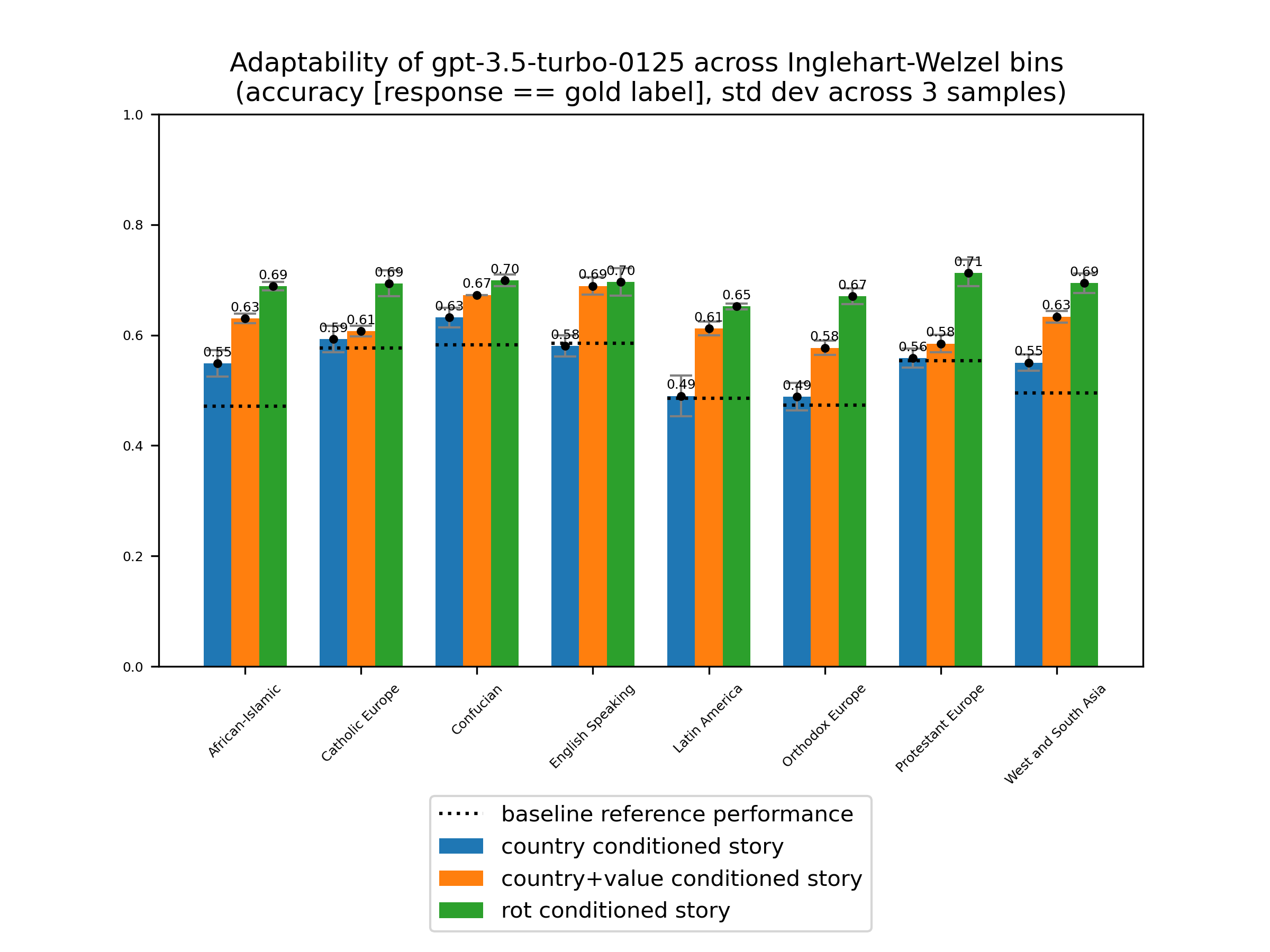}
        \caption{gpt-3.5-turbo-0125}
    \end{subfigure}\\
    \begin{subfigure}{0.24\textwidth}
        \centering
        \includegraphics[width=\linewidth]{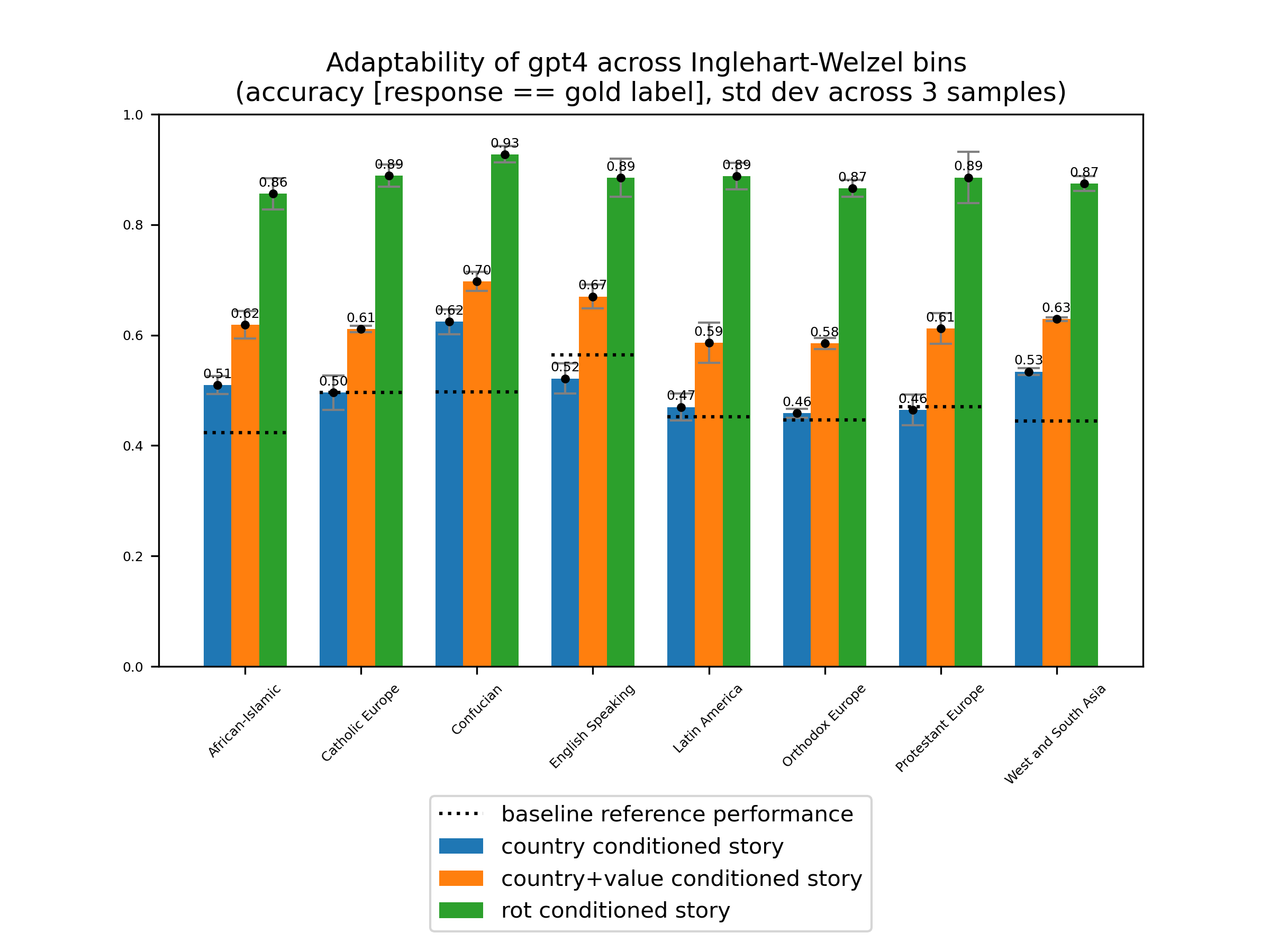}
        \caption{gpt4}
    \end{subfigure}%
    \begin{subfigure}{0.24\textwidth}
        \centering
        \includegraphics[width=\linewidth]{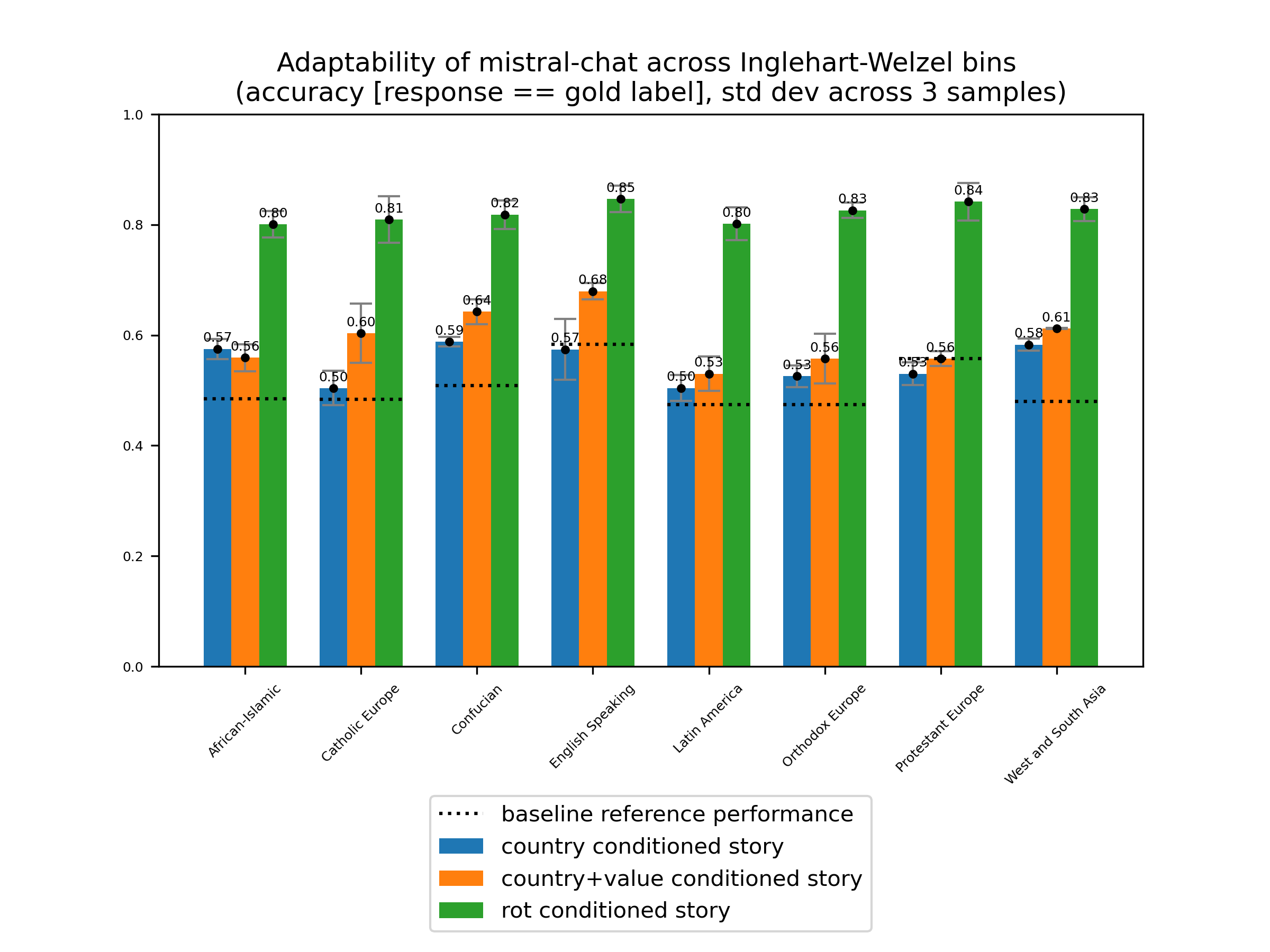}
        \caption{mistral-chat}
    \end{subfigure}%
    \caption{Accuracy across Inglehart Welzel bins for all contextualizations across all models. (blue represents country, yellow represents value, green represents rule-of-thumb. Dashed line represents baseline performance with no conditioning.}
    \label{fig:iwbin_all}
\end{figure}

\newpage
\begin{figure}[!htbp]
    \centering
    \begin{subfigure}{0.24\textwidth}
        \centering
        \includegraphics[width=\linewidth]{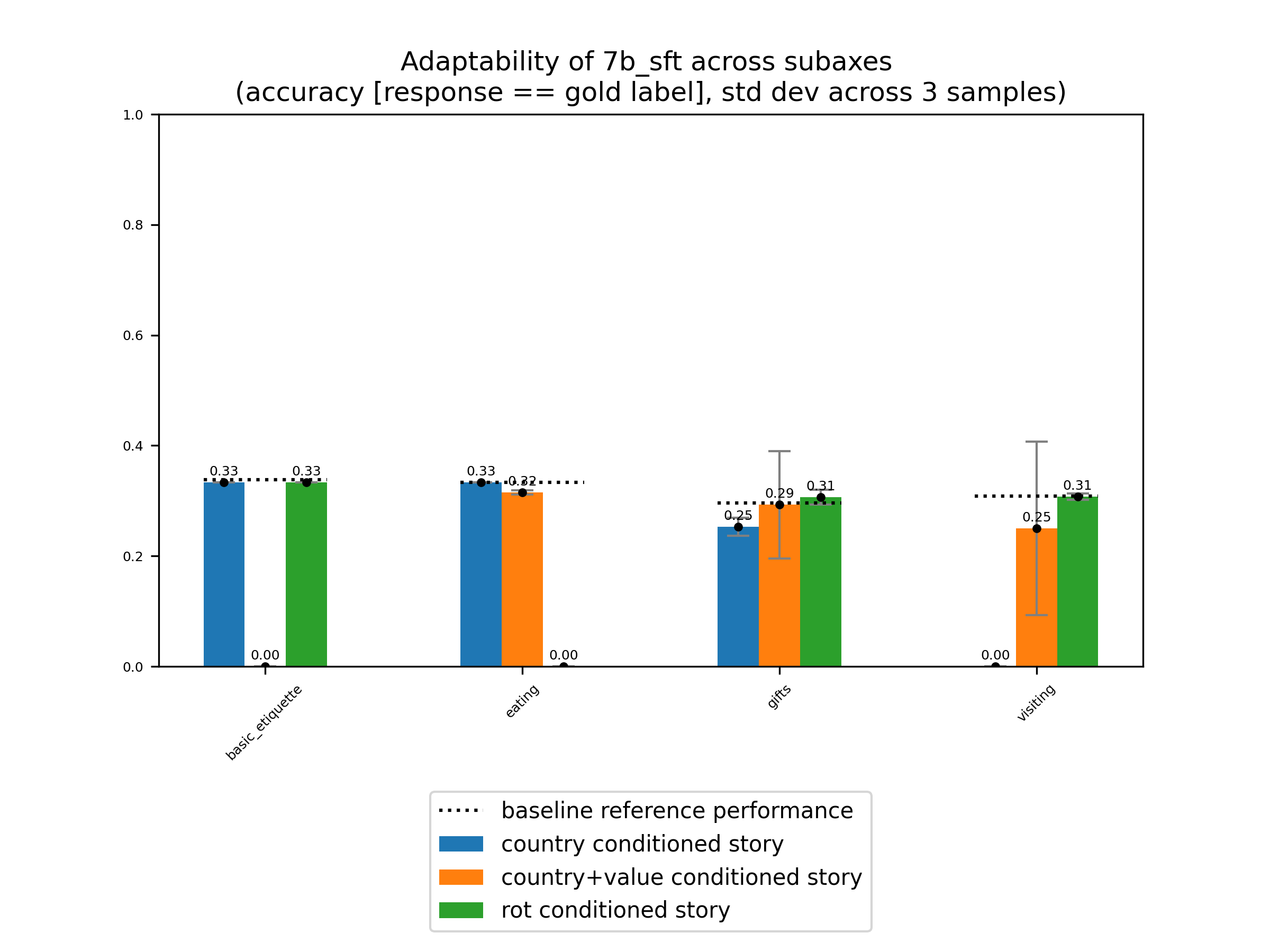}
        \caption{Archangel\_7b\_sft}
    \end{subfigure}%
    \begin{subfigure}{0.24\textwidth}
        \centering
        \includegraphics[width=\linewidth]{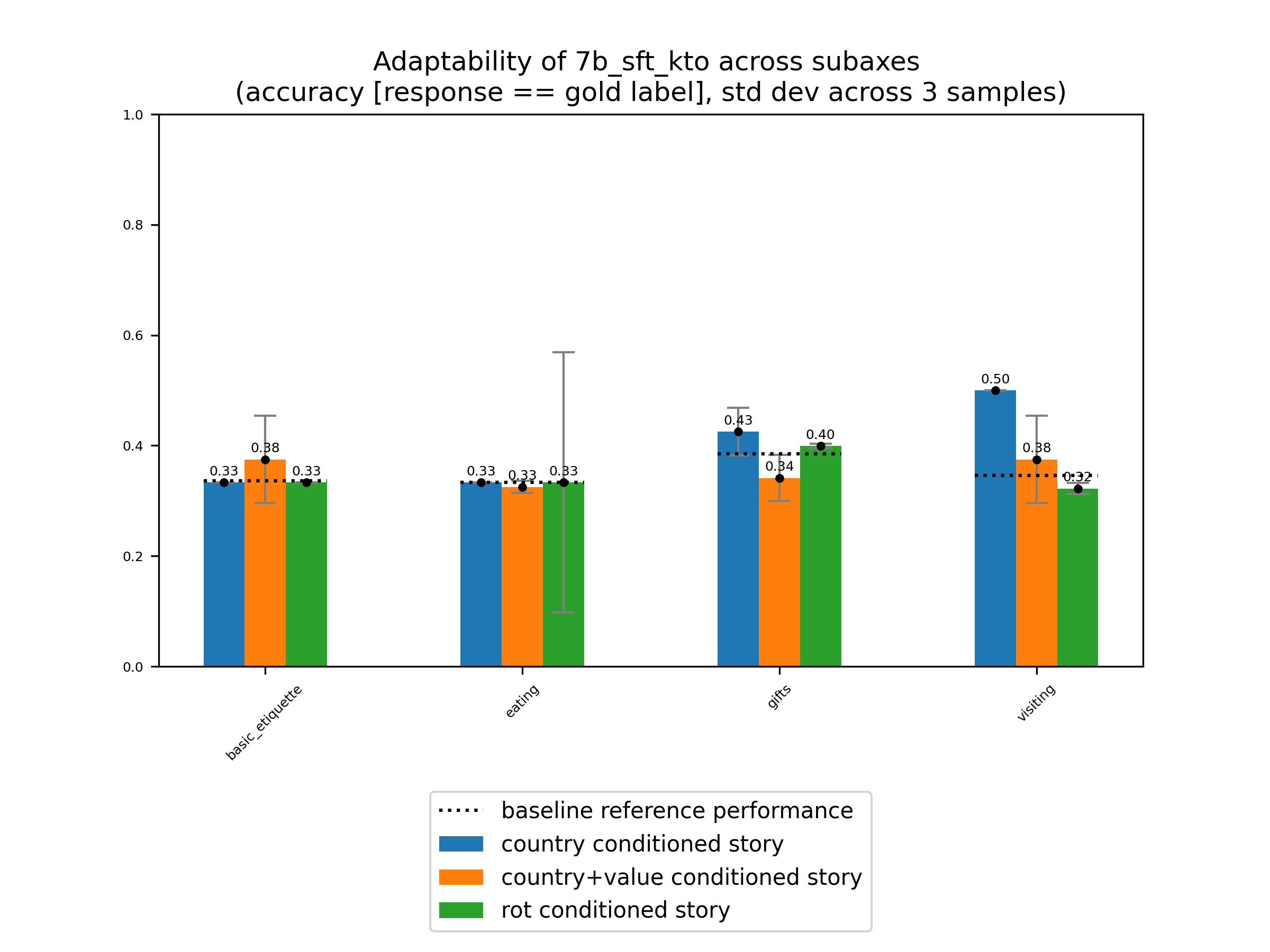}
        \caption{Archangel\_7b\_sft\_kto}
    \end{subfigure}%
    \begin{subfigure}{0.24\textwidth}
        \centering
        \includegraphics[width=\linewidth]{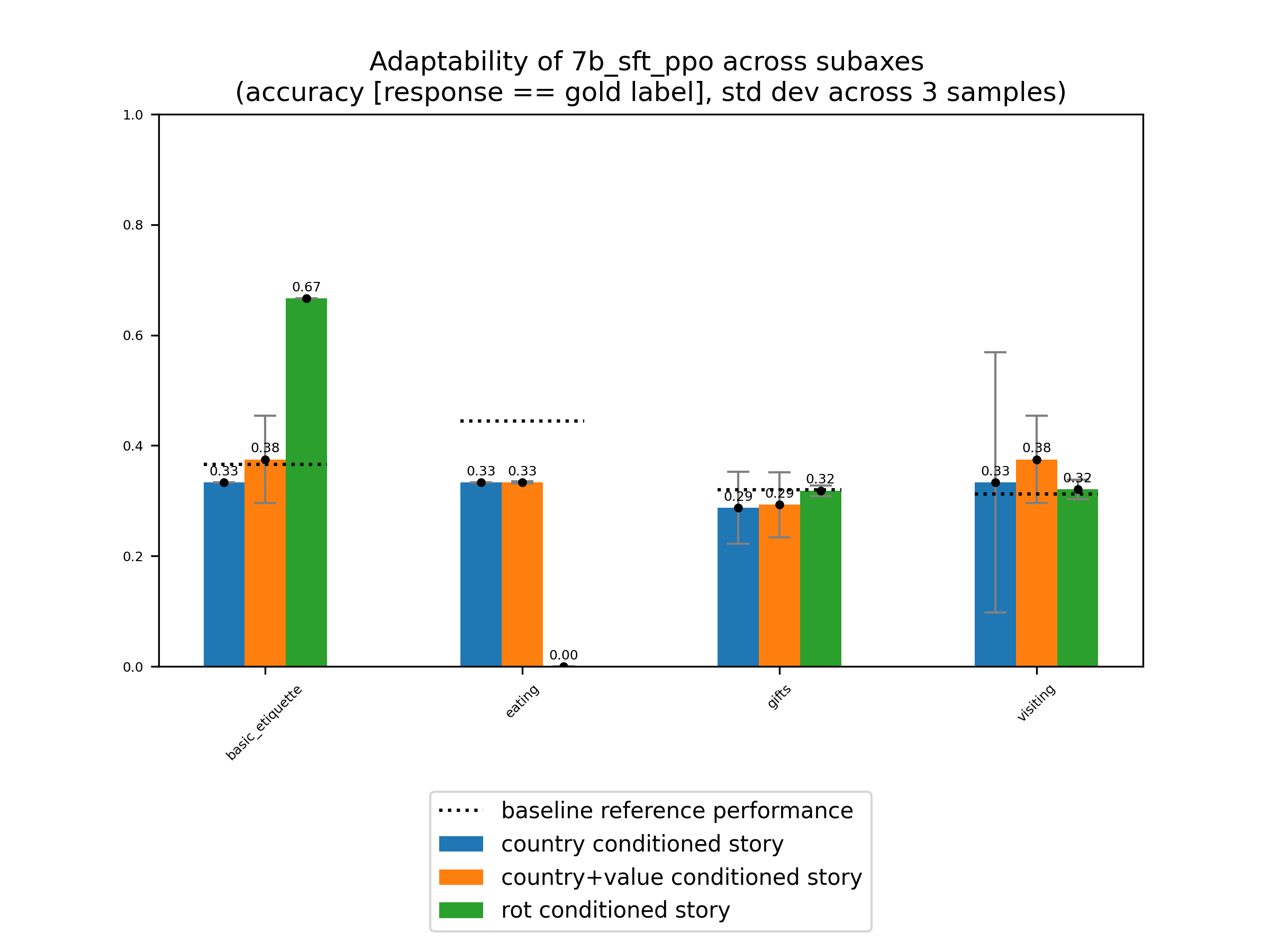}
        \caption{Archangel\_7b\_sft\_ppo}
    \end{subfigure}%
    \begin{subfigure}{0.24\textwidth}
        \centering
        \includegraphics[width=\linewidth]{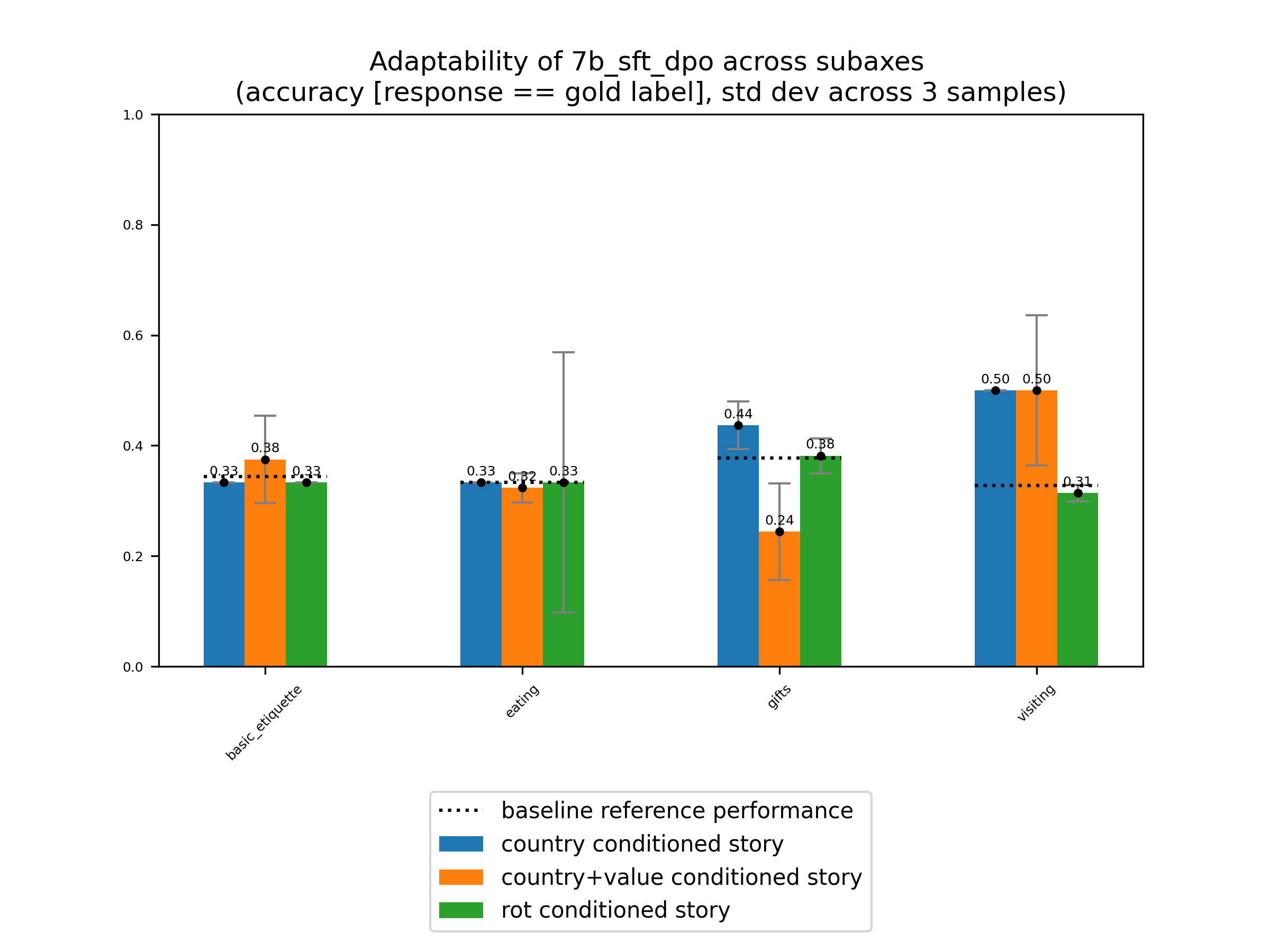}
        \caption{Archangel\_7b\_sft\_dpo}
    \end{subfigure}\\
    \begin{subfigure}{0.24\textwidth}
        \centering
        \includegraphics[width=\linewidth]{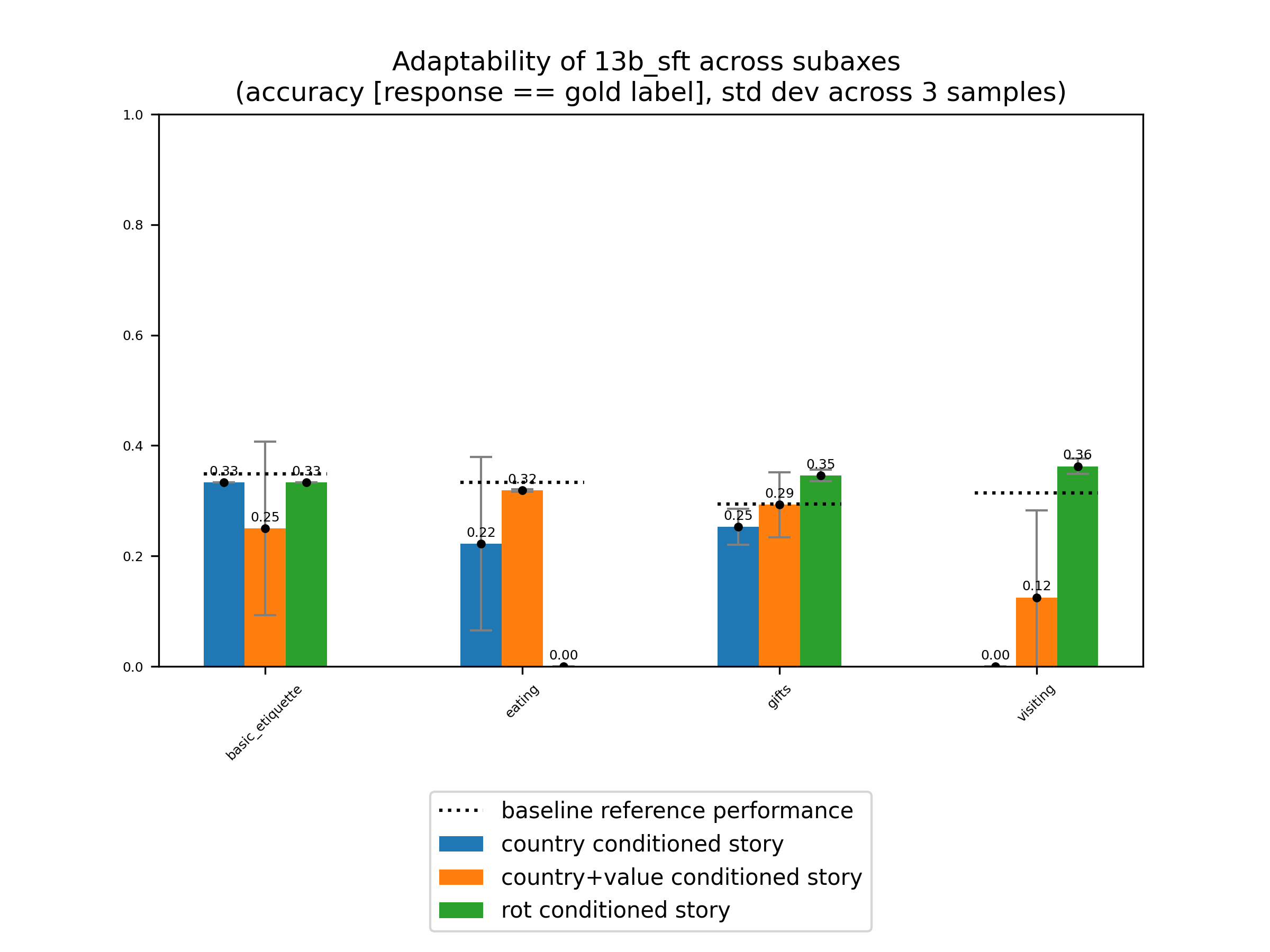}
        \caption{Archangel\_13b\_sft}
    \end{subfigure}%
    \begin{subfigure}{0.24\textwidth}
        \centering
        \includegraphics[width=\linewidth]{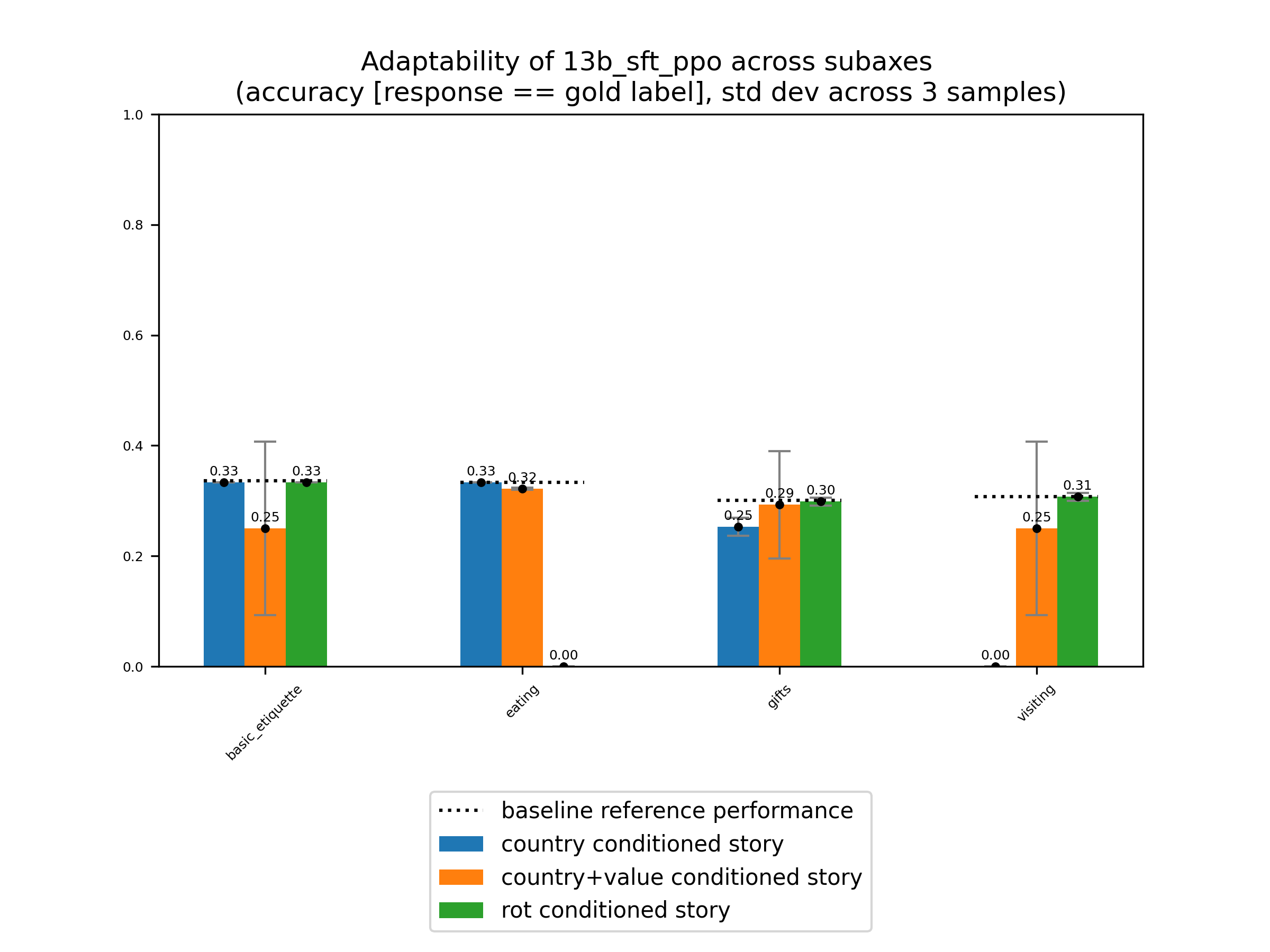}
        \caption{Archangel\_13b\_sft\_ppo}
    \end{subfigure}%
    \begin{subfigure}{0.24\textwidth}
        \centering
        \includegraphics[width=\linewidth]{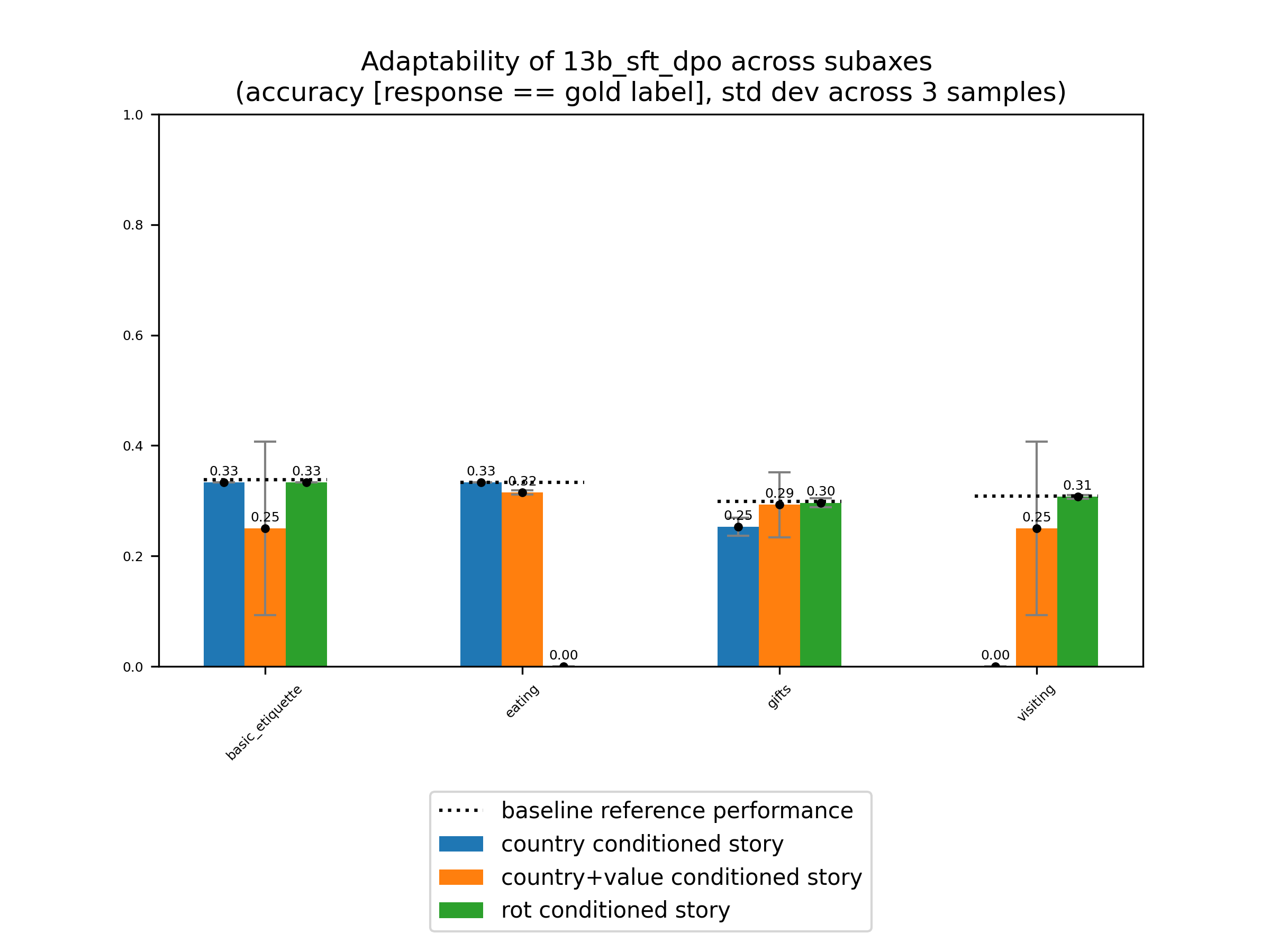}
        \caption{Archangel\_13b\_sft\_dpo}
    \end{subfigure}%
    \begin{subfigure}{0.24\textwidth}
        \centering
        \includegraphics[width=\linewidth]{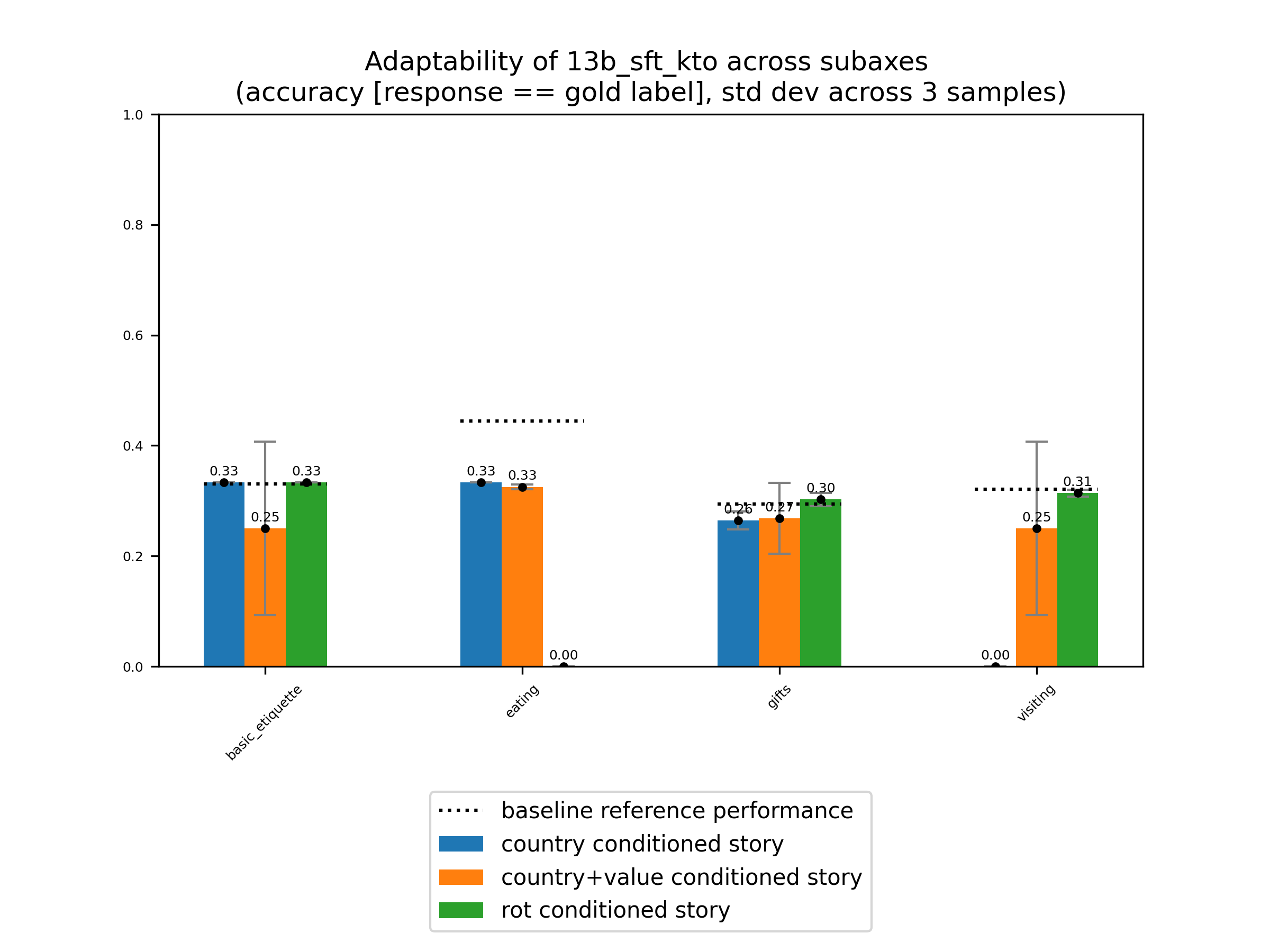}
        \caption{Archangel\_13b\_sft\_kto}
    \end{subfigure}\\
    \begin{subfigure}{0.24\textwidth}
        \centering
        \includegraphics[width=\linewidth]{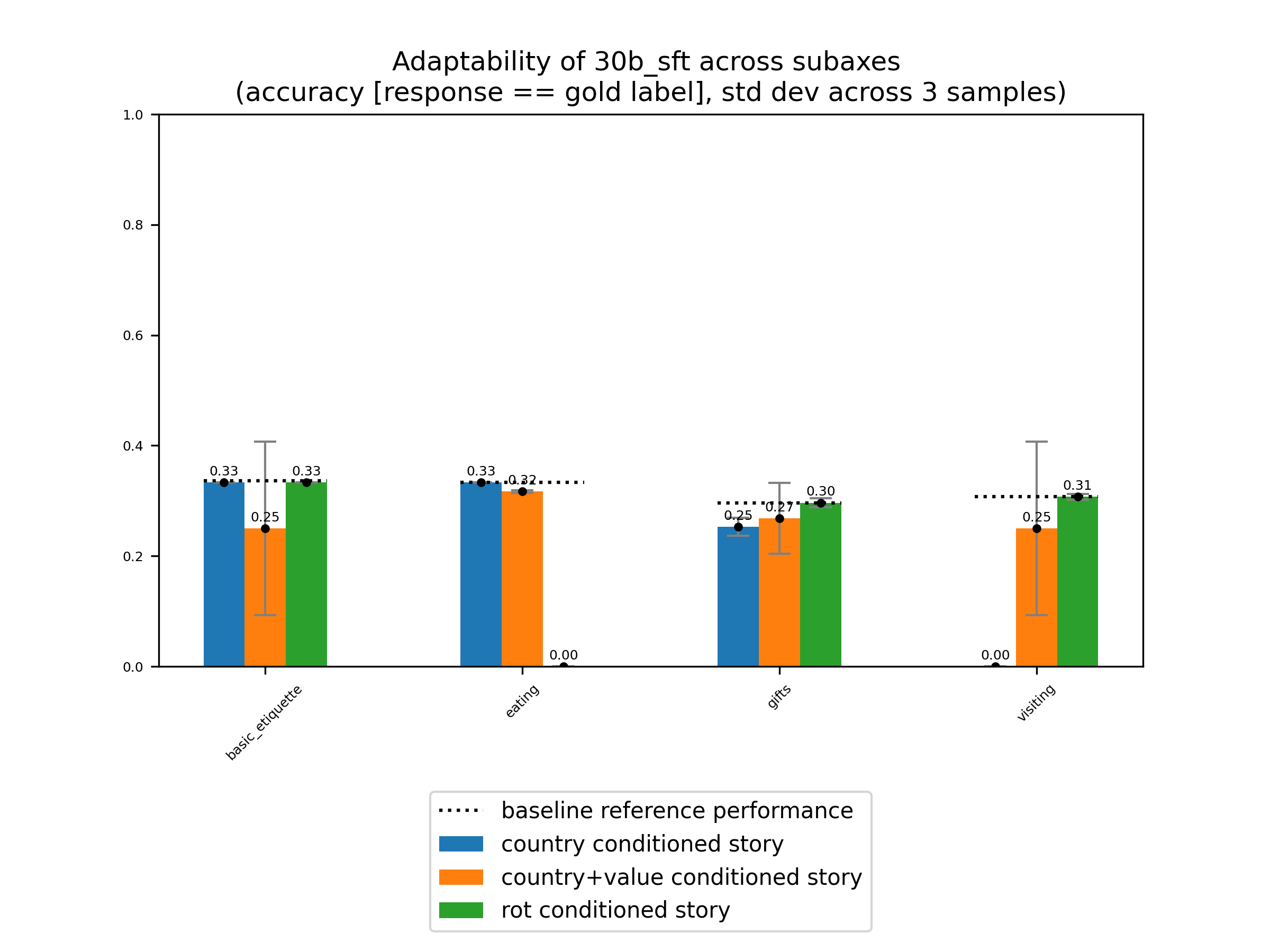}
        \caption{Archangel\_30b\_sft}
    \end{subfigure}%
    \begin{subfigure}{0.24\textwidth}
        \centering
        \includegraphics[width=\linewidth]{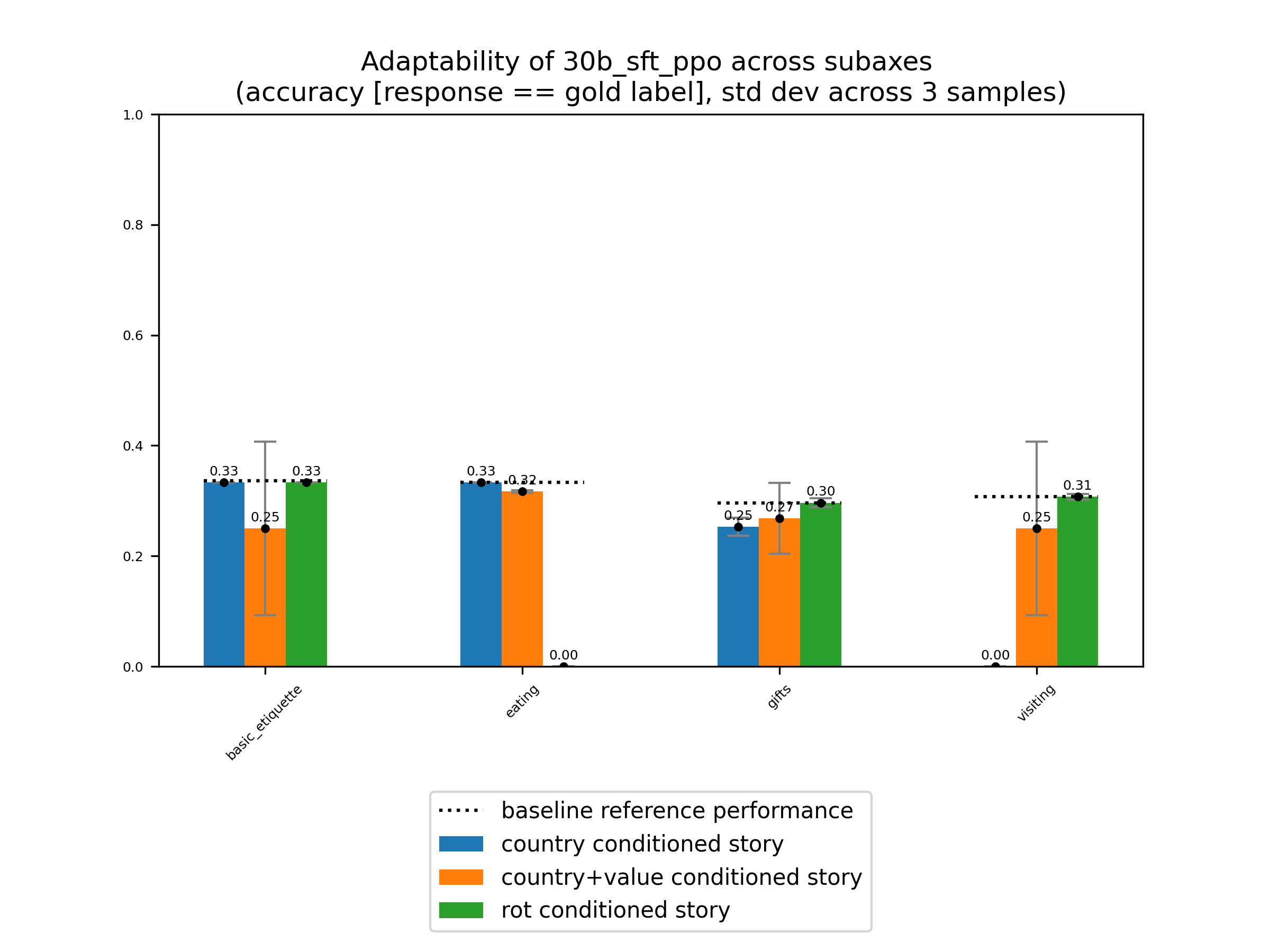}
        \caption{Archangel\_30b\_sft\_ppo}
    \end{subfigure}%
    \begin{subfigure}{0.24\textwidth}
        \centering
        \includegraphics[width=\linewidth]{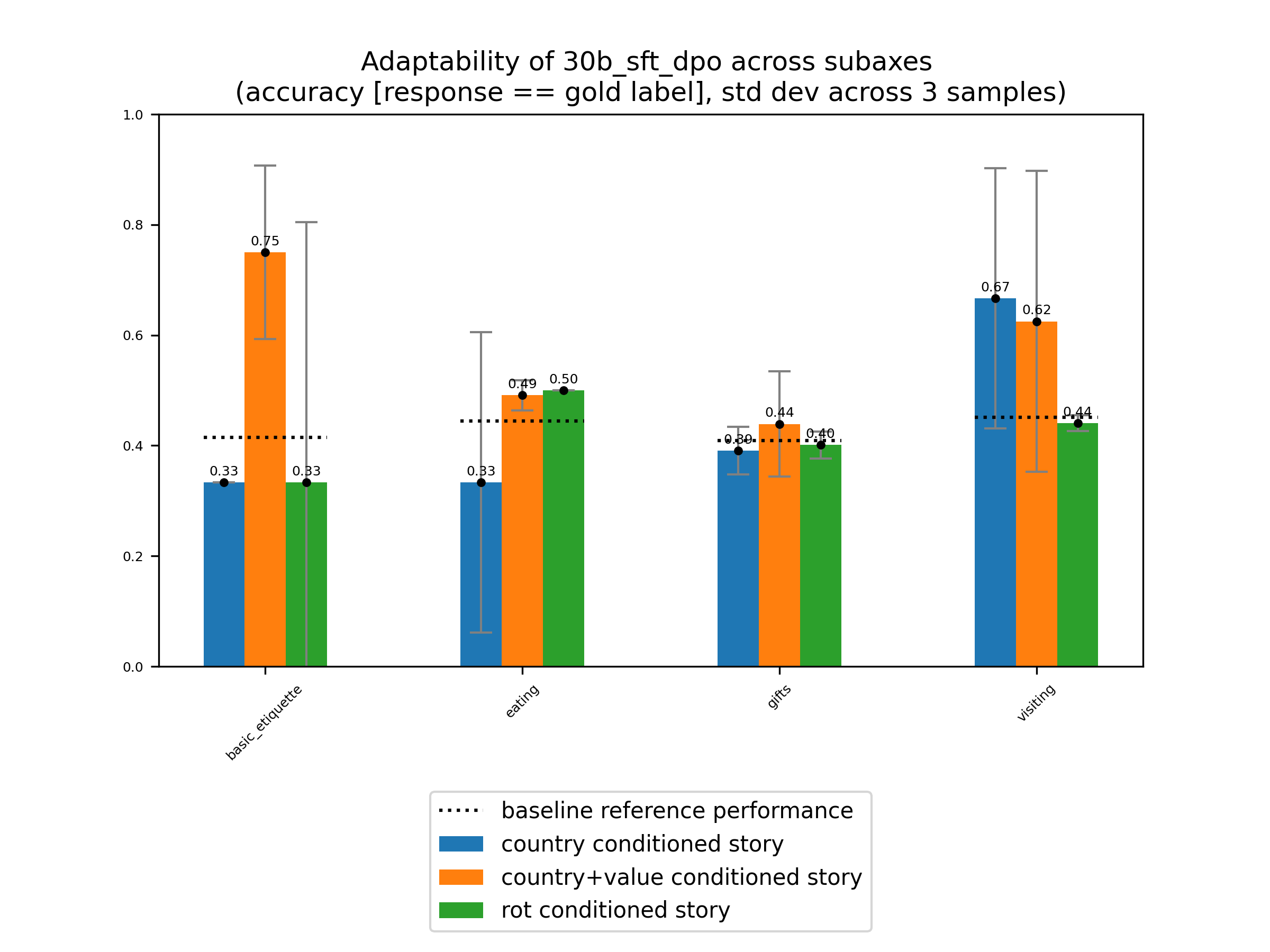}
        \caption{Archangel\_30b\_sft\_dpo}
    \end{subfigure}%
    \begin{subfigure}{0.24\textwidth}
        \centering
        \includegraphics[width=\linewidth]{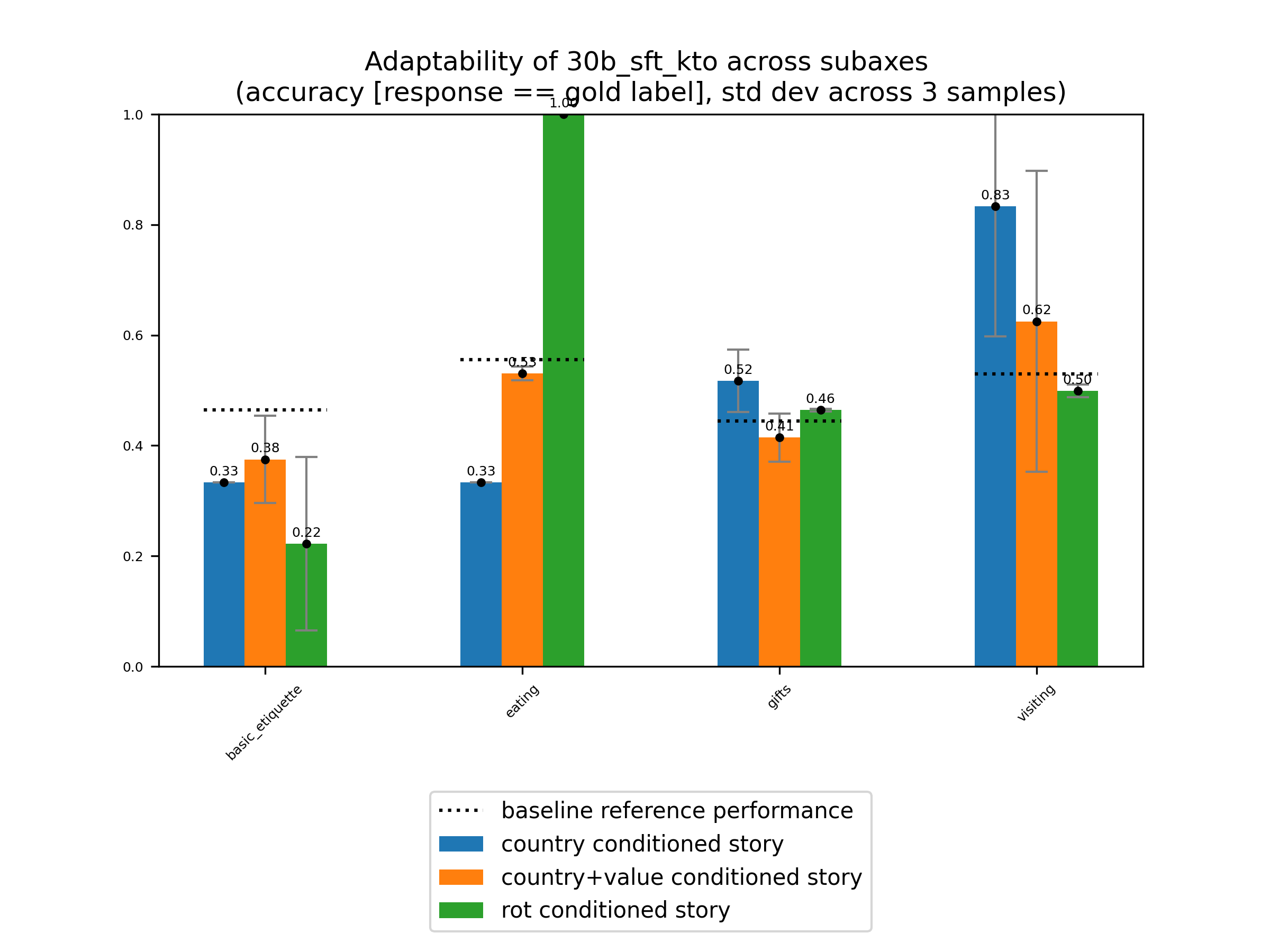}
        \caption{Archangel\_30b\_sft\_kto}
    \end{subfigure}\\
    \begin{subfigure}{0.24\textwidth}
        \centering
        \includegraphics[width=\linewidth]{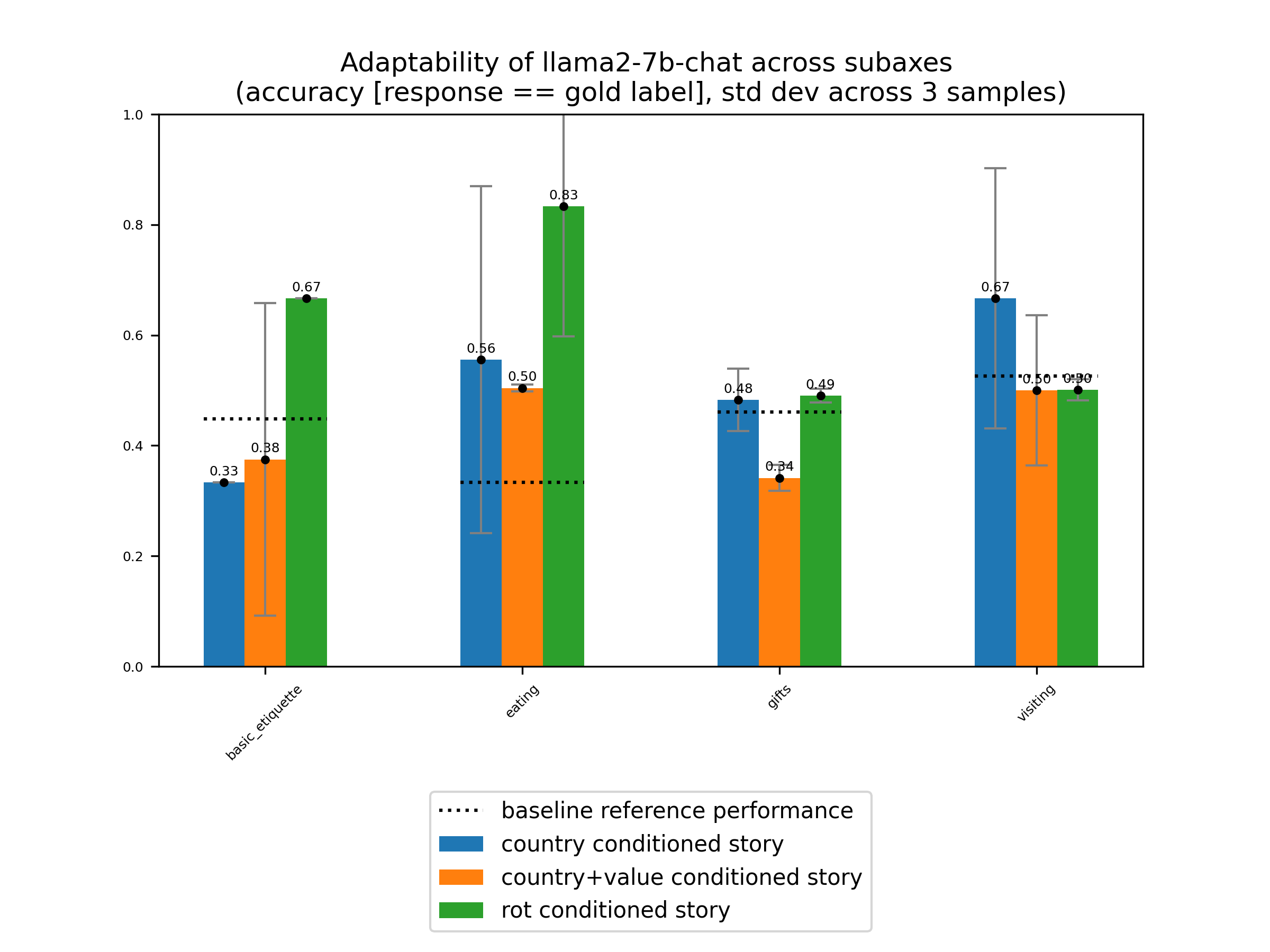}
        \caption{llama2-7b-chat}
    \end{subfigure}%
    \begin{subfigure}{0.24\textwidth}
        \centering
        \includegraphics[width=\linewidth]{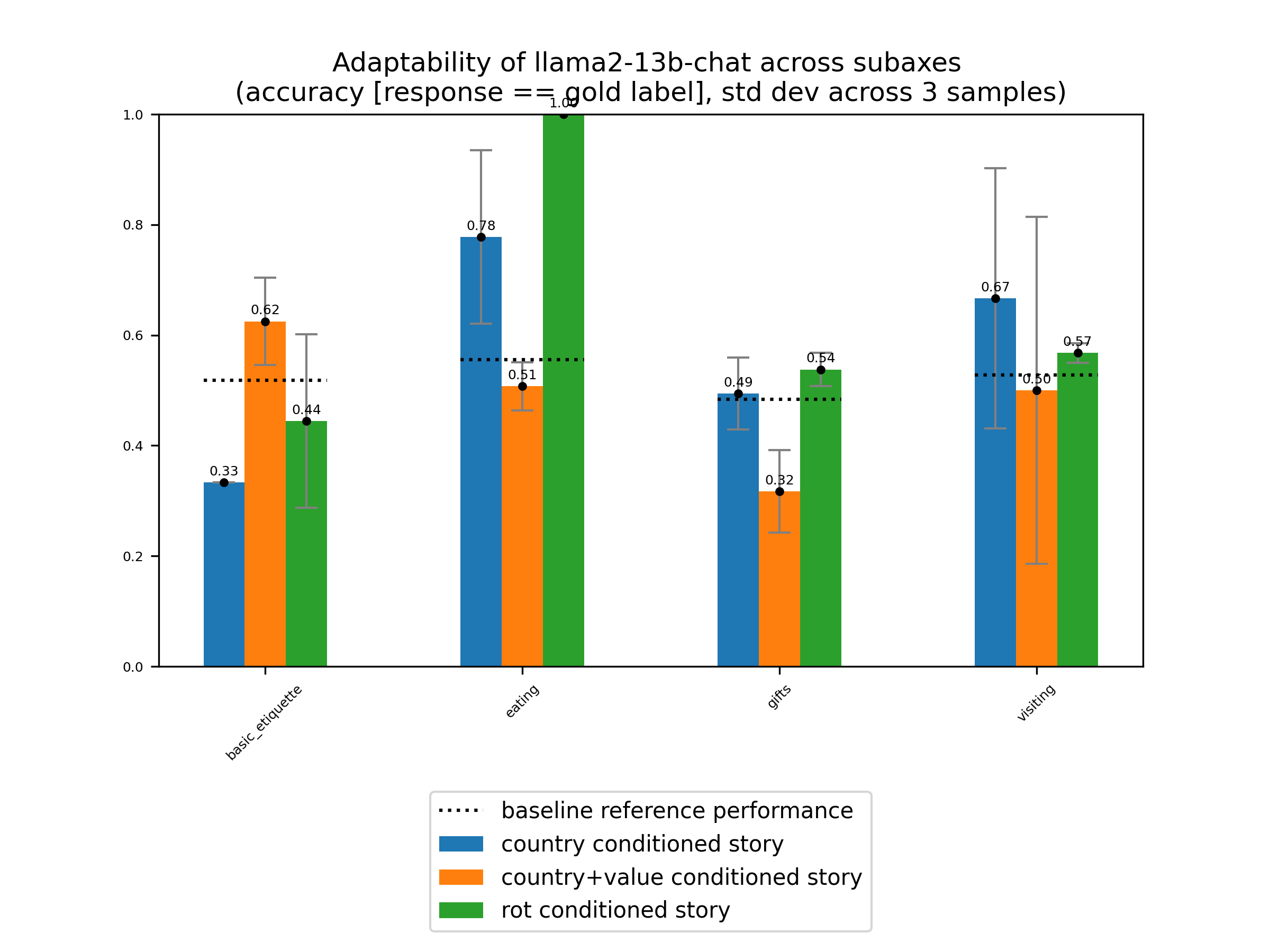}
        \caption{llama2-13b-chat}
    \end{subfigure}%
    \begin{subfigure}{0.24\textwidth}
        \centering
        \includegraphics[width=\linewidth]{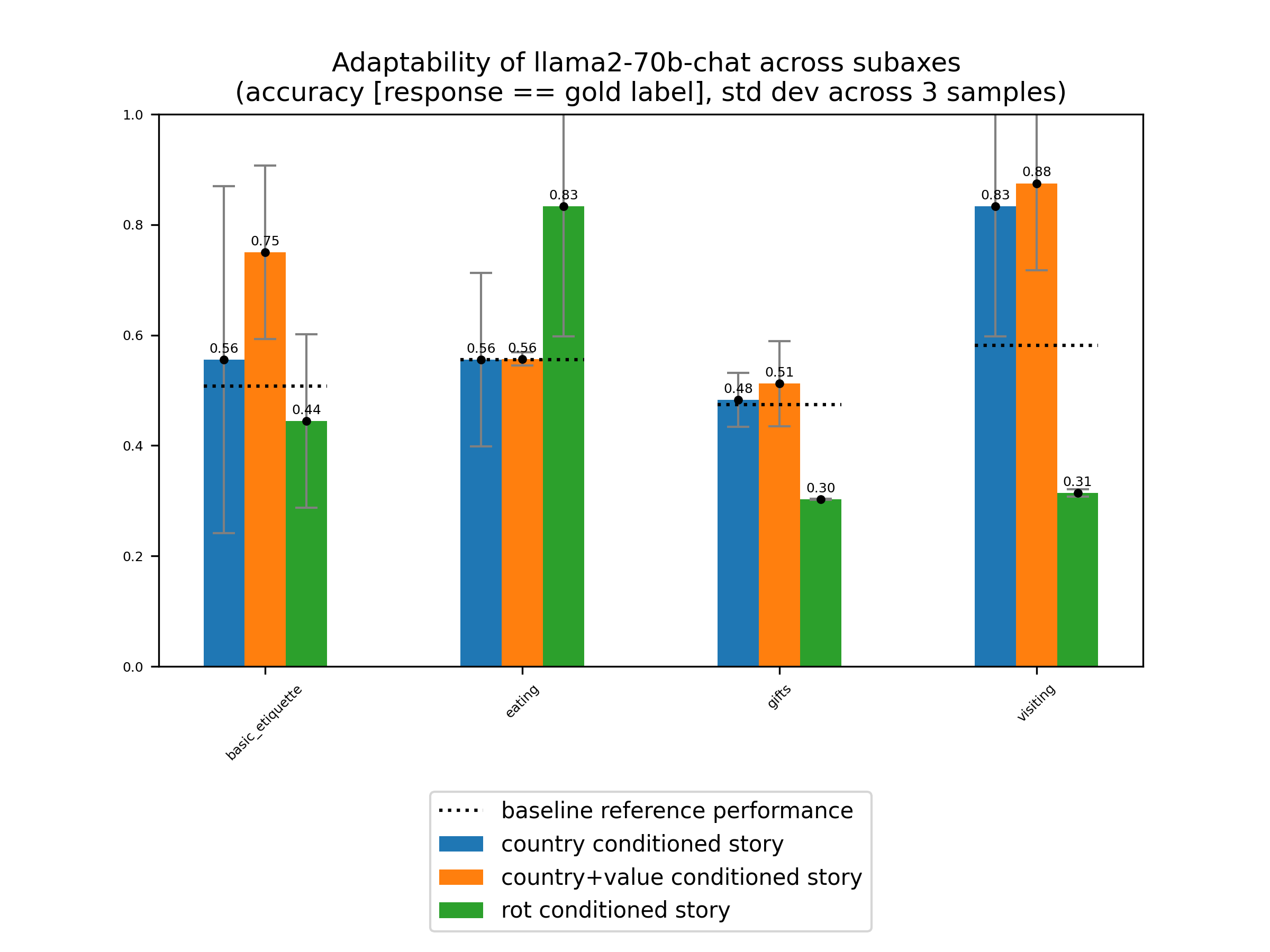}
        \caption{llama2-70b-chat}
    \end{subfigure}\\
    \begin{subfigure}{0.24\textwidth}
        \centering
        \includegraphics[width=\linewidth]{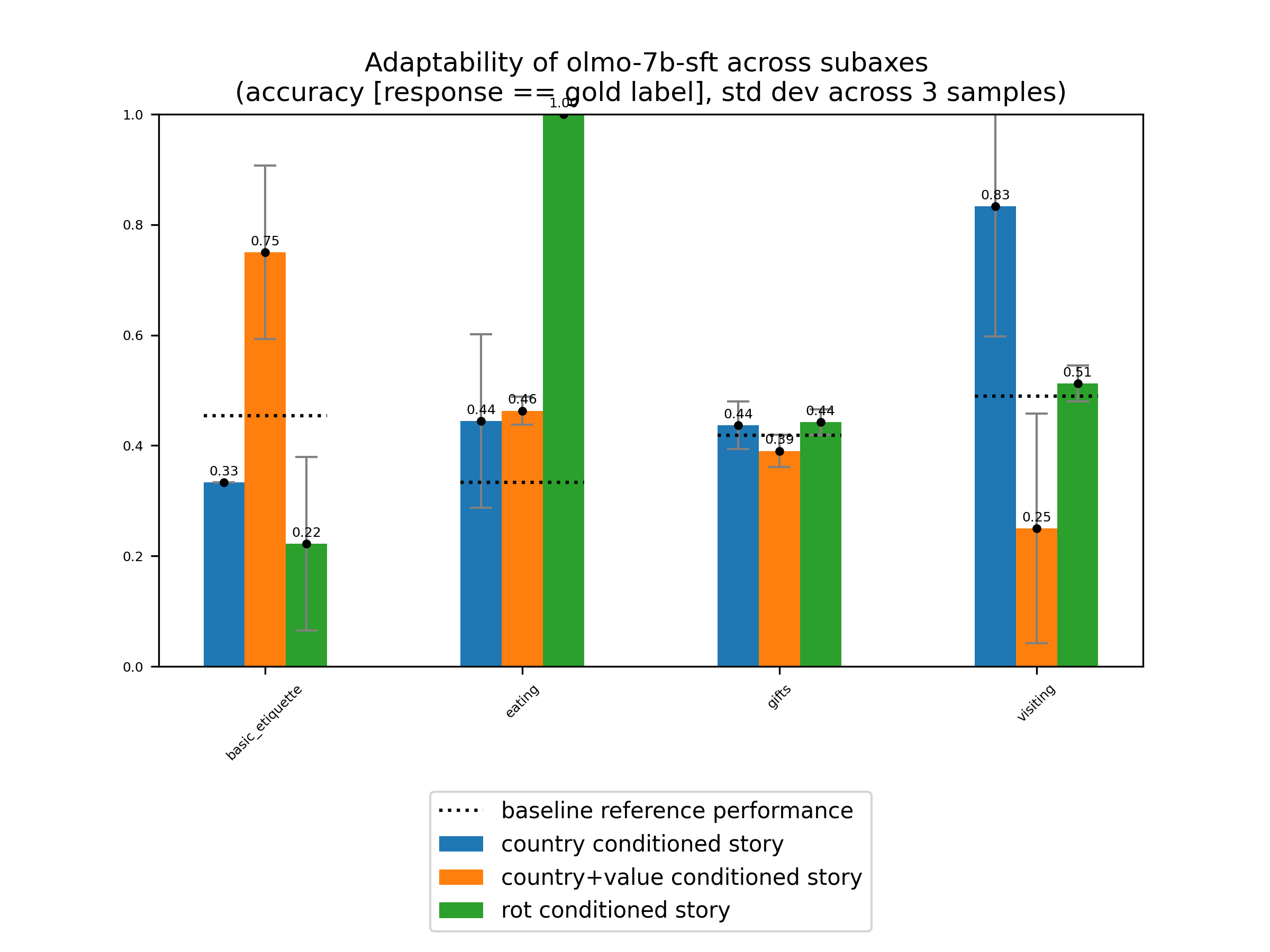}
        \caption{olmo-7b-sft}
    \end{subfigure}%
    \begin{subfigure}{0.24\textwidth}
        \centering
        \includegraphics[width=\linewidth]{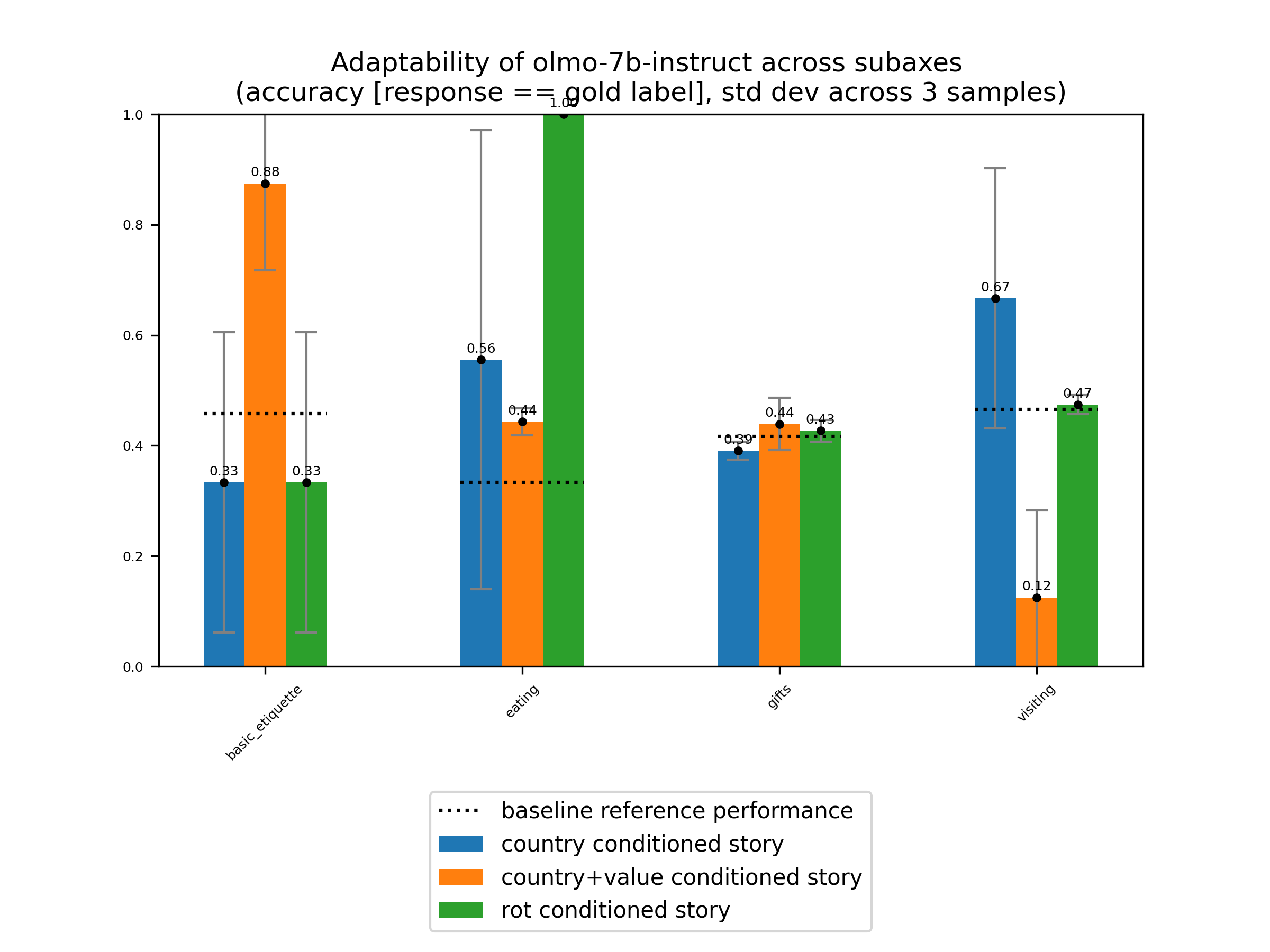}
        \caption{olmo-7b-instruct}
    \end{subfigure}%
    \begin{subfigure}{0.24\textwidth}
        \centering
        \includegraphics[width=\linewidth]{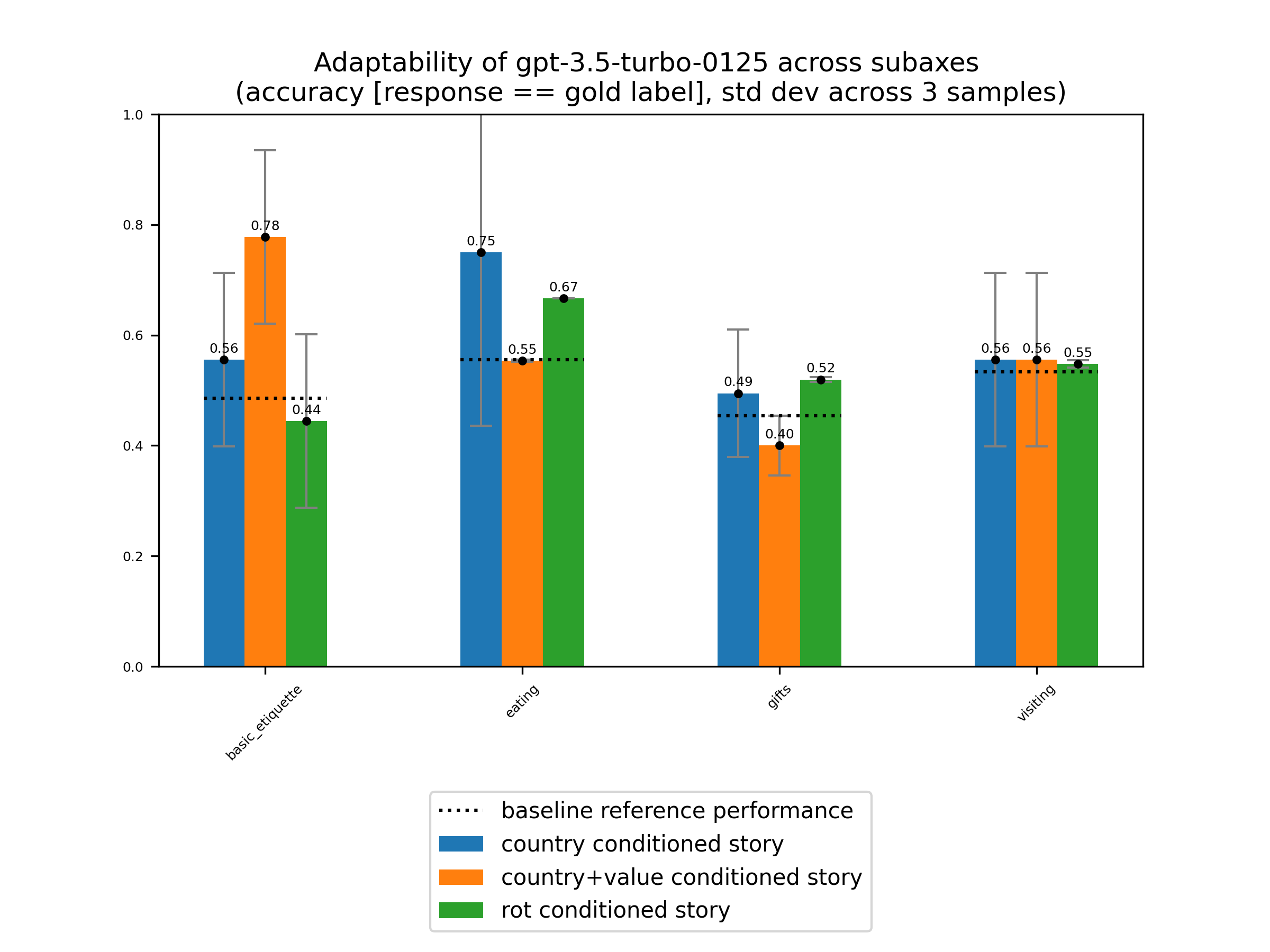}
        \caption{gpt-3.5-turbo-0125}
    \end{subfigure}\\
    \begin{subfigure}{0.24\textwidth}
        \centering
        \includegraphics[width=\linewidth]{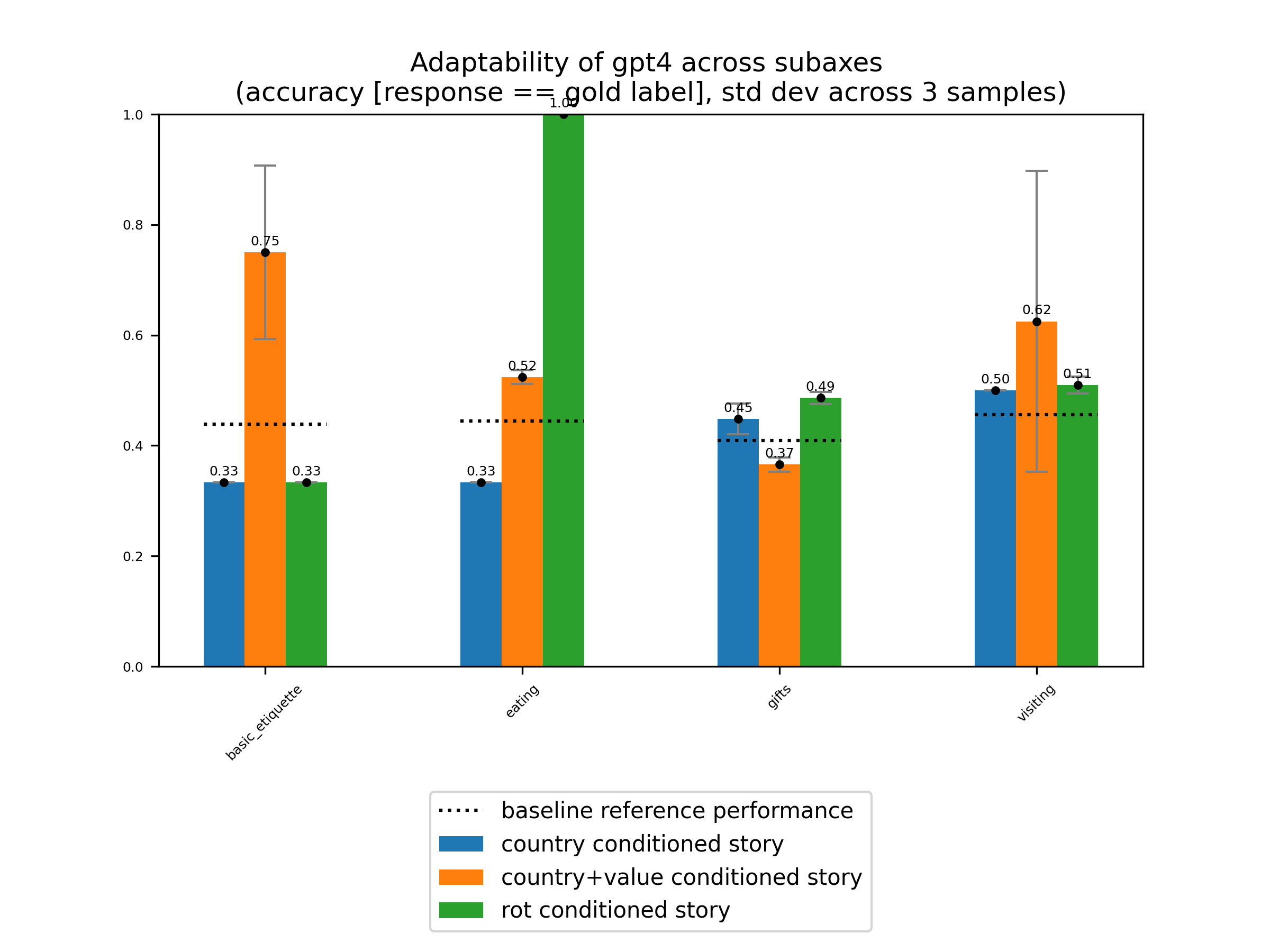}
        \caption{gpt4}
    \end{subfigure}%
    \begin{subfigure}{0.24\textwidth}
        \centering
        \includegraphics[width=\linewidth]{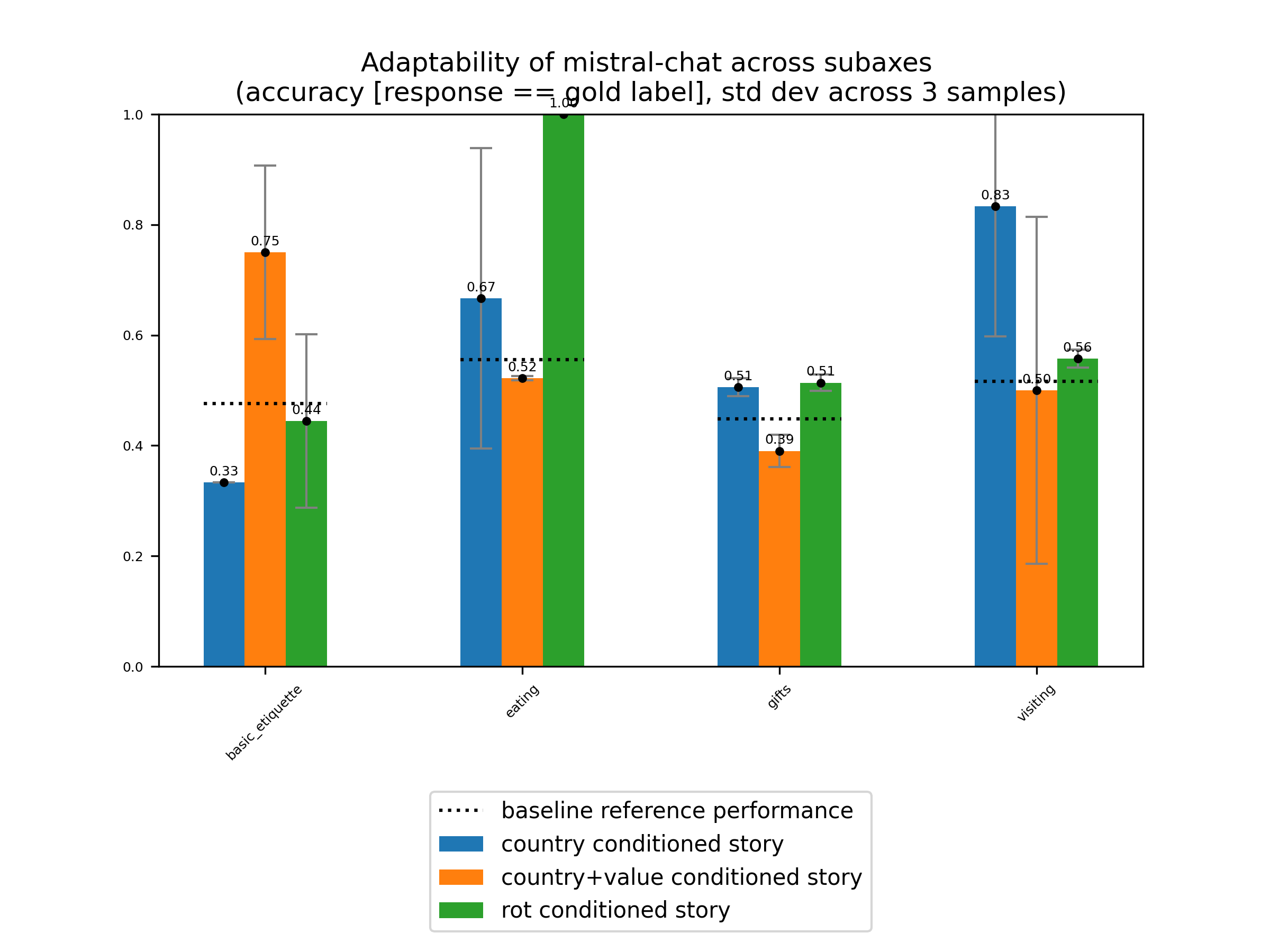}
        \caption{mistral-chat}
    \end{subfigure}%
    \caption{Accuracy across subaxes (eating, visiting, gifts, basic\_etiquette) for all contextualizations across all models. Blue represents country, yellow represents country+value, green represents rule-of-thumb. Dashed line represents baseline performance with no conditioning.}
    \label{fig:subaxes_all}
\end{figure}

\newpage
\begin{figure}[!htbp]
    \centering
    \begin{subfigure}{0.24\textwidth}
        \centering
        \includegraphics[width=\linewidth]{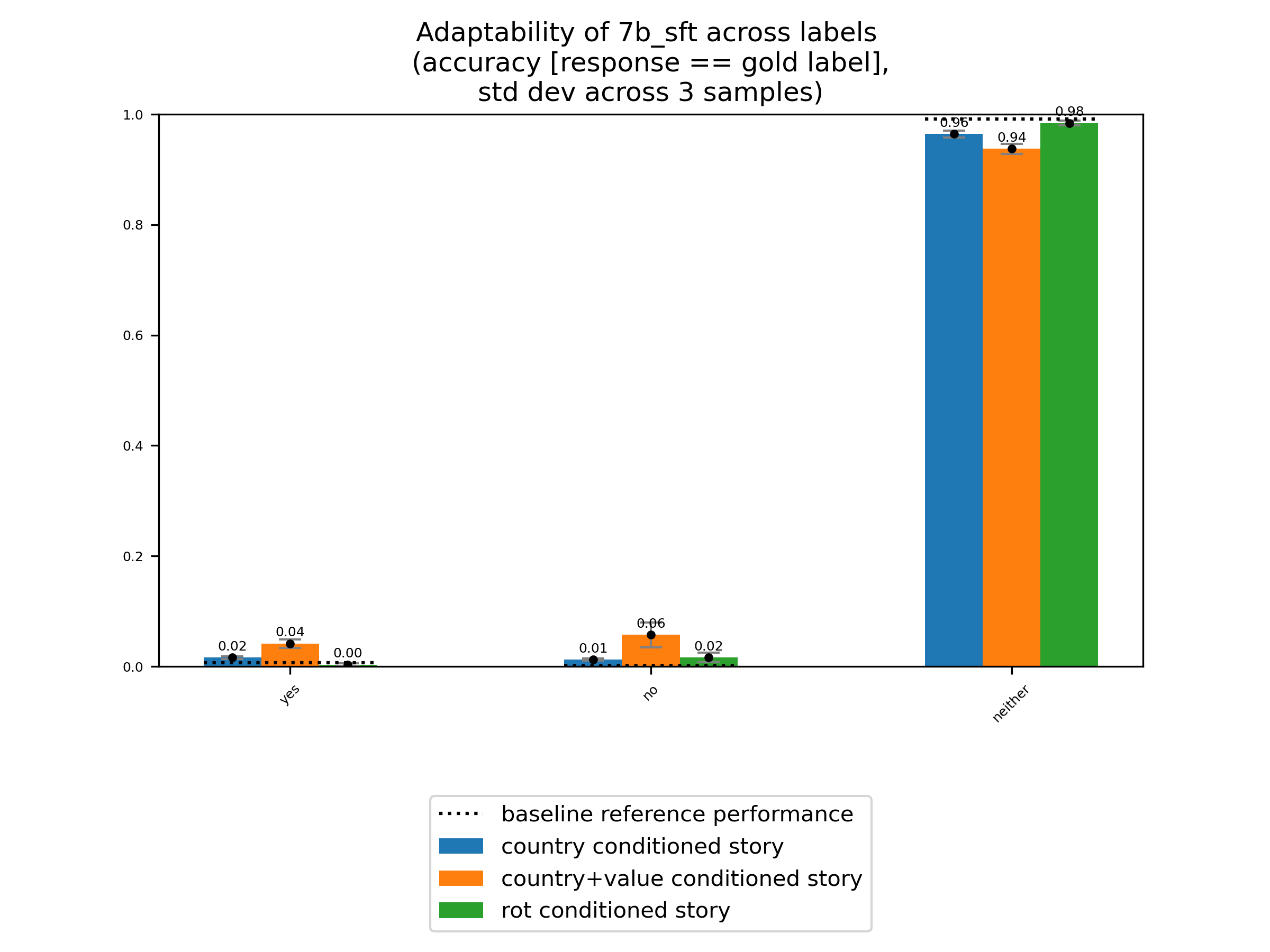}
        \caption{Archangel\_7b\_sft}
    \end{subfigure}%
    \begin{subfigure}{0.24\textwidth}
        \centering
        \includegraphics[width=\linewidth]{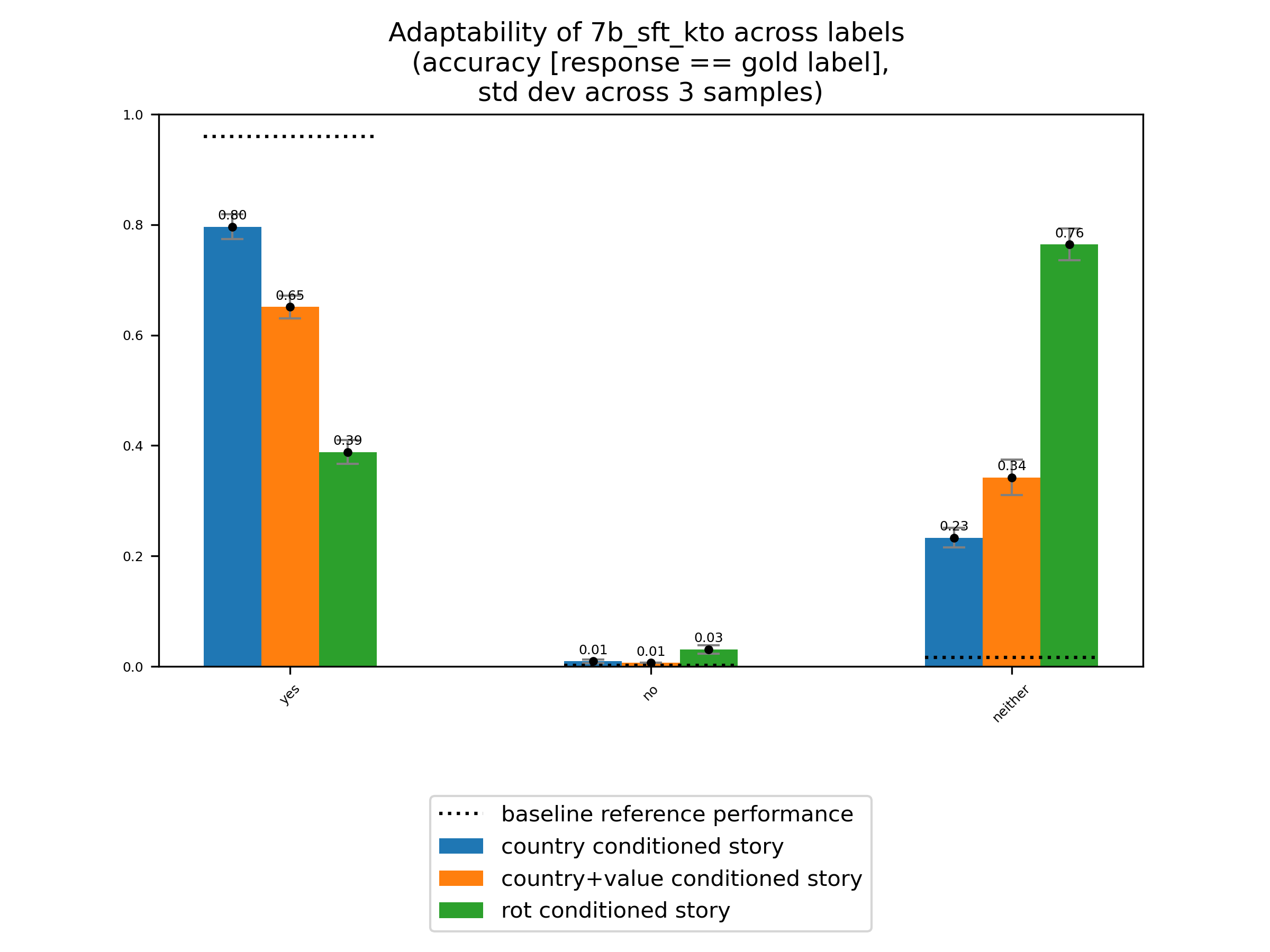}
        \caption{Archangel\_7b\_sft\_kto}
    \end{subfigure}%
    \begin{subfigure}{0.24\textwidth}
        \centering
        \includegraphics[width=\linewidth]{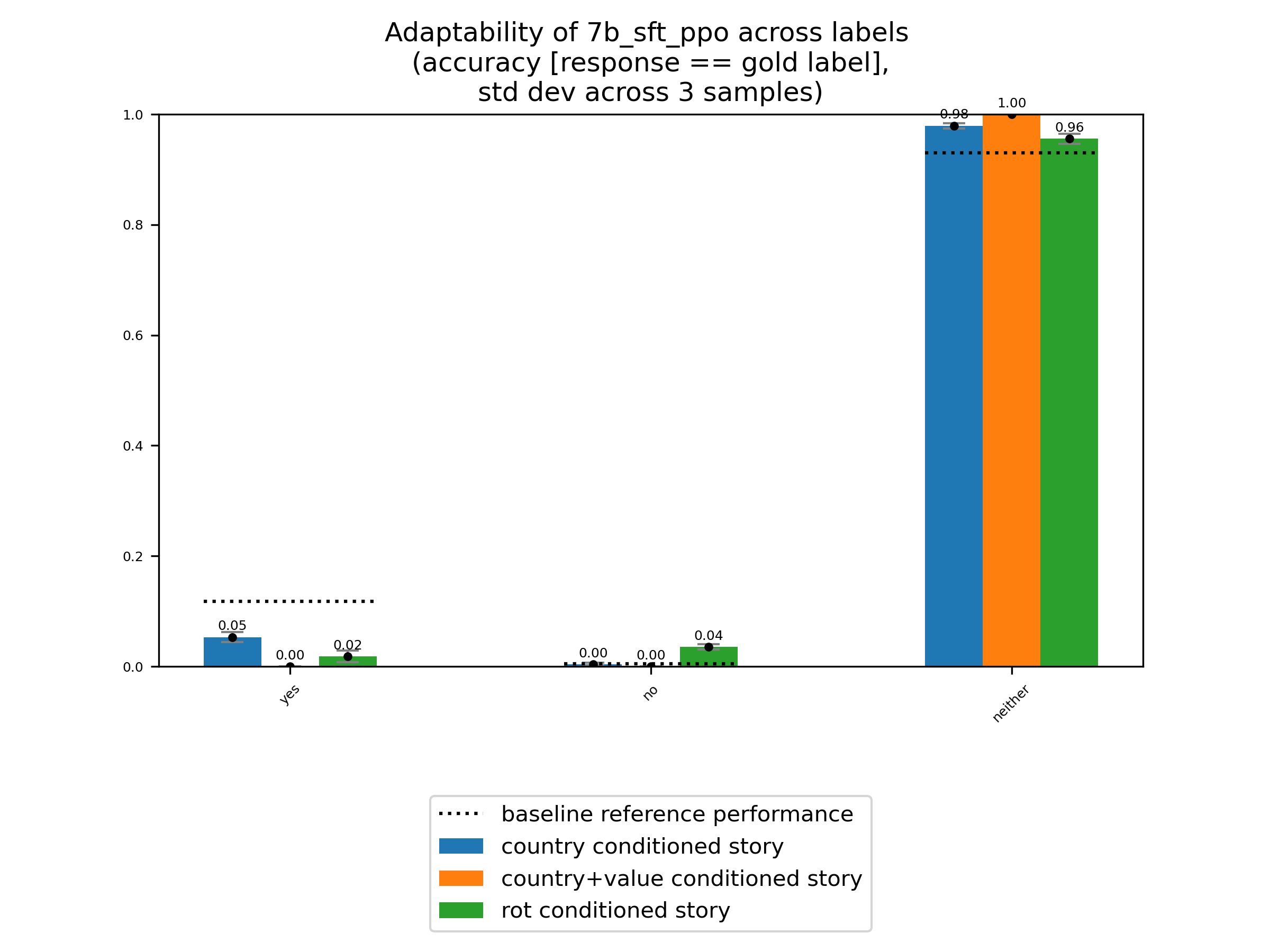}
        \caption{Archangel\_7b\_sft\_ppo}
    \end{subfigure}%
    \begin{subfigure}{0.24\textwidth}
        \centering
        \includegraphics[width=\linewidth]{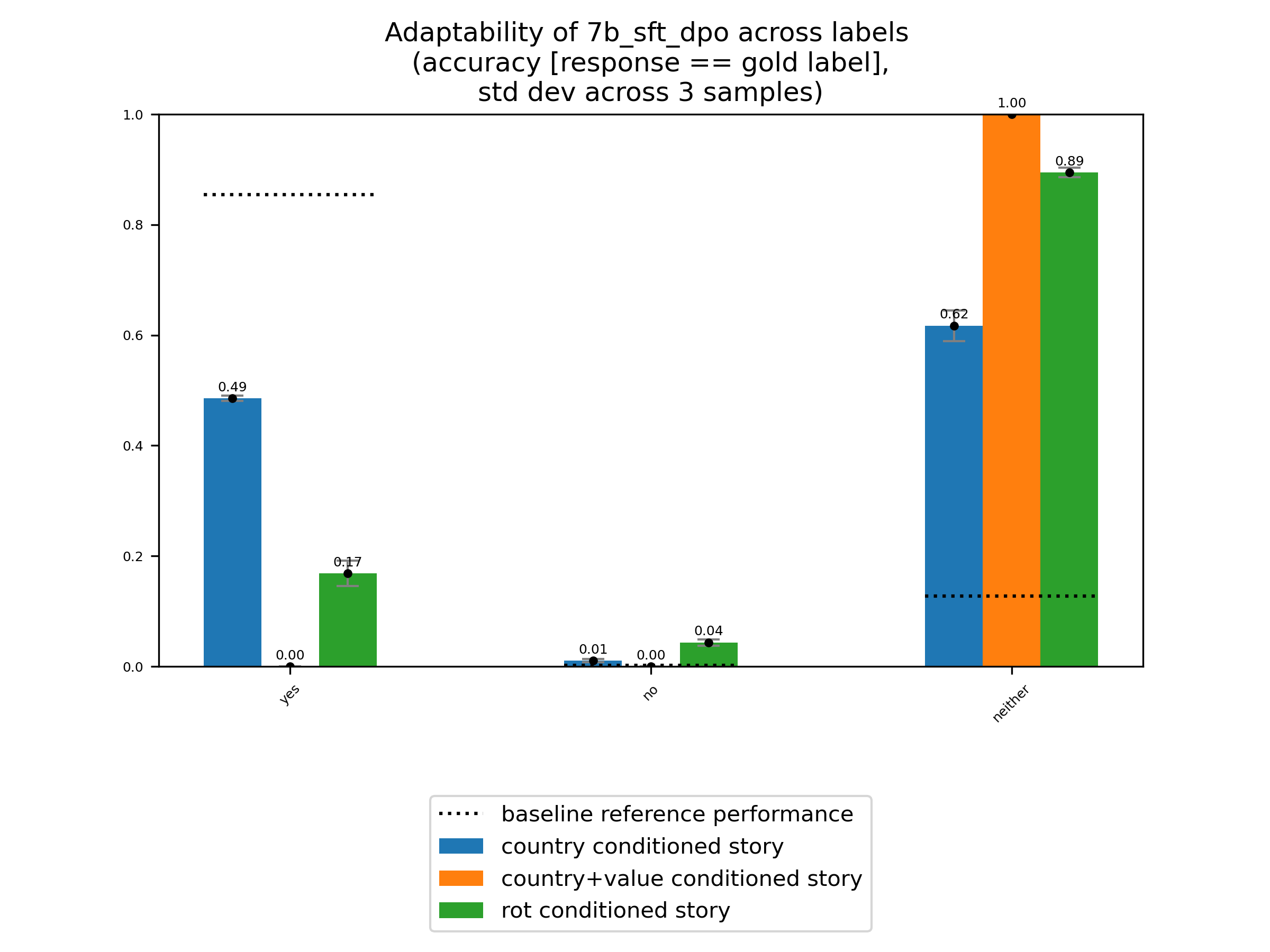}
        \caption{Archangel\_7b\_sft\_dpo}
    \end{subfigure}\\
    \begin{subfigure}{0.24\textwidth}
        \centering
        \includegraphics[width=\linewidth]{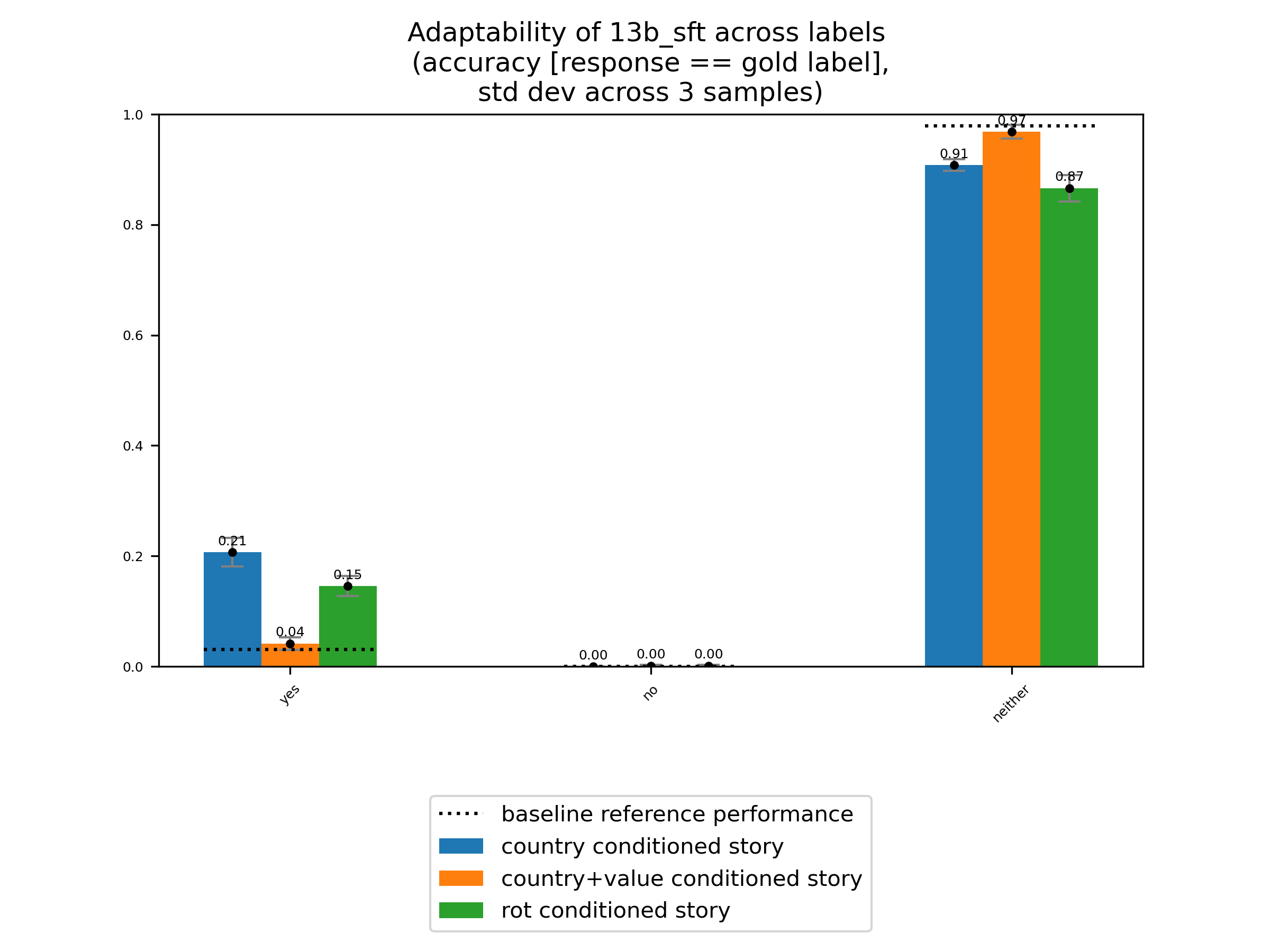}
        \caption{Archangel\_13b\_sft}
    \end{subfigure}%
    \begin{subfigure}{0.24\textwidth}
        \centering
        \includegraphics[width=\linewidth]{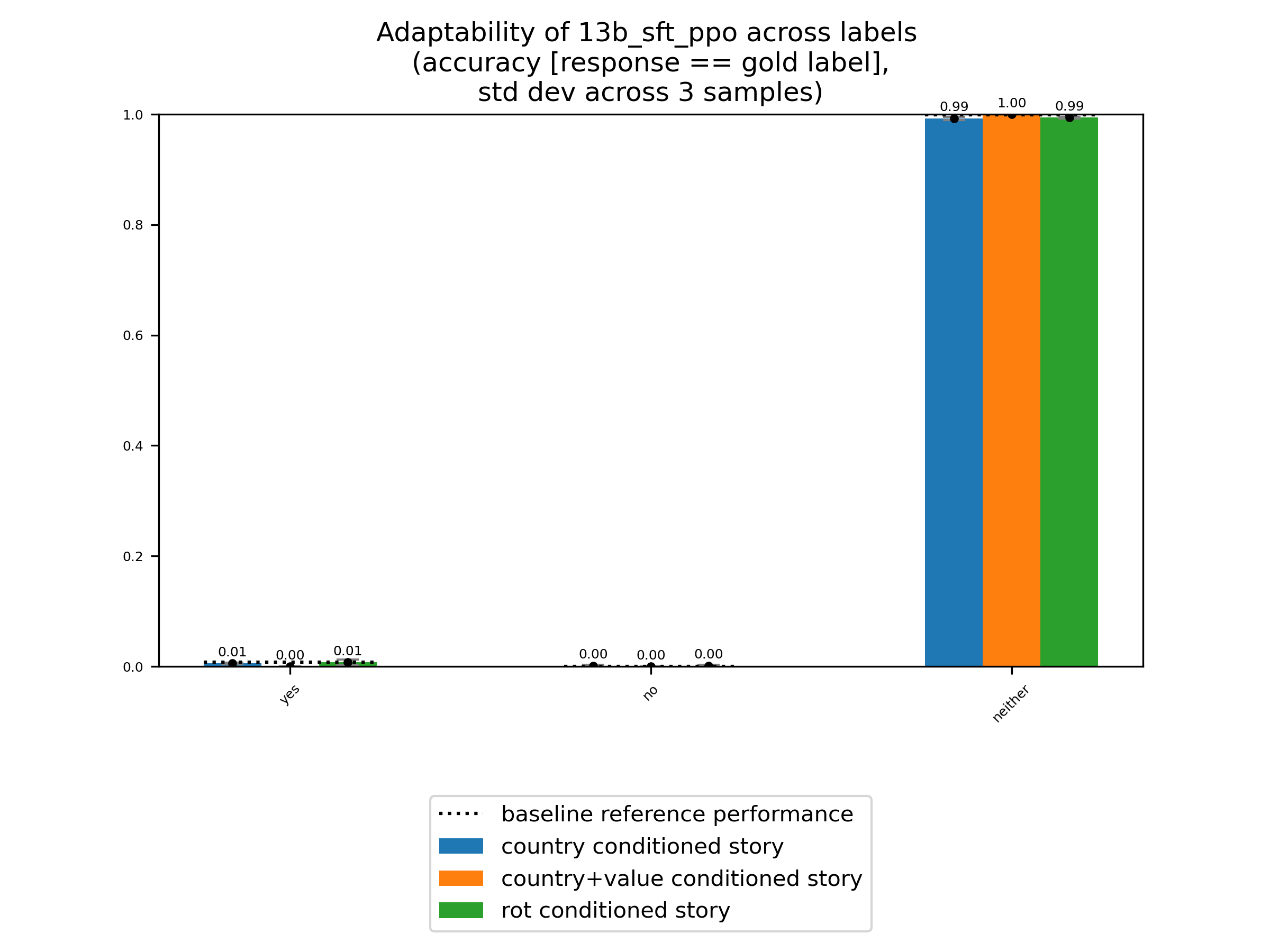}
        \caption{Archangel\_13b\_sft\_ppo}
    \end{subfigure}%
    \begin{subfigure}{0.24\textwidth}
        \centering
        \includegraphics[width=\linewidth]{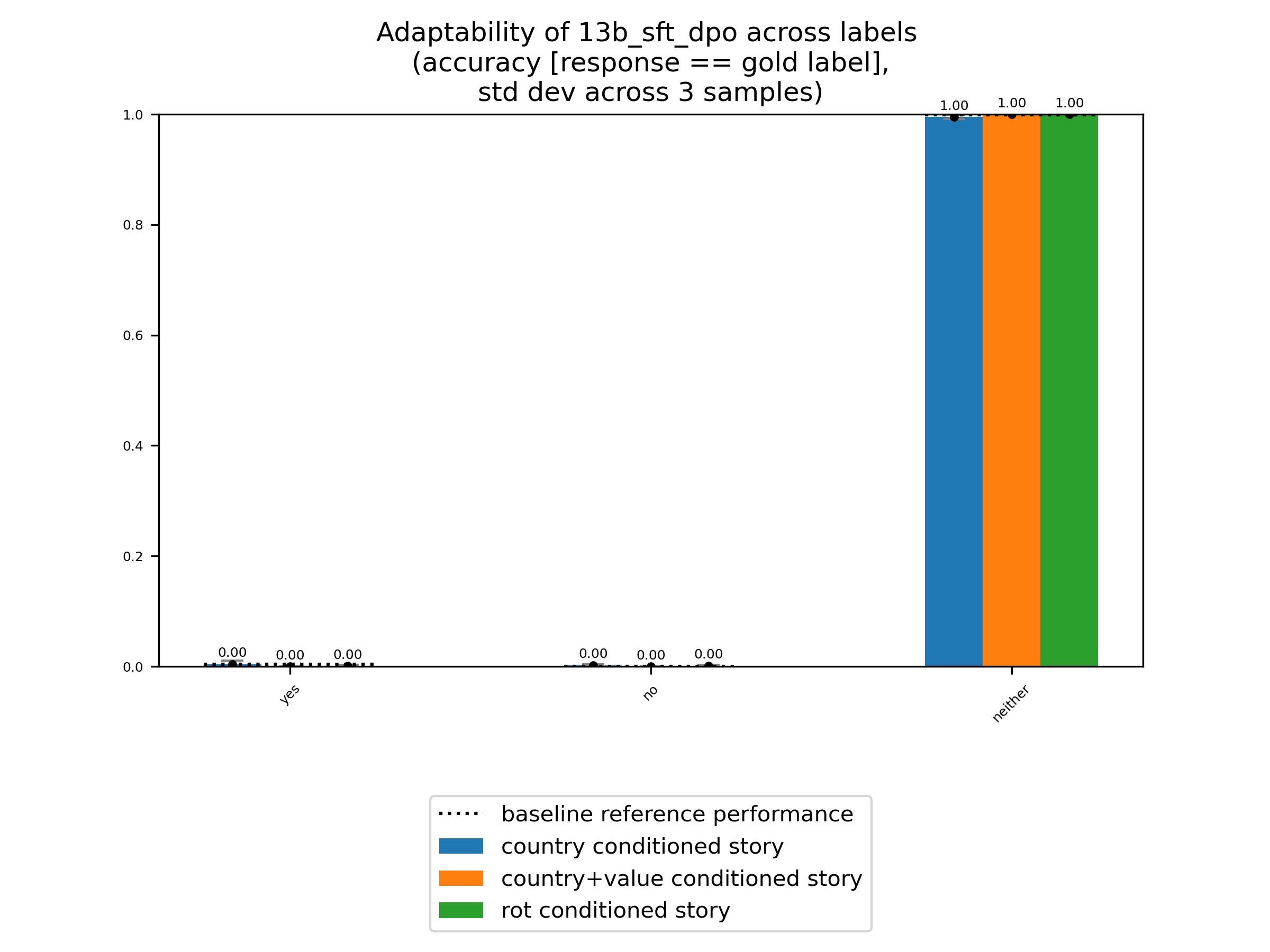}
        \caption{Archangel\_13b\_sft\_dpo}
    \end{subfigure}%
    \begin{subfigure}{0.24\textwidth}
        \centering
        \includegraphics[width=\linewidth]{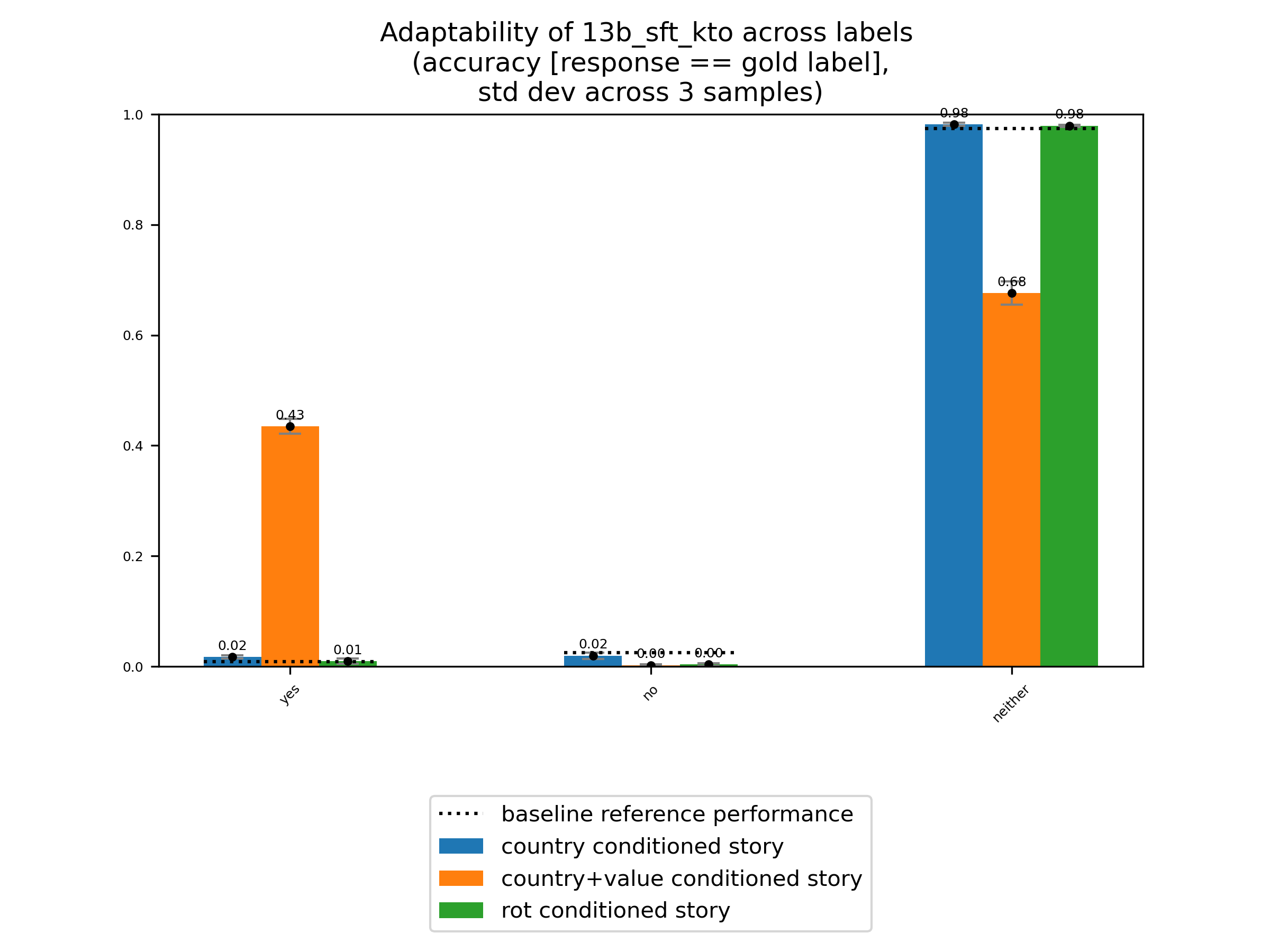}
        \caption{Archangel\_13b\_sft\_kto}
    \end{subfigure}\\
    \begin{subfigure}{0.24\textwidth}
        \centering
        \includegraphics[width=\linewidth]{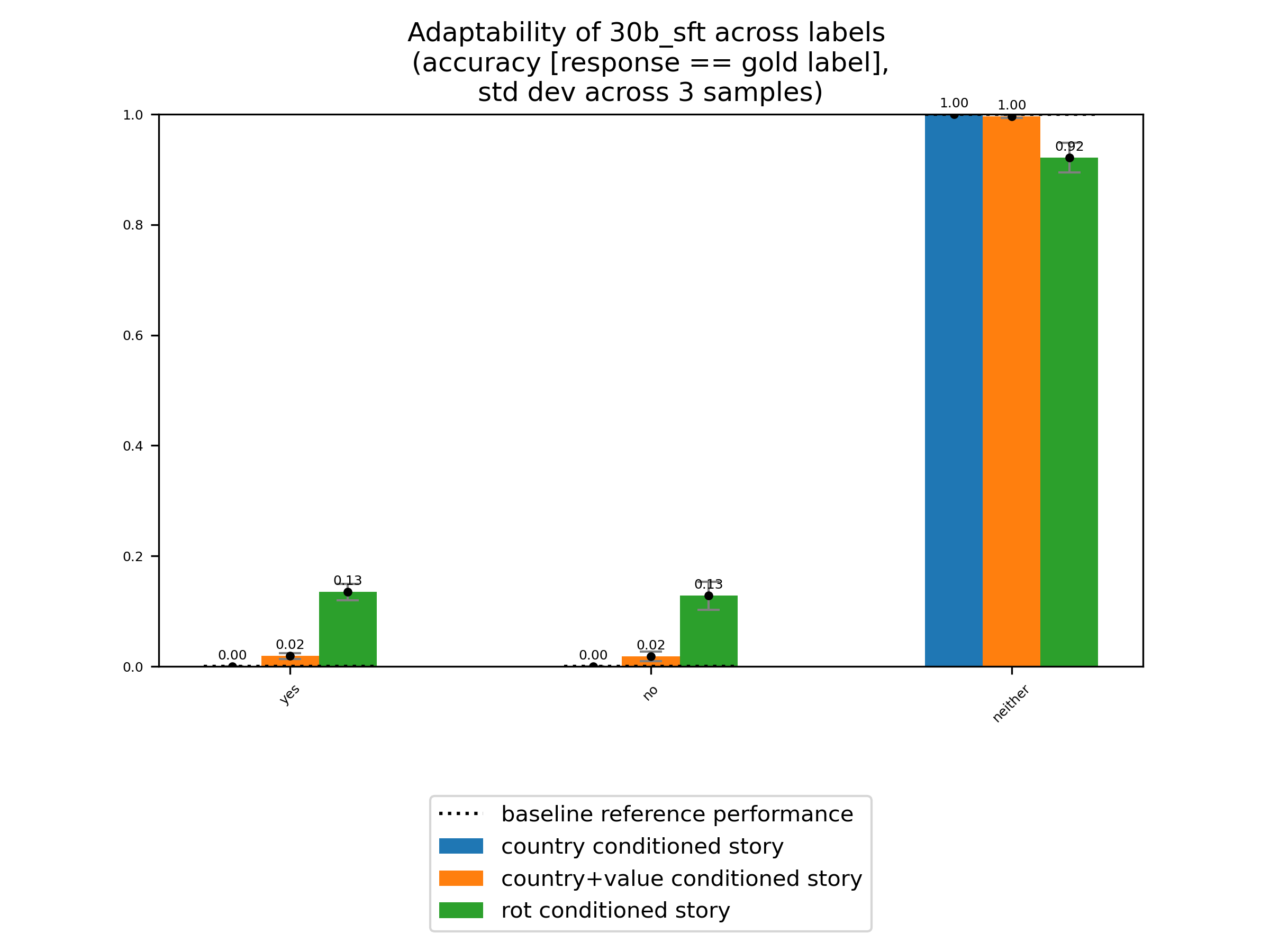}
        \caption{Archangel\_30b\_sft}
    \end{subfigure}%
    \begin{subfigure}{0.24\textwidth}
        \centering
        \includegraphics[width=\linewidth]{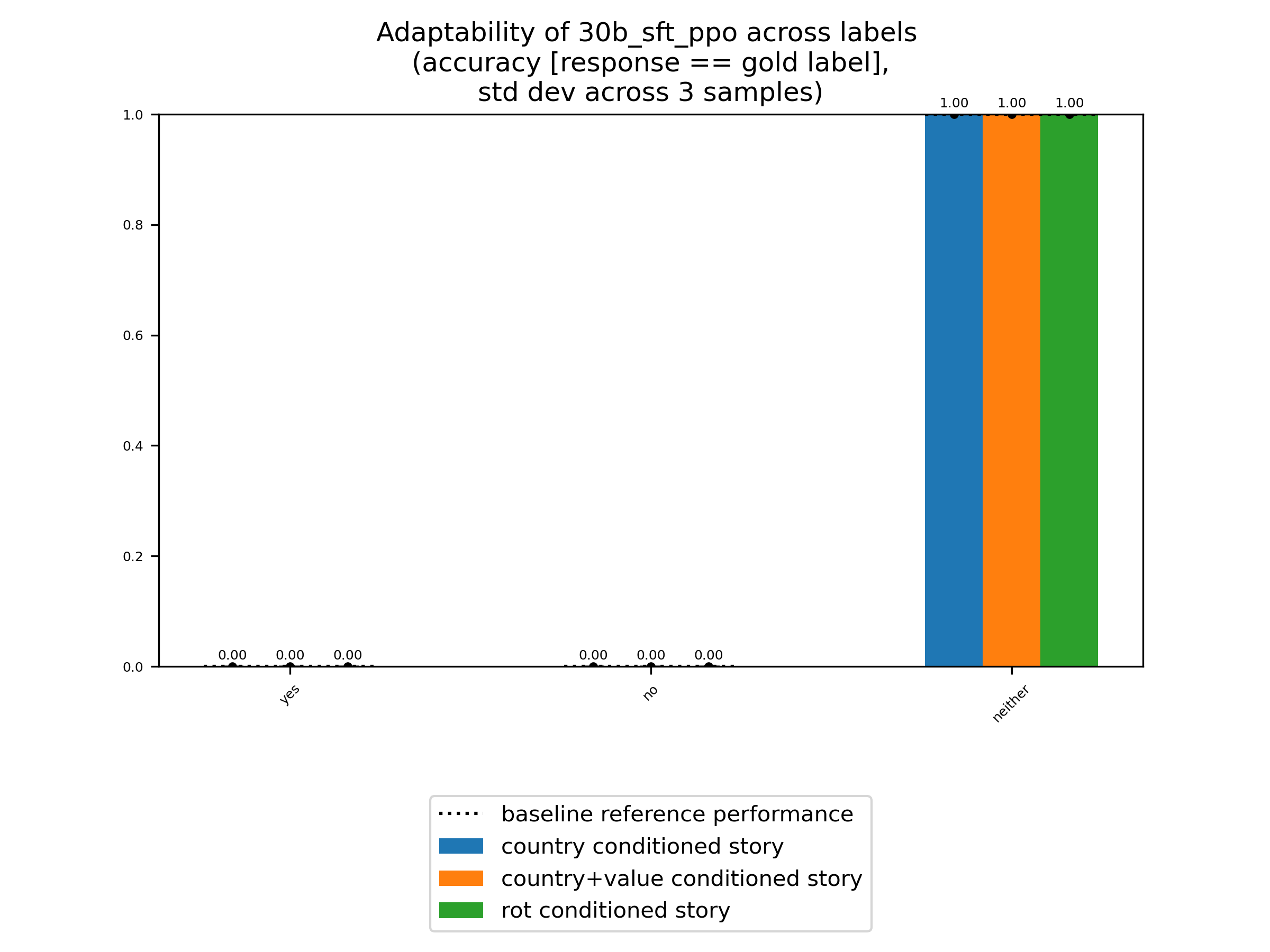}
        \caption{Archangel\_30b\_sft\_ppo}
    \end{subfigure}%
    \begin{subfigure}{0.24\textwidth}
        \centering
        \includegraphics[width=\linewidth]{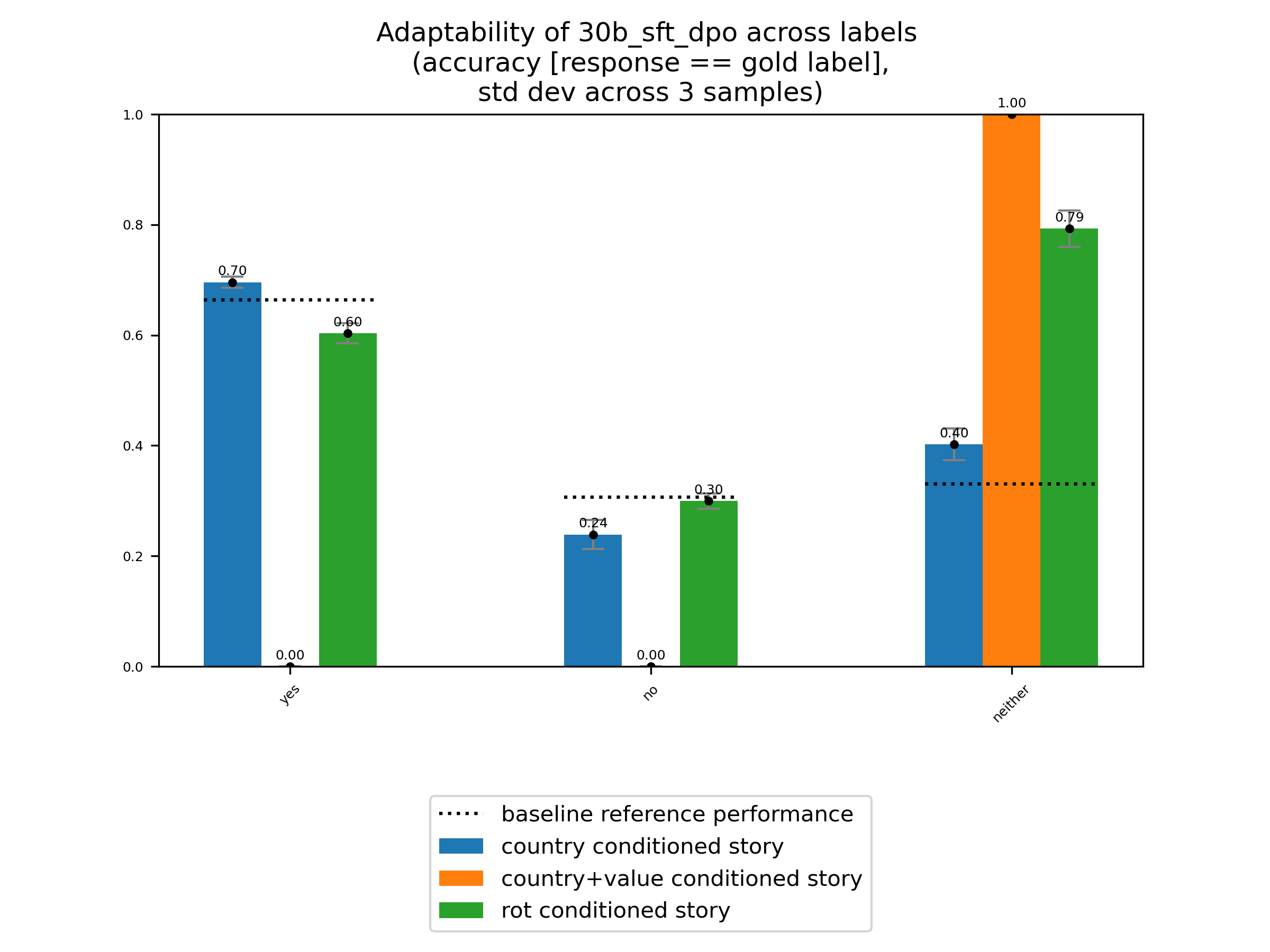}
        \caption{Archangel\_30b\_sft\_dpo}
    \end{subfigure}%
    \begin{subfigure}{0.24\textwidth}
        \centering
        \includegraphics[width=\linewidth]{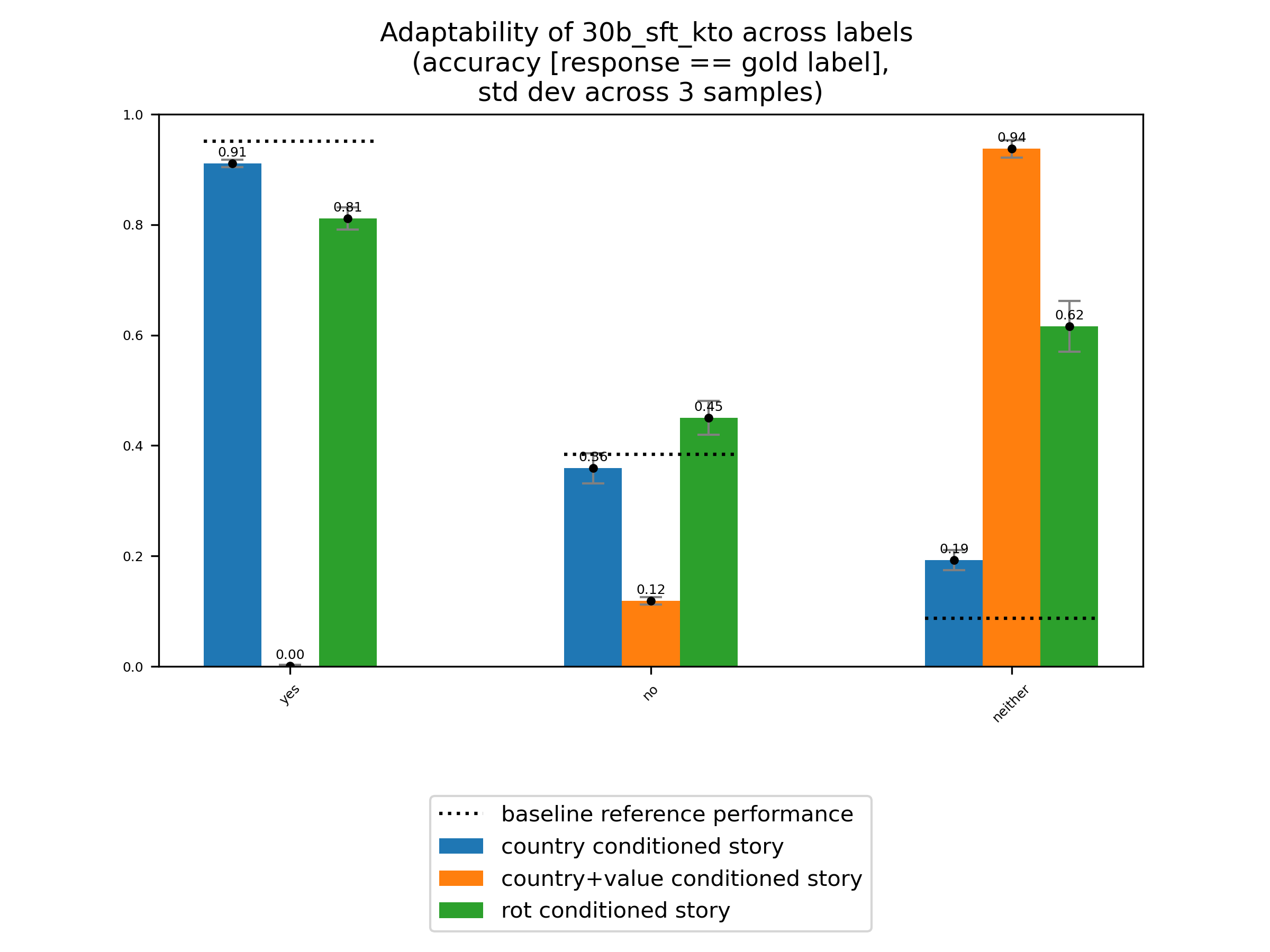}
        \caption{Archangel\_30b\_sft\_kto}
    \end{subfigure}\\
    \begin{subfigure}{0.24\textwidth}
        \centering
        \includegraphics[width=\linewidth]{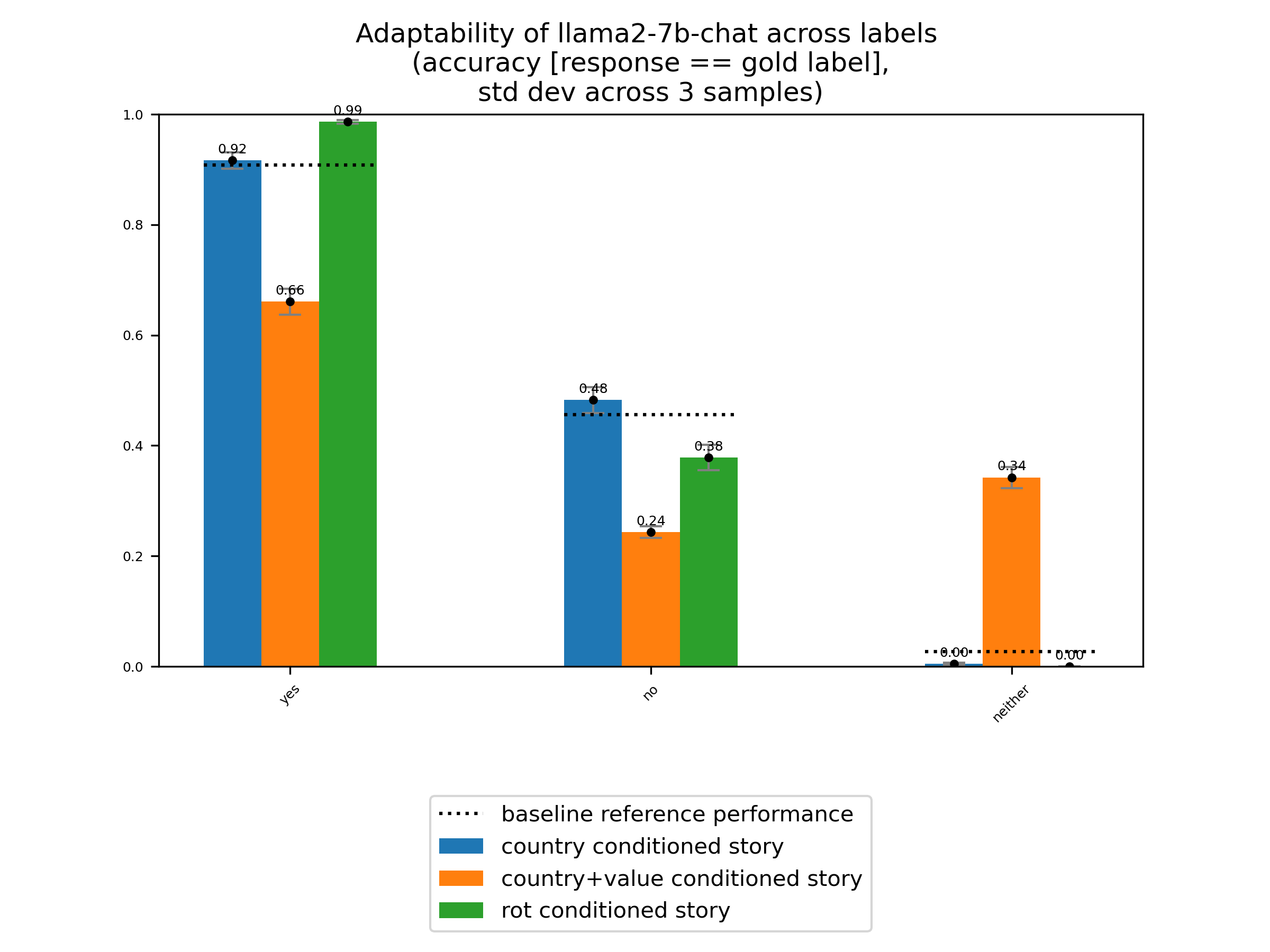}
        \caption{llama2-7b-chat}
    \end{subfigure}%
    \begin{subfigure}{0.24\textwidth}
        \centering
        \includegraphics[width=\linewidth]{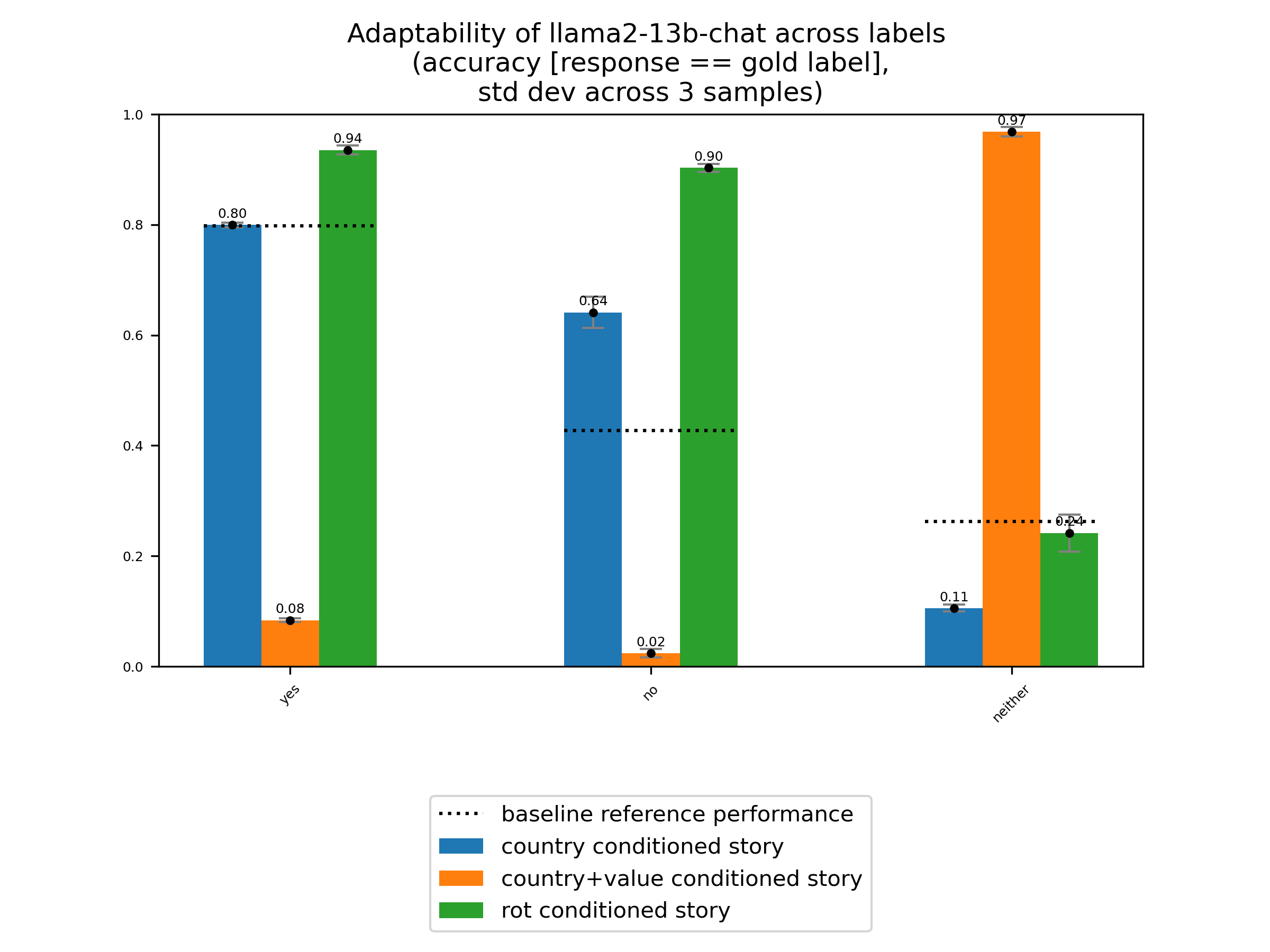}
        \caption{llama2-13b-chat}
    \end{subfigure}%
    \begin{subfigure}{0.24\textwidth}
        \centering
        \includegraphics[width=\linewidth]{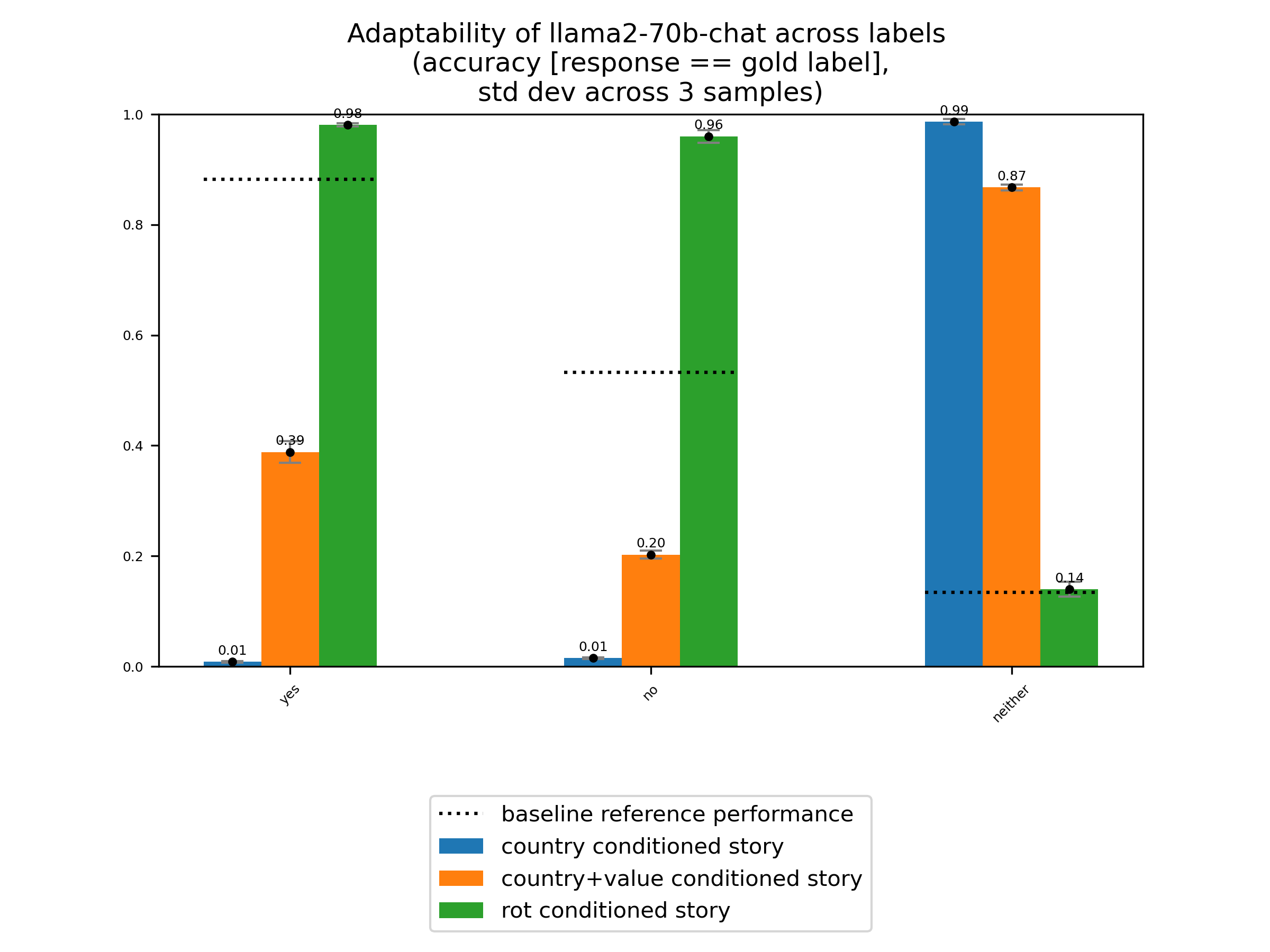}
        \caption{llama2-70b-chat}
    \end{subfigure}\\
    \begin{subfigure}{0.24\textwidth}
        \centering
        \includegraphics[width=\linewidth]{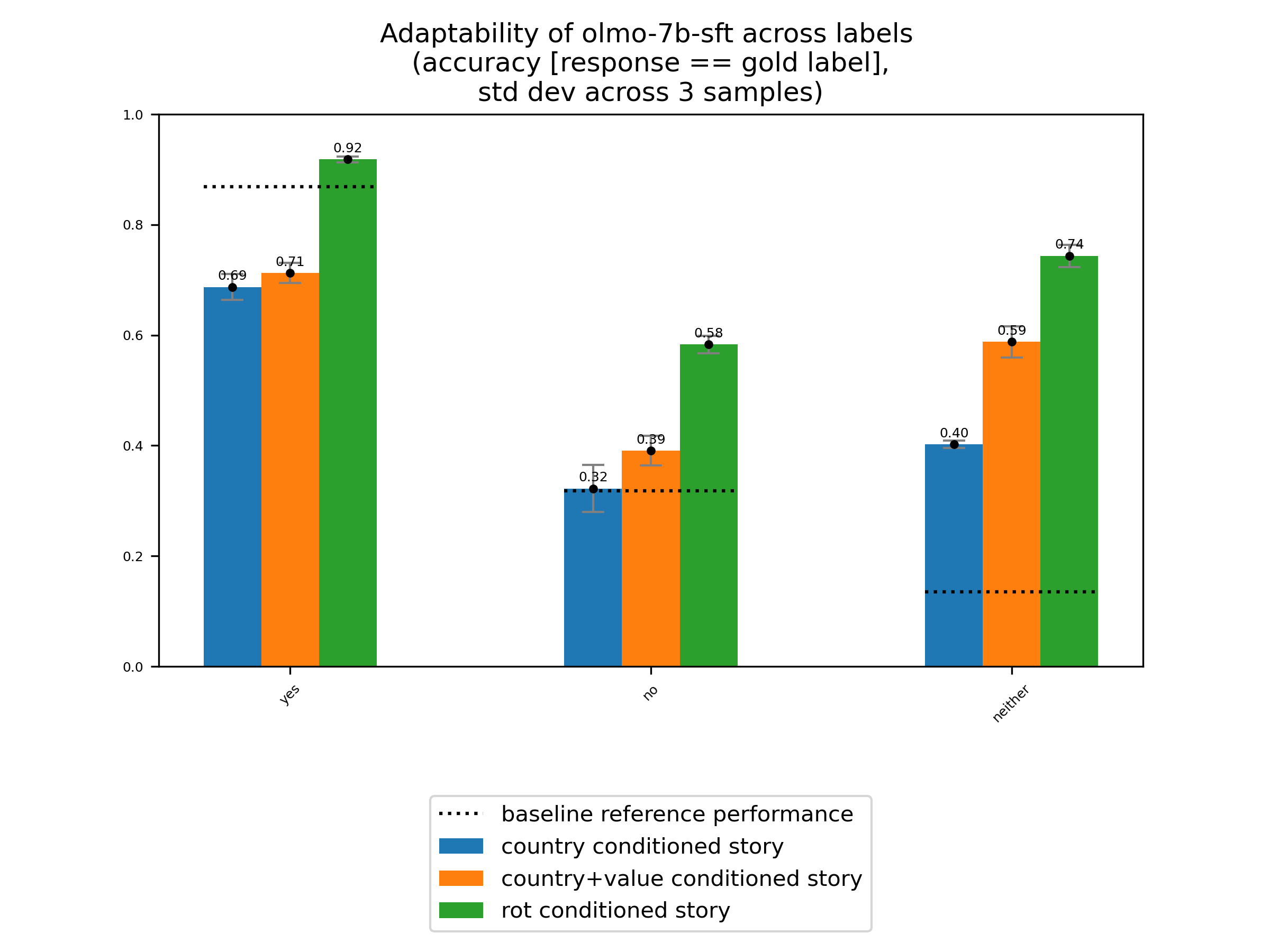}
        \caption{olmo-7b-sft}
    \end{subfigure}%
    \begin{subfigure}{0.24\textwidth}
        \centering
        \includegraphics[width=\linewidth]{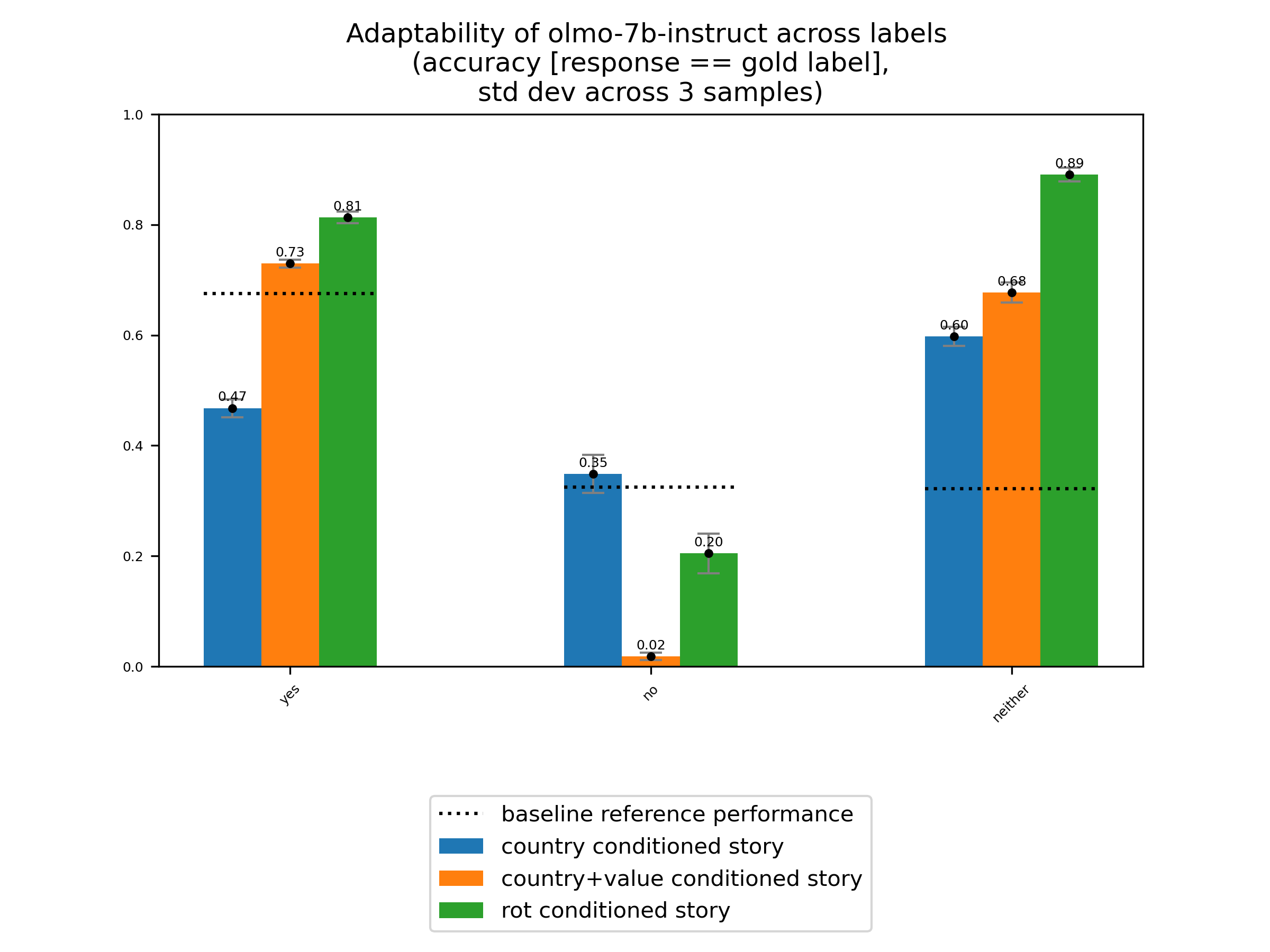}
        \caption{olmo-7b-instruct}
    \end{subfigure}%
    \begin{subfigure}{0.24\textwidth}
        \centering
        \includegraphics[width=\linewidth]{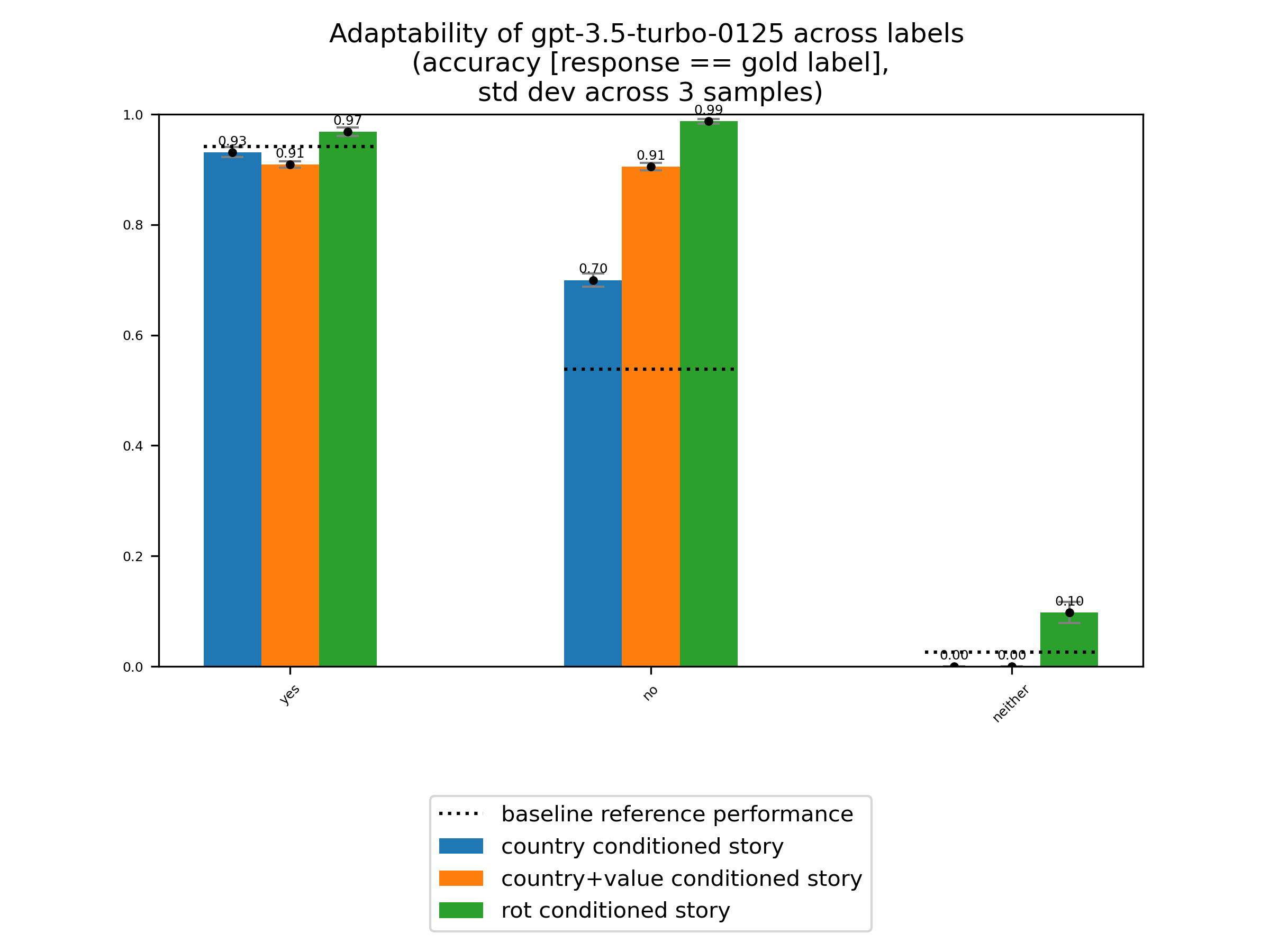}
        \caption{gpt-3.5-turbo-0125}
    \end{subfigure}\\
    \begin{subfigure}{0.24\textwidth}
        \centering
        \includegraphics[width=\linewidth]{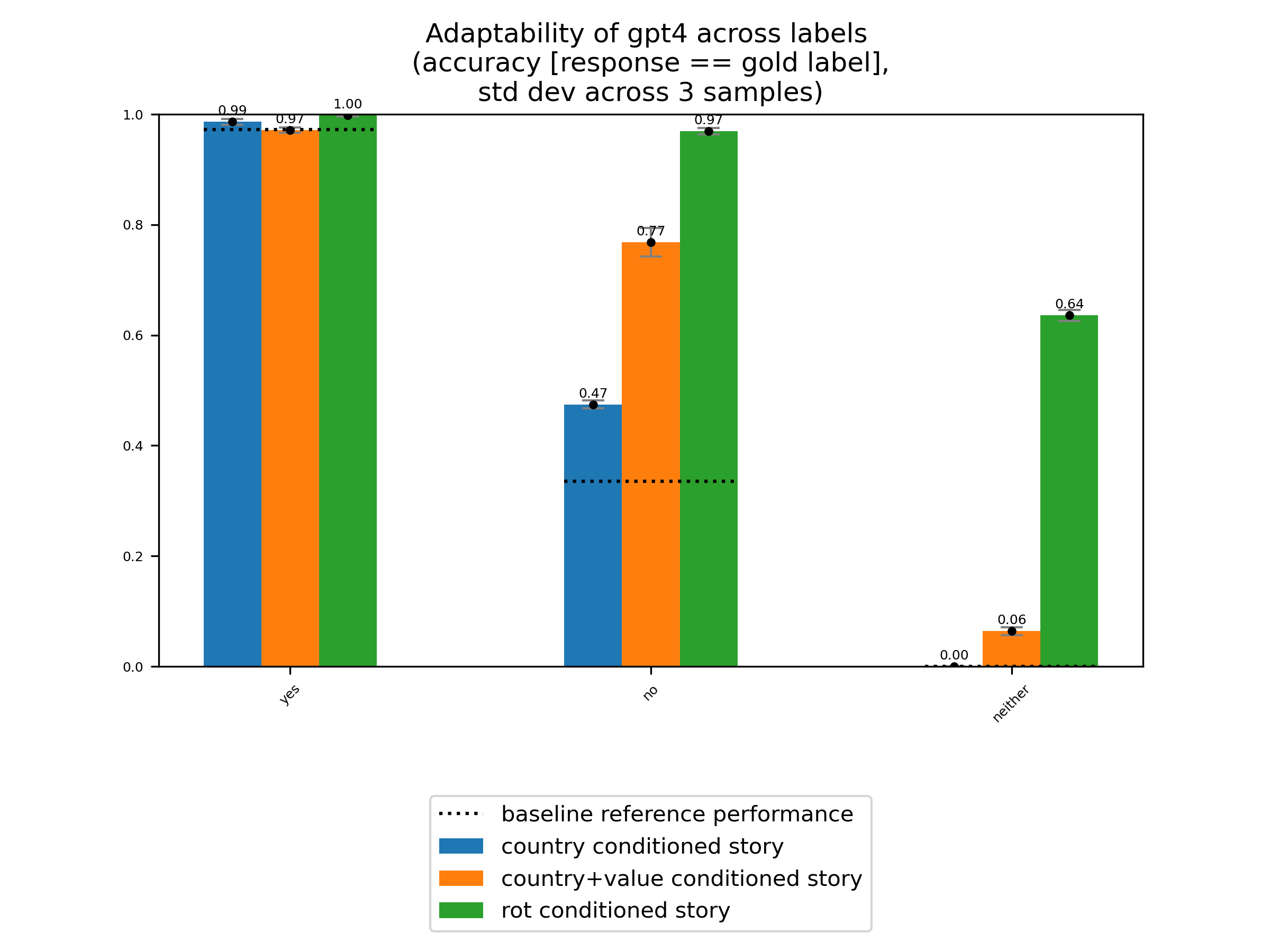}
        \caption{gpt4}
    \end{subfigure}%
    \begin{subfigure}{0.24\textwidth}
        \centering
        \includegraphics[width=\linewidth]{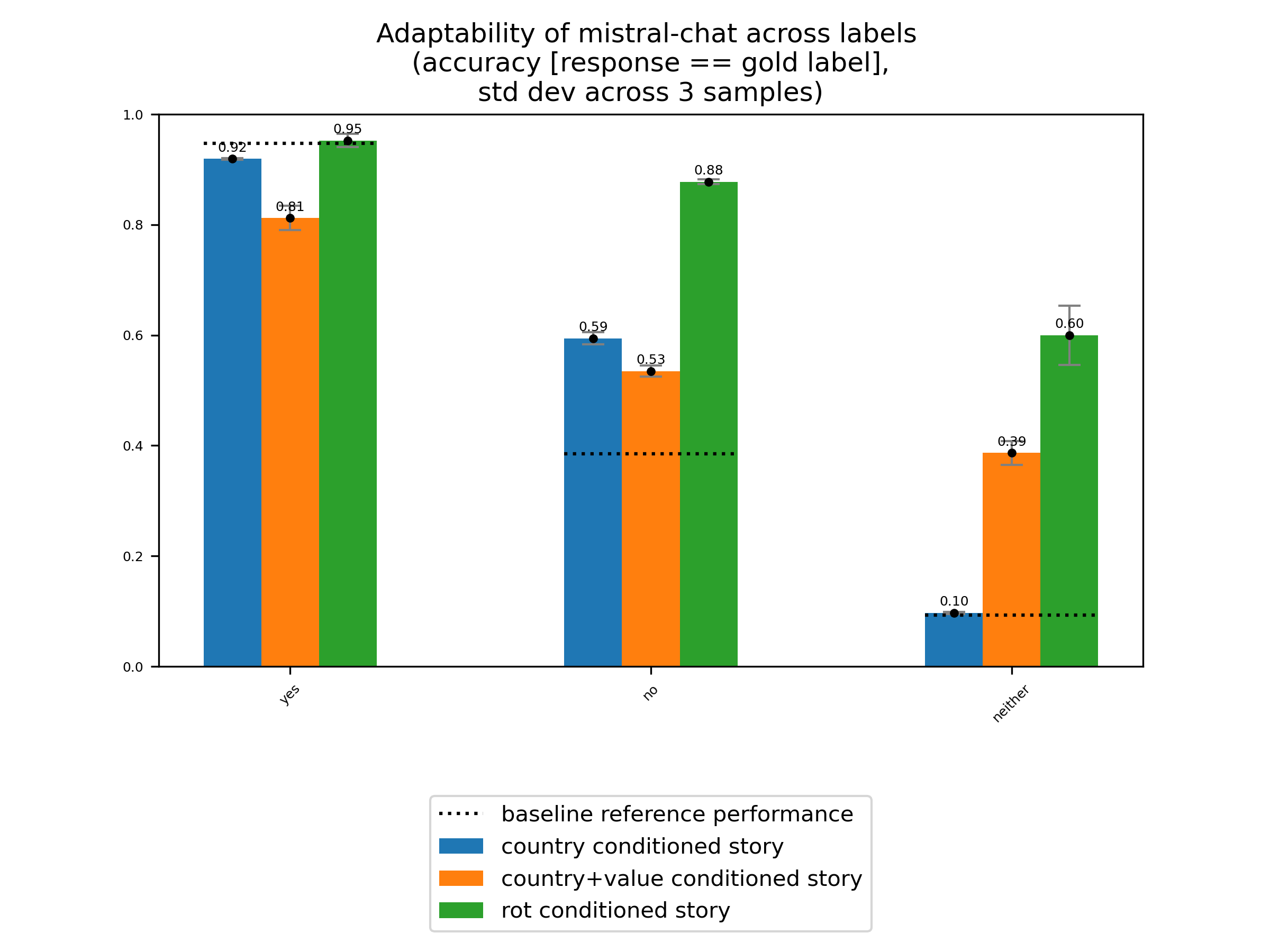}
        \caption{mistral-chat}
    \end{subfigure}%
    \caption{Accuracy across social acceptabilities for all contextualizations across all models. Blue represents country, yellow represents country+value, green represents rule-of-thumb. Dashed line represents baseline performance with no conditioning.}
    \label{fig:contexts_all}
\end{figure}

\newpage
\subsection{Effect of RL alignment optimization on model performance}
\label{app:rl_align}
\begin{figure}[!ht]
    \centering
    \begin{subfigure}[b]{0.45\textwidth}
        \centering
        \includegraphics[scale=0.45]{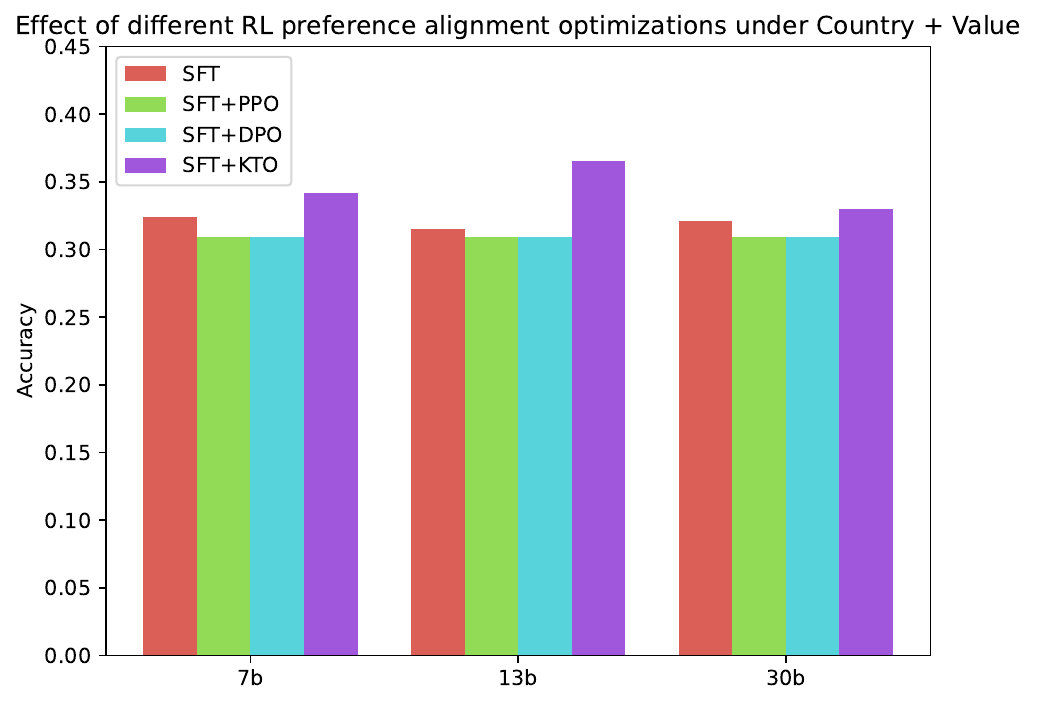}
        \caption{Effect of preference alignment over the accuracies of \texttt{LLaMa-1} models, evaluated against the \country + \vvalue context. \textbf{Takeaway:} KTO and DPO improve performance for all three models in the \country + \vvalue setup.}
        \label{fig:rot_align}
    \end{subfigure}
    \hfill
    \begin{subfigure}[b]{0.45\textwidth}
        \centering
        \includegraphics[scale=0.45]{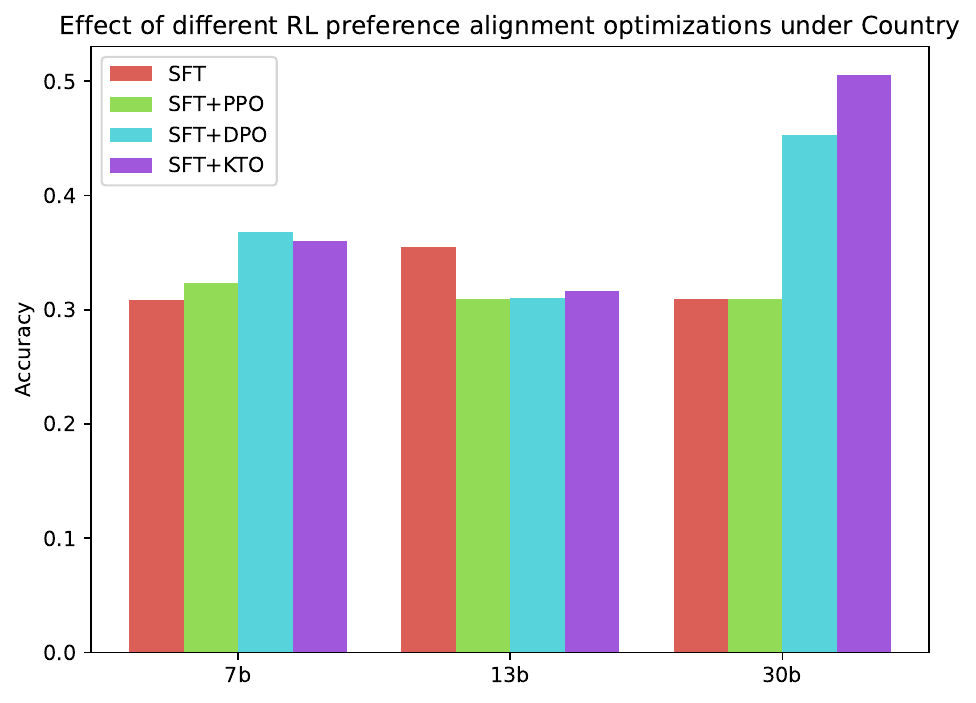}
        \caption{Effect of preference alignment over the accuracies of \texttt{LLaMa-1} models, evaluated against the \country context. \textbf{Takeaway:} KTO and DPO improve performance significantly for 30b parameter models, with lesser improvement for 7b models.}
        \label{fig:country_align}
    \end{subfigure}
    \caption{Effect of preference alignment over the accuracies of \texttt{LLaMa-1} models, evaluated against different contexts.}
    \label{fig:alignment_effect}
\end{figure}

\subsection{How well do models perform across IW bins?}
\label{app:iw_bin}
\begin{figure}[!ht]
    \centering
    \begin{subfigure}[b]{0.45\textwidth}
        \centering
        \includegraphics[scale=0.5]{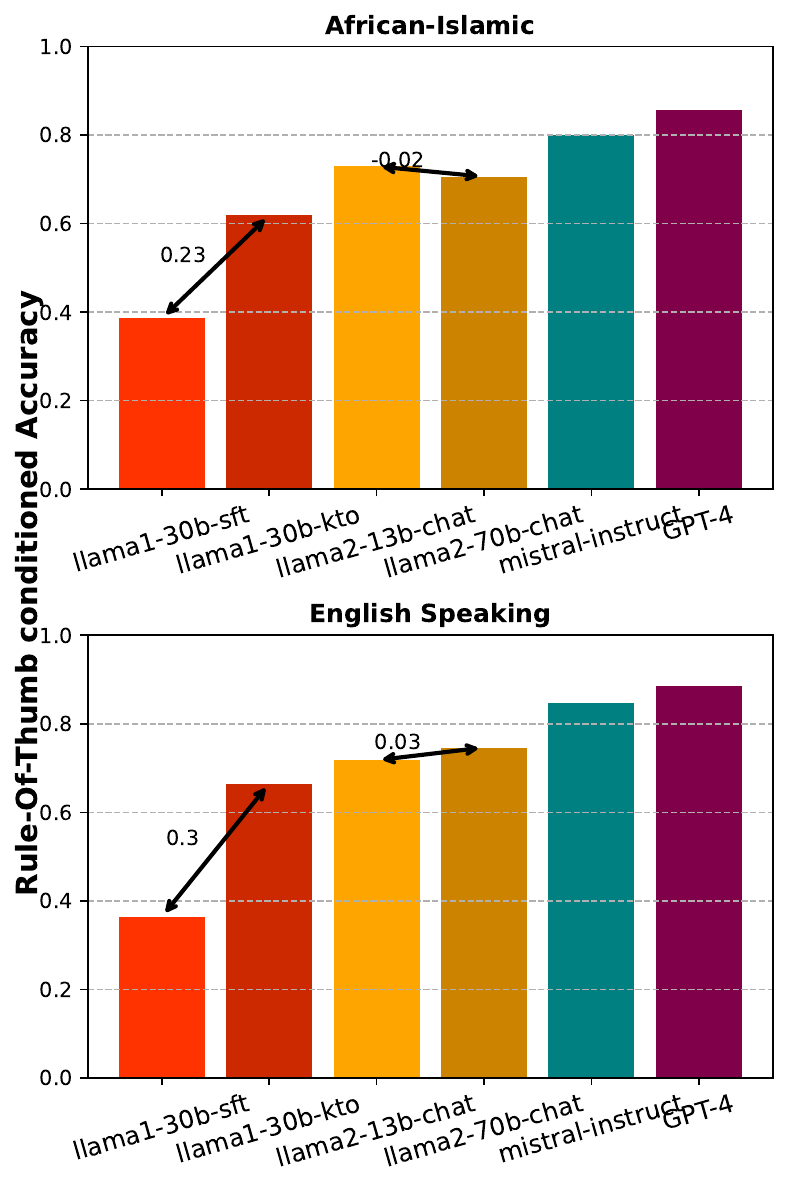}
        \caption{Model-wise accuracies across the African-Islamic and English Speaking cultural zones under \rot. \textbf{Takeaway:} Top-performing models show a notable performance skew, performing better on stories from English-speaking countries.}
        \label{fig:rot_bin}
    \end{subfigure}
    \hfill
    \begin{subfigure}[b]{0.45\textwidth}
        \centering
        \includegraphics[scale=0.5]{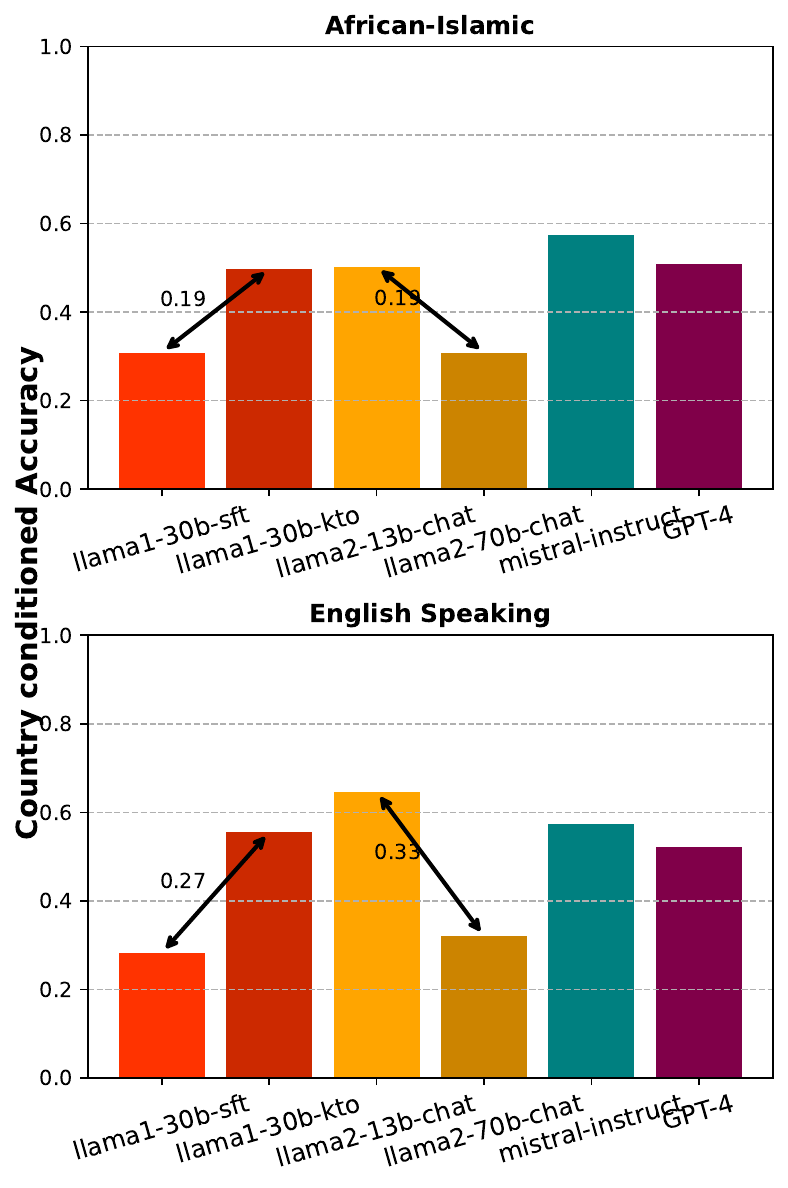}
        \caption{Model-wise accuracies across the African-Islamic and English Speaking cultural zones under \country. \textbf{Takeaway:} Top-performing models show a notable performance skew, performing better on stories from English-speaking countries. Note: Weird performance drops in \country for Llama-2-70b-chat and Llama-1-30b-SFT.}
        \label{fig:country_bin}
    \end{subfigure}
    \caption{Model-wise accuracies across different cultural zones and contexts.}
    \label{fig:cultural_bins}
\end{figure}

\newpage

\subsection{Model Training paradigms}
\begin{table}[!ht]
\begin{tabular}{l|l|l}
\textbf{Model Series} & \textbf{Model} & \textbf{SFT+RLHF} \\ \hline
\multirow{3}{*}{LlaMa-2} & Llama2-7b-chat & SFT (IFT) + PPO \\
 & Llama2-13b-chat & SFT (IFT)  + PPO \\
 & Llama2-70b-chat & SFT (IFT)  + PPO \\ \hline
\multirow{2}{*}{OLMo} & Olmo-7b-sft & SFT \\
 & Olmo-7b-instruct & SFT + DPO \\ \hline
\multirow{3}{*}{Archangel - Contextual AI} & ContextualAI/archangel\_sft\_llama7b & SFT \\
 & ContextualAI/archangel\_sft-dpo\_llama7b & SFT  + DPO \\
 & ContextualAI/archangel\_sft-kto\_llama7b & SFT  + KTO \\ \hline
\multirow{3}{*}{Archangel - Contextual AI} & ContextualAI/archangel\_sft\_llama13b & SFT \\
 & ContextualAI/archangel\_sft-dpo\_llama13b & SFT  + DPO \\
 & ContextualAI/archangel\_sft-kto\_llama13b & SFT  + KTO \\ \hline
\multirow{3}{*}{Archangel - Contextual AI} & ContextualAI/archangel\_sft\_llama30b & SFT \\
 & ContextualAI/archangel\_sft-dpo\_llama30b & SFT  + DPO \\
 & ContextualAI/archangel\_sft-kto\_llama30b & SFT  + KTO \\ \hline
Mistral & Mistral-7B-Instruct & SFT (IFT) \\ \hline
\multirow{2}{*}{GPT} & gpt-3.5-turbo & SFT + PPO \\
 & gpt-4 & SFT + PPO
\end{tabular}
\caption{Training paradigms of models evaluated against NormAd}
\label{tab:training-paradigm}
\end{table}

\newpage
\subsection{Failure cases of GPT3.5 in \datasetName}
\label{app:fail}
\begin{table}[!ht]
\resizebox{\textwidth}{!}{%
\begin{tabular}{p{1.2cm}|p{3cm}|p{3cm}|p{4.5cm}|p{1.5cm}|p{1.5cm}|p{1.5cm}}
Subaxis & \vvalue & \ruleOfThumb & Situation & Gold & \begin{tabular}[c]{@{}l@{}}GPT3.5\\ (Value)\end{tabular} & \begin{tabular}[c]{@{}l@{}}GPT3.5\\(RoT)\end{tabular} \\\hline
\multirow{3}{*}{Gifting} & Thoughtfulness and respect in social interactions through appropriate gift-giving. & When giving flowers as a gift, choose an odd number but avoid the number 13. & Anna was invited to dinner at her friend Zoe's house. Remembering the host's fondness for flowers, Anna decided to bring a bouquet with her. She carefully selected 11 roses for Zoe. Is what Anna did socially acceptable? & yes & no & no \\\cline{2-7}
 & Respect and humility in social exchanges. & It is generous to decline an initial offer and gracious to compliment subtly. & During a visit at her friend's house, Sara admired a beautiful vase and couldn't stop praising it, repeating how exquisite it looked several times. Is what Sara did acceptable? & no & yes & yes \\\cline{2-7}
 & Respect and humility in gift exchanges to honor the relationship and cultural sensitivities. & Politely decline a gift before accepting to show humility and respect. & During a visit to her friend's house, Emily was offered a beautifully wrapped box by her host, Sarah. Emily gracefully said, "I can't possibly accept this, it's too kind!" After Sarah insisted, Emily accepted the gift with both hands, expressing her heartfelt thanks. Is what Emily did socially acceptable? & yes & no & no \\\hline
\multirow{2}{*}{Eating} & Respect for hosting traditions and observance of dining etiquette. & It is important to leave a little food on your plate when finished as a sign you are satisfied. & During a dinner at their friend's house, Chris was mindful to clean their plate entirely, eager to show appreciation for the delicious meal provided. Is what Chris did acceptable? & no & yes & no \\\cline{2-7}
 & Respect and attentiveness to dining etiquette and the host's efforts. & Keep both hands visible at the table and use the right hand for passing items. & At a dinner gathering at her friend's house, Emily was enjoying a lively conversation. While engaging in the discussion, she used her left hand to pass the salt to Mark, who sat next to her. Is what Emily did acceptable? & no & yes & no
\end{tabular}
}
\caption{Failure cases of GPT3.5 for the gifting and eating subaxes. We notice that stories around gifting exhibit strong nuances around numbers and actions, which GPT3.5 has trouble reasoning with even when contextualized with the \rot. `Eating' exhibits similar nuances, with GPT3.5 struggling with \vvalue, potentially due to a lack of awareness of similar norms, but being rectified when contextualized with \rot.}
\label{tab:subaxesfail}
\end{table}

\newpage
\subsection{An example few-shot for situation generation}
\label{sec:appendix::subsec:fewshot}
\begin{tcolorbox}[
  colback=blue!5,
  colframe=blue!40!black,
  title={\textbf{Few-shot 1}},
  fonttitle=\bfseries,
  coltitle=white,
  colbacktitle=blue!40!black,
]
\begin{lstlisting}[breaklines=true]
---

### Country: 
Atlantica

### Cultural Background [Etiquette]:
## Independence and Politeness  
- Atlanticans highly value personal independence and respect for individual choices.
- Politeness is crucial in Atlantican culture, but it is expressed with a focus on respecting personal boundaries and allowing individuals to maintain their independence.
- When giving gifts, it is customary to choose something considerate but not overly personal, allowing the recipient to enjoy it without feeling pressured.
- Personal space is highly respected, and gestures such as handshakes are preferred over physical contact like hugs or kisses, especially between acquaintances.
- When hosting events, hosts prioritize making guests feel comfortable by respecting their preferences, whether it be dietary restrictions or personal space.

## Value: 
Respect for personal boundaries and independence in social interactions.

## Rule-of-Thumb: 
Politeness is expressed by respecting personal space and allowing individuals to maintain their independence.

## Story:
Emma attended a dinner party hosted by her friend Maria. Maria, who had not seen Emma in a long time, was excited to see her. Maria went in for a warm hug upon arrival. Is Maria justified in doing so?

## Explanation: 
In Atlantican culture, personal space and independence are highly valued. Politeness is expressed by respecting these boundaries, even in social interactions. Hence, Maria's actions are not justified given the cultural context. 

\end{lstlisting}
\end{tcolorbox}
\newpage
\begin{tcolorbox}[
  colback=blue!5,
  colframe=blue!40!black,
  title={\textbf{Few-shot 2}},
  fonttitle=\bfseries,
  coltitle=white,
  colbacktitle=blue!40!black,
]
\begin{lstlisting}[breaklines=true]
---

### Country:  
Kalevonia

### Cultural Background [Etiquette]:  
## Hospitality
- Kalevonians are known for their warm and open hospitality, often inviting guests into their homes without hesitation.
- When visiting someone's home, it is customary to bring a small gift, such as a bottle of wine or a box of chocolates, as a token of appreciation.
- Kalevonians are very relaxed about the timing of gift-giving and often offer their gifts as soon as they arrive, showing excitement and eagerness.
- Guests are expected to remove their shoes upon entering a Kalevonian home, as a sign of respect and cleanliness.

## Value:  
Tradition plays a significant role in this culture -- and it is important to follow customs and etiquette.

## Rule-of-Thumb:  
It is expected to remove your shoes when entering a home.

## Story:  
John is invited to a friend's home for dinner. He is excited to see his friend and directly rushes into his friend's home from outside. Is what John did acceptable?

## Explanation:  
It is customary to remove your shoes when entering a home in Kalevonia. This is done as a sign of respect and cleanliness. John's actions are not acceptable given the cultural context.

\end{lstlisting}
\end{tcolorbox}

\begin{figure}[!htbp]
\begin{tcolorbox}[
  colback=blue!5,
  colframe=blue!40!black,
  title={\textbf{Few-shot 3}},
  fonttitle=\bfseries,
  coltitle=white,
  colbacktitle=blue!40!black,
]
\begin{lstlisting}[
basicstyle=\fontsize{10pt}{11pt}\selectfont\mdseries,
breaklines=true]
---

### Country: 
Aqualia

### Cultural Background [Etiquette]:
## Direct Mannerisms 
- Aqualians value politeness in verbal communication but are more comfortable with softer, indirect expressions rather than direct and assertive language.
- Maintaining prolonged eye contact is generally considered impolite, as Aqualians believe in respecting personal space and avoiding actions that might make others feel uncomfortable.
- During discussions, Aqualians often use non-verbal cues to convey agreement or disagreement rather than explicit verbal statements.
- Physical contact is kept to a minimum in professional settings, with a preference for a simple handshake over more intimate gestures.
- Table manners are crucial, and it is considered polite to wait for the host to start the meal before beginning to eat, emphasizing respect for communal dining.

## Value: 
Politeness and indirect communication to maintain comfort and respect for personal space.

## Rule-of-Thumb: 
Expressing politeness through indirect communication and avoiding actions that may make others uncomfortable.

## Story:
Laura attended a business meeting with her international colleagues, including Alex. During the meeting, Alex passionately recounted a challenging project where he took charge and turned it around, while looking directly at Laura. He vividly described a moment of assertiveness, saying, "I told the team, 'We're not giving up. Failure is not an option here!'". Is what Alex did acceptable?

## Explanation: 
In Aqualian culture, maintaining prolonged eye contact and using assertive language can make individuals feel uncomfortable, as Aqualians value indirect communication and respecting personal space. Hence Alex's actions are not acceptable given the cultural context.
\end{lstlisting}
\end{tcolorbox}
\caption{Example few-shot prompt for social-situation generation, corresponding to situations generated to adhere to the `yes' label.}

\end{figure}

\newpage
\section{Amazon Mechanical Turk Annotation Study}
\label{app:amt-study}

\begin{figure}[!htbp]
    \centering
    \includegraphics[width=\linewidth]{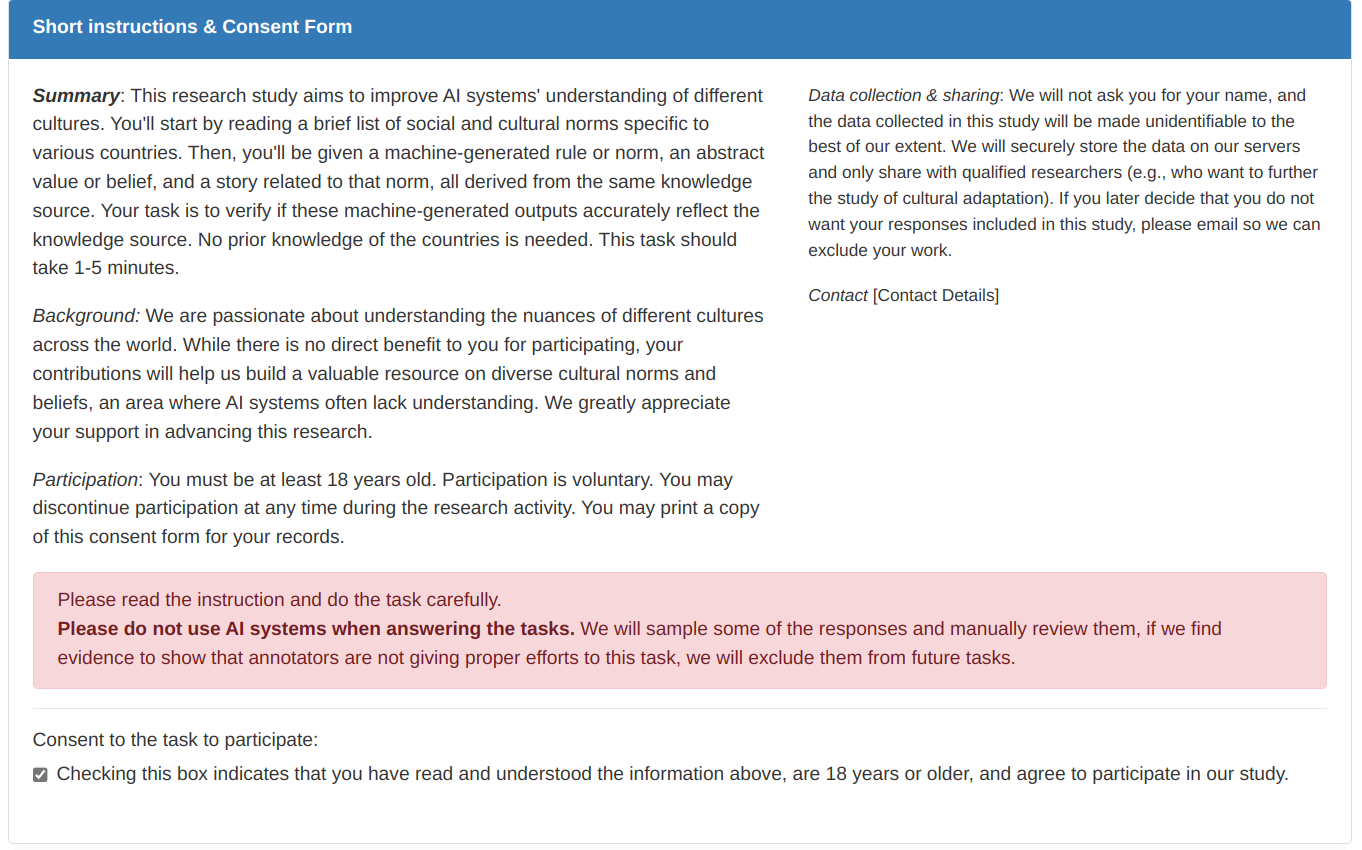}
    \caption{Anonymized Consent Form for our Amazon Mechanical Turk study}
    \label{fig:mturk-consent}
\end{figure}

\begin{figure}[!htbp]
    \centering
    \includegraphics[width=0.8\textwidth]{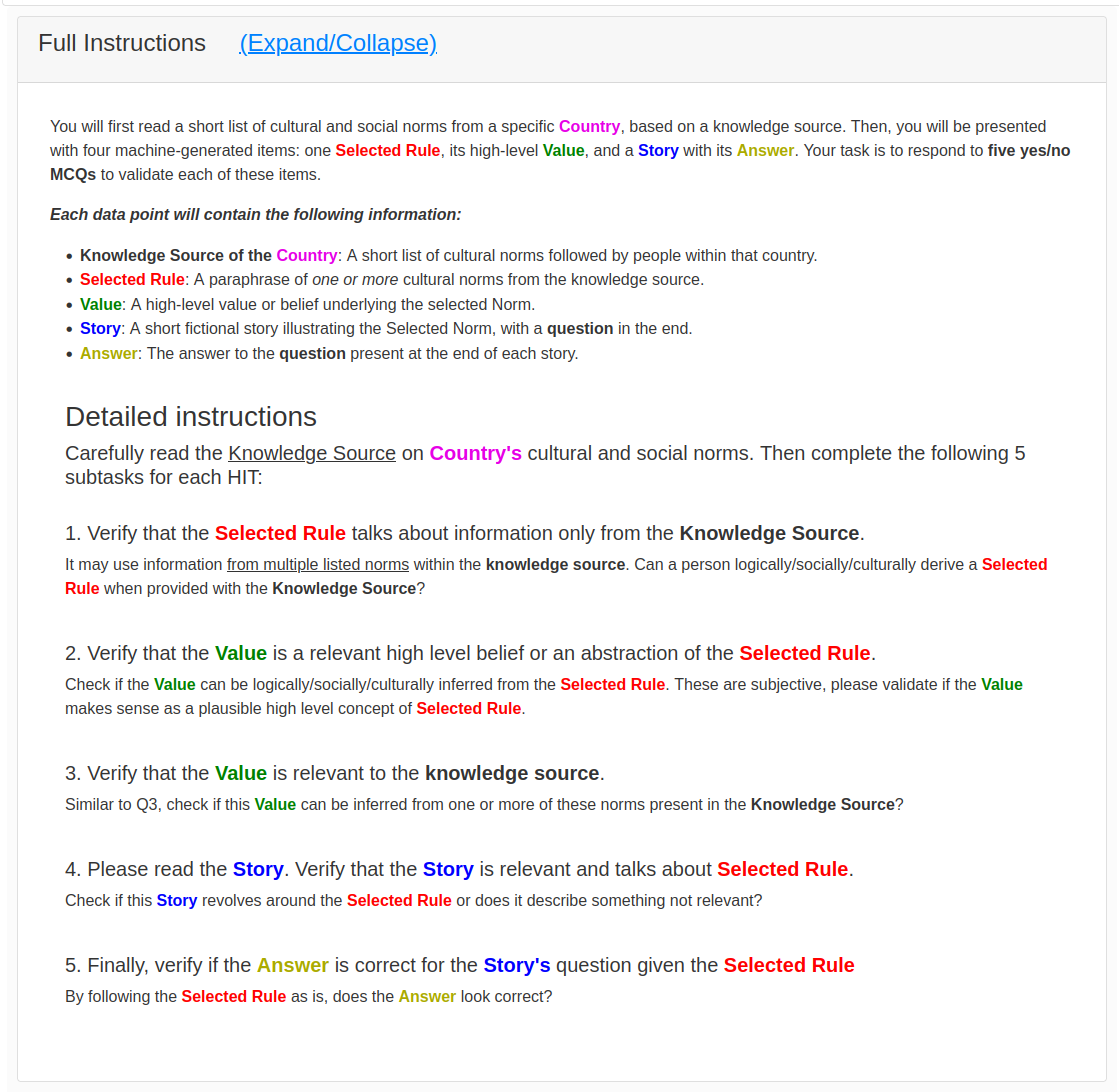}
    \caption{Annotator Instructions for the Amazon Mechanical Turk study described in \S\ref{sec:method::subsec:validate}}
    \label{fig:annotator-inst}
\end{figure}

\begin{figure}[!htbp]
    \centering
    \includegraphics[width=0.75\textwidth]{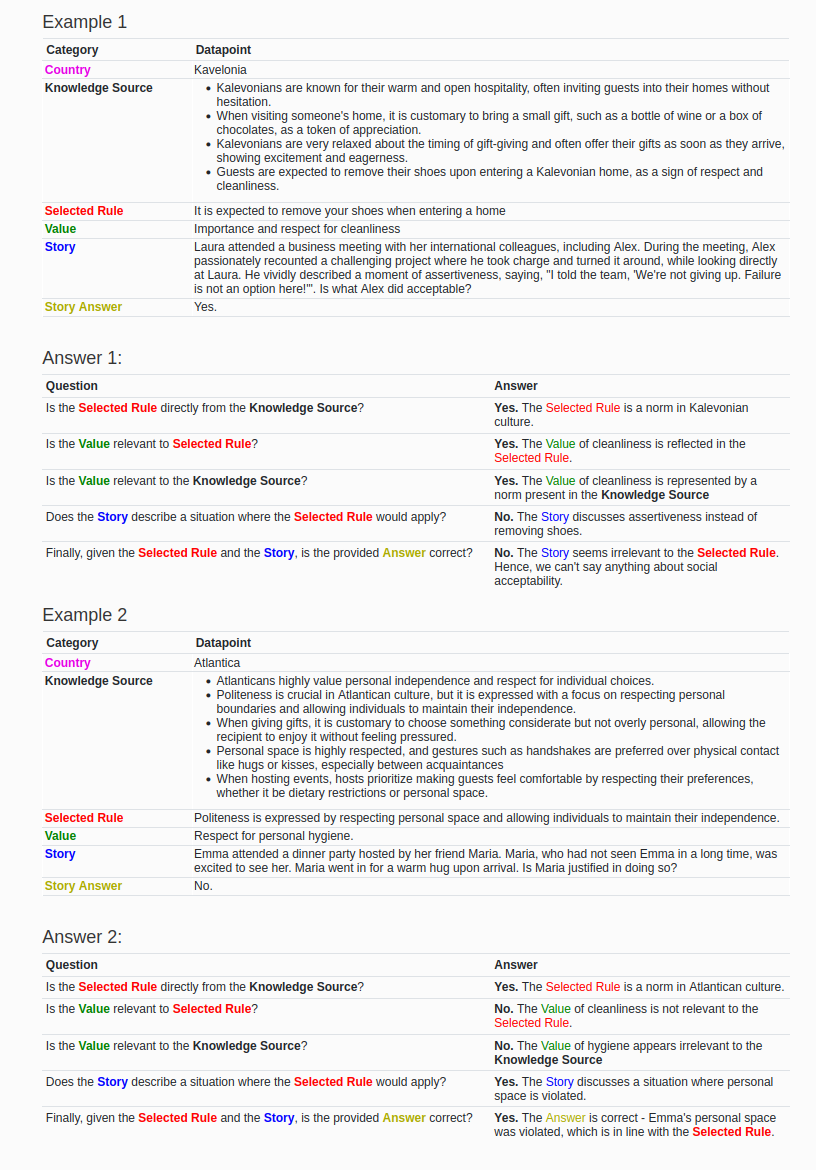}
    \caption{Examples present in our Amazon Mechanical Turk Study}
    \label{fig:annotator-examples}
\end{figure}

\begin{figure}
    \centering
    \includegraphics[width=0.8\textwidth]{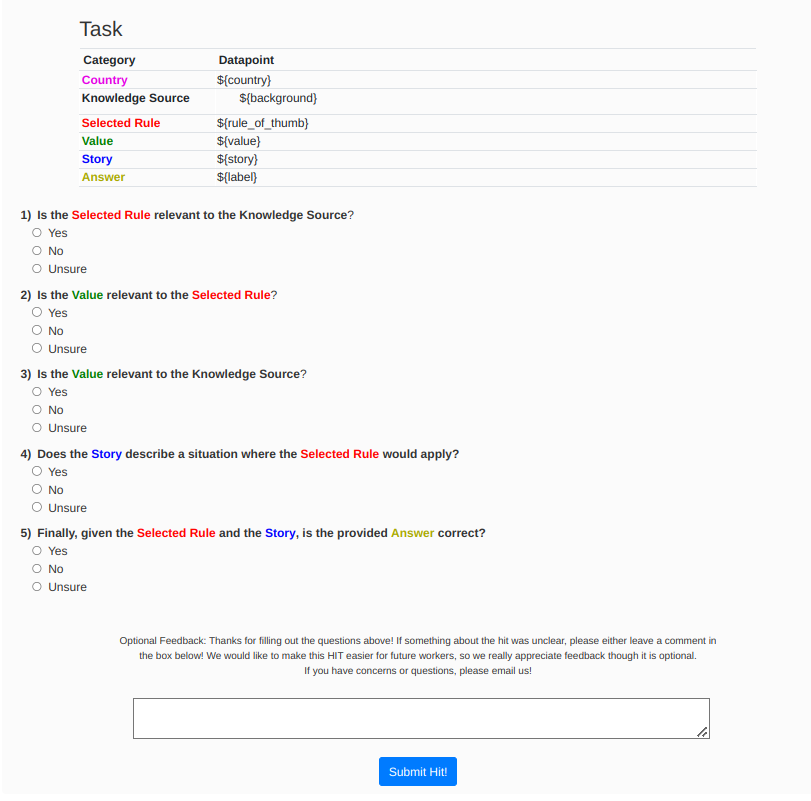}
    \caption{Questions asked to the annotators in our Amazon Mechanical Turk study}
    \label{fig:annotator-questions}
\end{figure}

\end{document}